%% file: aistats2020-arxiv.tex
\let\mypdfximage\pdfximage
\def\pdfximage{\immediate\mypdfximage}
\documentclass[twoside]{article}

\usepackage[utf8]{inputenc}

\usepackage[accepted]{aistats2019}
\usepackage{microtype}
\usepackage{graphicx}
\usepackage{booktabs} 

\usepackage{algorithm2e}
\usepackage{setspace}
\usepackage{listings}

\usepackage{color}

\usepackage{subcaption}

\usepackage{graphicx}

\usepackage{array}

\usepackage{multirow}

\input{math_commands.tex}

\newcommand{\citep}{\cite}
\newcommand{\citet}{\cite}

\newcommand{\note}[1]{} 
\newcommand{\todo}[1]{} 

\newcommand{\blindfootnote}[1]{{\renewcommand\thefootnote{}\footnotetext{#1}}}

\begin{document}

%
\runningtitle{Independent Subspace Analysis for Unsupervised Learning of Disentangled Representations}

%

\twocolumn[

\aistatstitle{Independent Subspace Analysis for \\
           Unsupervised Learning of Disentangled Representations}

\aistatsauthor{ Jan St\"uhmer \And Richard E. Turner \And  Sebastian Nowozin }

\aistatsaddress{ Microsoft Research \And  University of Cambridge \And Google Brain\thanks{} } ]

\begin{abstract}
Recently there has been an increased interest in unsupervised learning of disentangled representations using the Variational Autoencoder (VAE) framework. Most of the existing work has focused largely on modifying the variational cost function to achieve this goal. We first show that these modifications, e.g. $\beta$-VAE, simplify the tendency of variational inference to underfit causing pathological over-pruning and over-orthogonalization of learned components. Second we propose a complementary approach: to modify the probabilistic model with a structured latent prior. This prior allows to discover latent variable representations that are structured into a hierarchy of independent vector spaces. The proposed prior has three major advantages: First, in contrast to the standard VAE normal prior the proposed prior is not rotationally invariant. This resolves the problem of unidentifiability of the standard VAE normal prior. Second, we demonstrate that the proposed prior encourages a disentangled latent representation which facilitates learning of disentangled representations. Third, extensive quantitative experiments demonstrate that the prior significantly mitigates the trade-off between reconstruction loss and disentanglement over the state of the art. 
\end{abstract}


\section{Introduction}

Recently there has been an increased interest in unsupervised learning of disentangled representations. The term \emph{disentangled} usually describes two main objectives: First, to identify each true factor of variation with a latent variable, and second, interpretability of these latent factors~\citep{Schmidhuber-1992,Ridgeway-2016,Achille-Soatto-2017}. Most of this recent work is inspired by the $\beta$-VAE concept introduced in~\citet{Higgins-et-al-2016}, which proposes to re-weight the terms in the evidence lower bound~(ELBO) objective. In~\citet{Higgins-et-al-2016} a higher weight for the Kullback-Leibler divergence~(KL) between approximate posterior and prior is proposed, and putative mechanistic explanations for the effects of this modification are studied in~\citet{Burgess-et-al-2018,Chen-et-al-2018}. Two recent approaches, \citet{Kim-and-Mnih-2018} and \citet{Chen-et-al-2018}, propose to penalize the total correlation between the dimensions of the latent representation, therefore encouraging a factorized distribution.
\blindfootnote{$^1$ Work done while at Microsoft Research.}
\todo{discuss DIP-VAE Kumar et al. [30 in Locatello]}

These modifications of the evidence lower bound however lead to a trade-off between disentanglement and reconstruction loss and therefore the quality of the learned model. This trade-off is directly encoded in the modified objective: by increasing the $\beta$-weight of the KL-term, the relative weight of the reconstruction loss term is more and more decreased. Therefore, optimization of the modified ELBO will lead to latent encodings which have a lower KL-divergence from the prior, but at the same time lead to a higher reconstruction loss. Furthermore, we discuss in \secref{sec:betavae} that using a higher weight for the KL-term amplifies existing biases of variational inference, potentially to a catastrophic extent.

There is a foundational contradiction in many approaches to disentangling deep generative models (DGMs):
the standard model employed is not identifiable as it employs a standard normal prior which then undergoes a linear transformation.
Any rotation of the latent space can be absorbed into the linear transform and is therefore statistically indistinguishable. If interpretability is desired, the modelling choices are setting us up to fail.


\newpage
We make the following contributions:
\begin{itemize}
\item
We show that current state of the art approaches based on modified cost functions employ a trade-off between reconstruction loss and disentanglement of the latent representation.
\item
In \secref{sec:ortho} we show that variational inference techniques are biased: the estimated components are biased towards having orthogonal effects on the data and the number of components is underestimated.
\item
We provide a novel description of the origin of disentanglement in $\beta$-VAE
and demonstrate in \secref{sec:betavae} that increasing the weight of the KL term increases the over-pruning bias of variational inference.
\item
To mitigate these drawbacks of existing approaches, we propose a family of rotationally asymmetric distributions for the latent prior, which removes the rotational ambiguity from the model.
\item
The prior allows to decompose the latent space into independent subspaces. Experiments demonstrate that this prior facilitates disentangled representations even for the unmodified ELBO objective.
\item Extensive quantitative experiments demonstrate that the prior significantly mitigates the trade-off between disentanglement and reconstruction quality.
\end{itemize}

\section{Background}

We briefly discuss previous work on variational inference in deep generative models and two modifications of the learning objective that have been proposed to learn a disentangled representation. We discuss characteristic biases of variational inference and how the modifications of the learning objective actually accentuate these biases.

\subsection{Disentangled Representation Learning}

\paragraph{Variational Autoencoder} The variational autoencoder introduced in
\citet{Kingma-Welling-2014} combines a generative model, the decoder, with an inference network, the encoder. Training is performed by optimizing the \emph{evidence lower bound} (ELBO) averaged over the empirical distribution:

{
~\\[-16pt]
\small
\begin{equation}
\mathcal{L}_{\textrm{ELBO}} =
\mathbb{E}_{q_\phi(\vz|\vx)} \left[ \log p_\theta(\vx|\vz) \right]
- D_{\textrm{KL}}(q_\phi(\vz|\vx) \| p(\vz))
\, ,
\end{equation}
\normalsize
}
where the decoder $p_\theta(\vx | \vz)$ is a deep learning model with parameters $\theta$
and $\vz$ is sampled from the encoder $\vz \sim q_\phi(\vz|\vx)$ with variational parameters $\phi$. 
When choosing appropriate families of distributions, gradients through the samples $\vz$ can be estimated using the \emph{reparameterization trick}. The approximate posterior $q_\phi(\vz|\vx)$ is usually modelled as a multivariate Gaussian with diagonal covariance matrix and the prior $p(\vz)$ is typically the standard normal distribution.

\paragraph{$\beta$-VAE}

\citet{Higgins-et-al-2016} propose to modify the evidence lower bound objective and penalize the KL-divergence of the ELBO:

~\\[-16pt]
\small
\begin{equation}
\mathcal{L}_{\beta\textrm{-ELBO}} =
\mathbb{E}_{q_\phi(\vz|\vx)} \left[ \log p_\theta(\vx|\vz) \right]
- \beta D_{\textrm{KL}}(q_\phi(\vz|\vx) \| p(\vz))
,
\label{eq:betavae}
\end{equation}
\normalsize
where $\beta>1$ is a free parameter that should encourage a disentangled representation. 
In \citet{Burgess-et-al-2018} the authors provide further thoughts on the mechanism that leads to these disentangled representations.
However we will show in \secref{sec:betavae} that this parameter amplifies biases of variational inference towards orthogonalization and pruning.

\paragraph{$\beta$-TCVAE}{ }
In \citet{Chen-et-al-2018} the authors propose an alternative decomposition of the ELBO, that leads to the recent variant of $\beta$-VAE called $\beta$-TCVAE. They demonstrate that $\beta$-TCVAE allows to learn representations with higher MIG score than $\beta$-VAE~\citep{Higgins-et-al-2016}, InfoGAN~\citep{Chen-et-al-2016} and FactorVAE~\citep{Kim-and-Mnih-2018}.
The authors propose to decompose the KL-term in the ELBO objective into three parts and to weight them independently:

~\\[-16pt]
\small
\begin{align}
\nonumber &\mathbb{E}_{p_\theta(\vx)} \left[ D_{\textrm{KL}}(q_\phi(\vz|\vx) \| p(\vz)) \right] =\\[1em]
\nonumber & = D_{\textrm{KL}}(q_\phi(\vz|\vx) \| q_\phi(\vz)p_\theta(\vx)) + \\[1em]
\nonumber & + D_{\textrm{KL}}(q_\phi(\vz) \| \prod\limits_j q_\phi(\vz_j)) + \\
& + \sum\limits_j D_{\textrm{KL}}(q_\phi(\vz_j) \| p(\vz_j))
\, .
\label{eq:tcvae}
\end{align}
\normalsize
The first term is the index-code mutual information, the second term is the total correlation and the third term the dimension-wise KL-divergence.
Because the index-code mutual information can be viewed as an estimator for the mutual information between $p_\theta(\vx)$ and $q_\phi(\vz)$, the authors propose to exclude this term when reweighting the KL-term with the $\beta$ weight. 
In addition to the improved objective, the authors propose a quantitative metric of disentanglement, the mutual information gap (MIG). To compute this metric, first the mutual information between every latent variable and the underlying generative factors of the dataset are evaluated. The mutual information gap is then defined as the difference of the mutual information between the latent variables with highest and second highest correlation with an underlying factor.
\todo{change spacing between figures. prevent linebreak for beta=}
\begin{figure*}[h!]
\captionsetup[subfigure]{justification=centering}
\begin{center}
\begin{subfigure}[t]{0.3\linewidth}
	\centering
	\includegraphics[height=3.5cm]{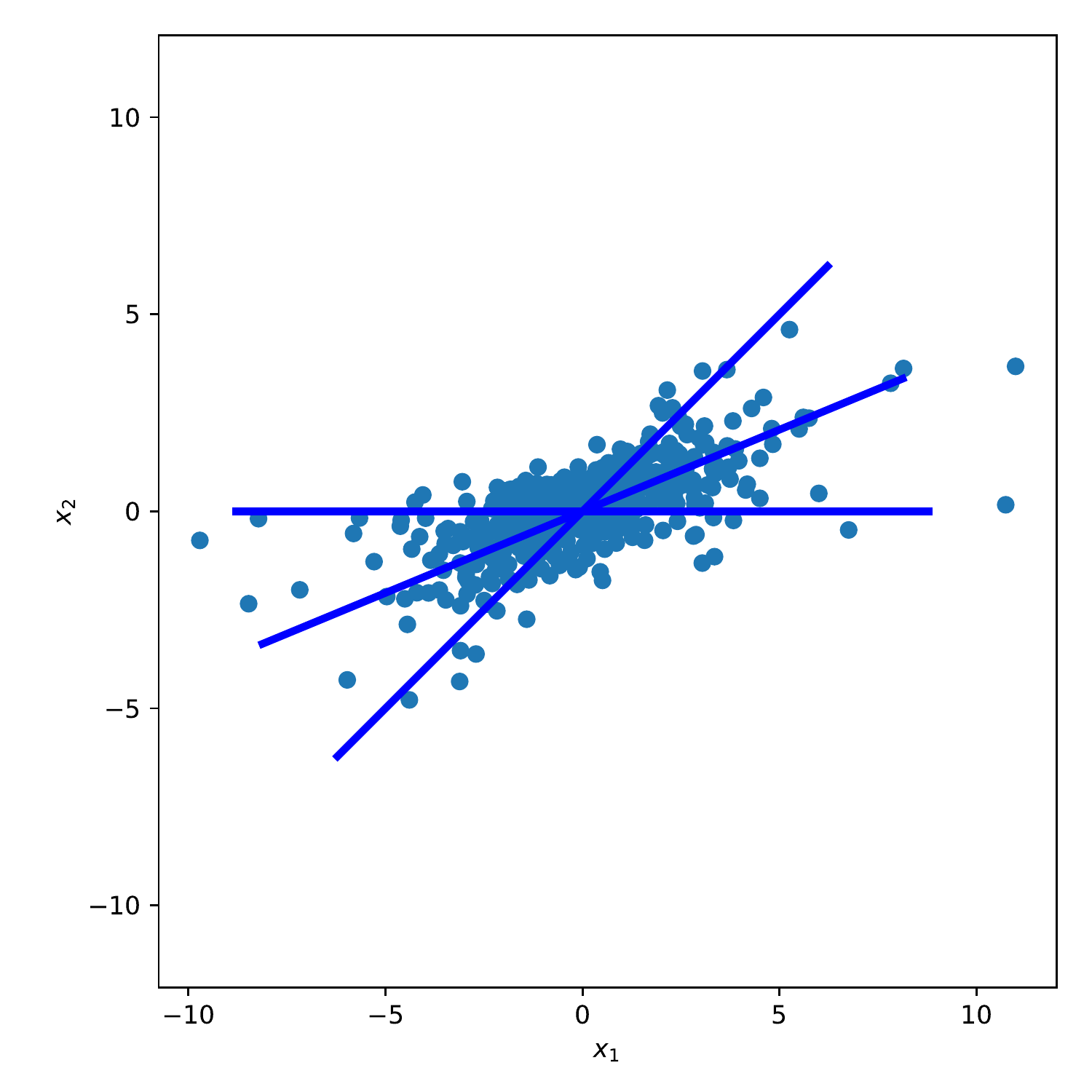}
	\caption{Sparse student's t-distribution generated by three latent components.}
\end{subfigure}
~~~
\begin{subfigure}[t]{0.3\linewidth}
	\centering
	\includegraphics[height=3.5cm]{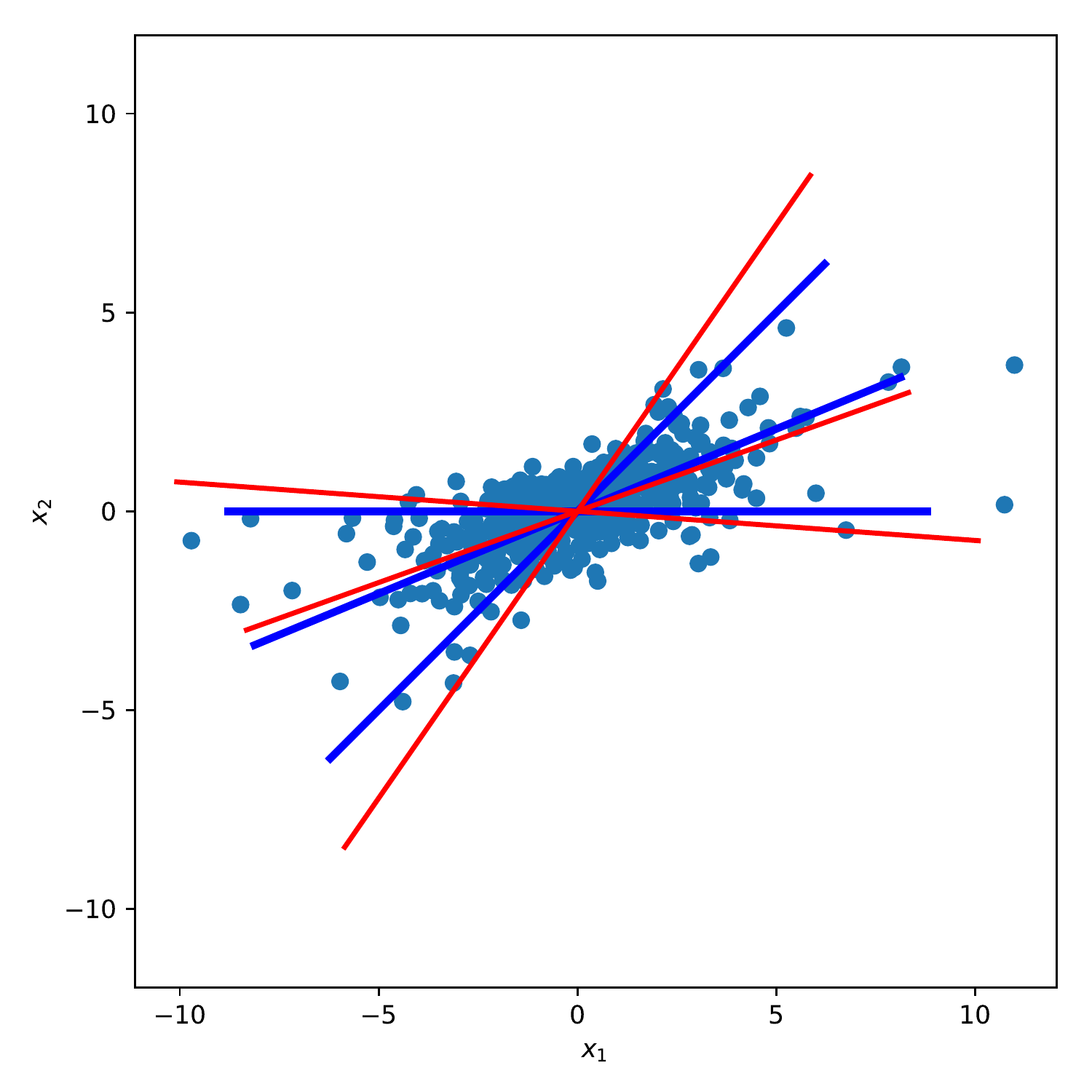}
	\caption{Variational inference\\ with \mbox{$\beta=1.0$}.}
\end{subfigure}
~~~
\begin{subfigure}[t]{0.3\linewidth}
	\centering
	\includegraphics[height=3.5cm]{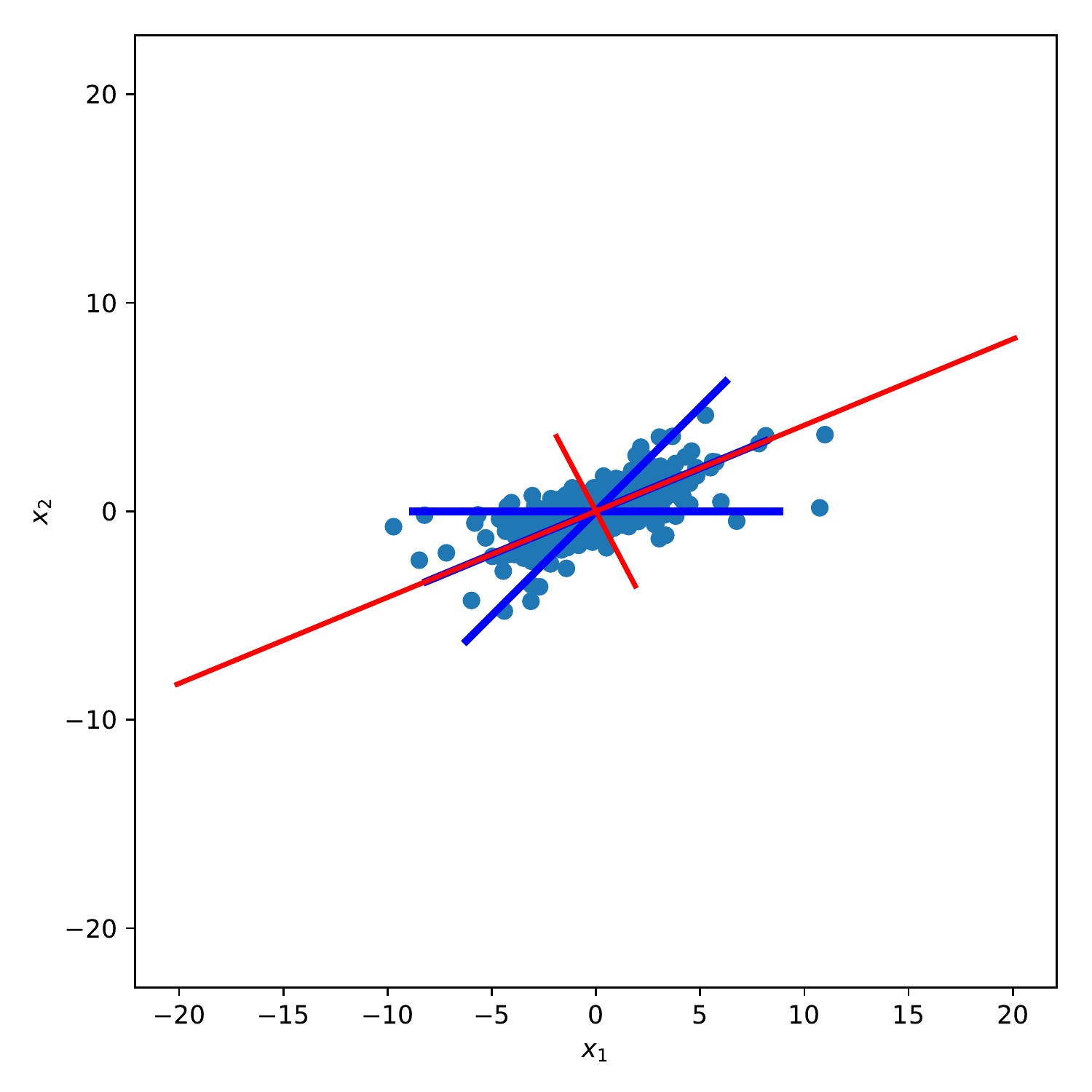}
	\caption{Variational inference\\ with \mbox{$\beta=5.0$}.}
\end{subfigure}
\end{center}
\caption{The modified ELBO objective of $\beta$-VAE emphasizes orthogonalization and pruning with increasing $\beta$-weight of the Kullback Leibler divergence term.}
\label{fig-ica-example}
\end{figure*}
\subsection{Related Work}

Recent work has shown an increased interest into learning of interpretable representations. In addition to the work mentioned already, we briefly review some of the influential papers:
\citet{Chen-et-al-2016} present a variant of a GAN that encourages an interpretable latent representation by maximizing the mutual information between the observation and a small subset of latent variables. The approach relies on optimizing a lower bound of the intractable mutual information.

\citet{Kim-and-Mnih-2018} propose a learning objective equivalent to $\beta$-TCVAE, and train it with the density ratio trick~\citep{Sugiyama-et-al-2012}. \citet{kumar2017variational} introduce a regulariser of the KL-divergence between the approximate posterior and the prior distribution. A parallel line of research proposes not to train a perfect generative model but instead to find a simpler representation of the data~\citep{vedantam2017generative,hinton2011transforming}.
A similar strategy is followed in semi-supervised approaches that require implicit or explicit knowledge about the true underlying factors of the data~\citep{Kulkarni-et-al-2015, Kingma-et-al-2014, Reed-et-al-2014,Siddharth-et-al-2017,Hinton-et-al-2011,Mathieu-et-al-2016,Goroshin-et-al-2015,Hsu-et-al-2017, Denton-and-Birodkar-2017}.

Existing work on structured priors for VAEs, the VAMP prior~\cite{tomczak2017vamp} and the LORACs prior~\cite{vikram2018loracs}, are modelling a clustered latent space. In our work however we introduce a prior which is a latent subspace model.

The recent work of~\citet{Locatello-et-al-2019} challenges the whole field of unsupervised representation learning and presents a proof of unidentifiability of the latent representation. We want to emphasize however, that the proof presented in~\citet{Locatello-et-al-2019} only holds for priors that factorize over every latent dimension. A property which does not hold for the prior proposed in this work.

\subsection{Orthogonalization and Pruning in Variational Inference}
\label{sec:ortho}

There have been several interpretations of the behaviour of the $\beta$-VAE \citep{Chen-et-al-2018,Burgess-et-al-2018,Rolinek-et-al-2019}. Here we provide a complementary perspective: that it enhances well known statistical biases in VI \citep{turner+sahani:2011a} to produce disentangled, but not necessarily useful, representations. The form of these biases can be understood by considering the variational objective when written as an explicit lower-bound: the log-likelihood of the parameters minus the KL divergence between the approximate posterior and the true posterior
\begin{align}
\mathcal{L}_{\textrm{ELBO}} =
\log p_\theta(\vx)
- D_{\textrm{KL}}(q_\phi(\vz|\vx) \| p_\theta(\vz | \vx )) \, .
          \label{eq1}
\end{align}
From this form it is clear that VI's estimates of the parameters $\theta$ will be biased away from the maximum likelihood solution (the maximizer of the first term) in a direction that reduces the KL between the approximate and true posteriors. When factorized approximating distributions are used, VI will therefore be biased towards settings of the parameters that reduce the statistical dependence between the latent variables in the posterior. For example, this will bias learned components towards orthogonal directions in the output space as this reduces explaining away (e.g. in the factor analysis model, VI breaks the degeneracy of the maximum-likelihood solution finding the orthogonal PCA directions, see appendix~\ref{sec:ica-example}). Moreover, these biases often cause components to be pruned out (in the sense that they have no effect on the observed variables) since then their posterior sits at the prior, which is typically factorized (e.g.~in an over-complete factor analysis model VI prunes out components to return a complete model, see appendix~\ref{sec:ica-example}). For simple linear models these effects are not pathological: indeed VI is arguably selecting from amongst the degenerate maximum likelihood solutions in a sensible way. However, for more complex models the biases are more severe: often the true posterior of the underlying model has significant dependencies (e.g. due to explaining away) and the biases can prevent the discovery of some components. For example, VAEs are known to over-prune~\citep{burda2015importance, cremer2018inference}.  

\subsection{$\beta$-VAE Emphasizes Orthogonalization and Pruning}
\label{sec:betavae}

What happens to these biases in the $\beta$-VAE generalization when $\beta>1$? 
%
The short answer is that they grow. This can be understood by considering coordinate ascent of the modified objective. With $\theta$ fixed, optimising $q$ finds a solution that is closer to the prior distribution than VI due to the upweighting of the KL term in \ref{eq:betavae}. With $q$ fixed, optimization over $\theta$ returns the same solution as VI (since the prior does not depend on the parameters $\theta$ and so the value of $\beta$ is irrelevant). However, since $q$ is now closer to the prior than before, the KL bias in \eqref{eq:betavae} will be greater. These effects are shown in the ICA example in \figref{fig-ica-example}. Also refer to appendix~\ref{sec:ica-example} for further details.
VI ($\beta=1$) learns components that are more orthogonal than the underlying ones, but $\beta=5$ prunes out one component entirely and sets the other two to be orthogonal. This is disentangled, but arguably leads to incorrect interpretation of the data. This happens even though both methods are initialised at the true model. Arguably, the $\beta$-VAE is enhancing a statistical bug in VI and leveraging this as a feature. We believe that this can be dangerous, preventing the discovery of the underlying model. 
\todo{ICA example}

\begin{figure}
\begin{center}
\scalebox{1.25}{
\setlength{\tabcolsep}{0em}
\begin{tabular}{c c c c}
&
\begin{minipage}[b]{2cm}
\centering
\scalebox{.5}{$p_1=1.0$}
\end{minipage}
&
\begin{minipage}[b]{2cm}
\centering
\scalebox{.5}{$p_1=2.0$}
\end{minipage}
&
\begin{minipage}[b]{2cm}
\centering
\scalebox{.5}{$p_1=9.0$}
\end{minipage}
\\
\vspace{-1.2cm}
\\
\begin{minipage}[c][2cm][b]{0.25cm}
\centering
\rotatebox{90}{\scalebox{.5}{$p_0=1.0$}}
\end{minipage}
&\multicolumn{3}{c}{\multirow{3}{6cm}{\begin{minipage}[c][6cm][b]{6cm}
\includegraphics[width=6cm]{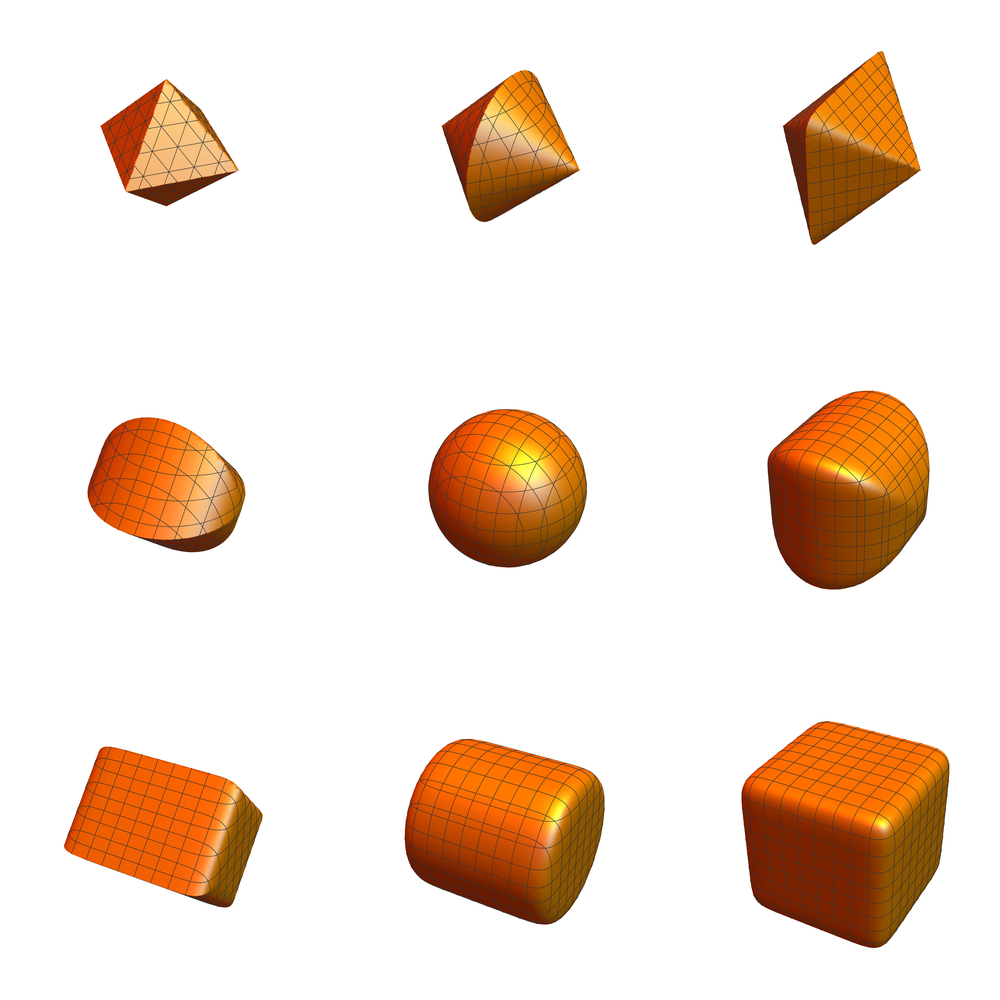}
\end{minipage}}}\\
\begin{minipage}[c][2cm][b]{0.25cm}
\centering
\rotatebox{90}{\scalebox{.5}{$p_0=2.0$}}
\end{minipage}&\\
\begin{minipage}[c][2cm][b]{0.25cm}
\centering
\rotatebox{90}{\scalebox{.5}{$p_0=9.0$}}
\end{minipage}&&&\\
\vspace{5pt}
\end{tabular}
}
\end{center}
\caption{Iso-contours of the $\normlp$-nested function example in \eqref{lpnested-example} for combinations of $p_0, p_1 \in \{1, 2, 9\}$.}
\label{fig:isocontours}
\end{figure}

\section{Latent Prior Distributions for Unsupervised Factorization}

In this section we describe an approach for unsupervised learning of disentangled representations. Instead of modifying the ELBO-objective, we propose to use certain families of prior distributions $p(\vz)$, that lead to identifiable and interpretable models. In contrast to the standard normal distribution, the proposed priors are not rotationally invariant, and therefore allow interpretability of the latent space.

\subsection{Independent Component Analysis}
\label{sec:ICA}

Independent Component Analysis
(ICA) seeks to factorize a distribution into non-Gaussian factors.
In order to avoid the ambiguities of latent space rotations, a non-Gaussian distribution (e.g. Laplace or Student-t distribution) is used as prior for the latent variables.\\

\begin{figure*}
\begin{center}
\begin{subfigure}[t]{0.45\linewidth}
	\centering
	\includegraphics[height=2cm]{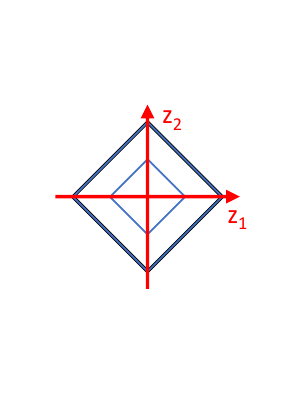}
	\includegraphics[height=2cm]{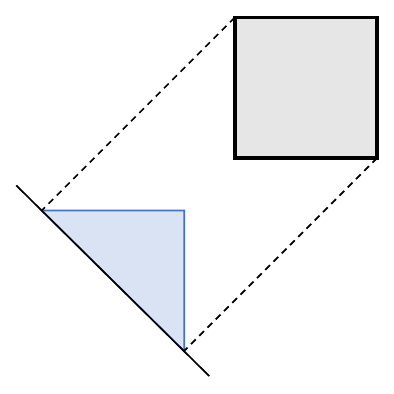}
	\includegraphics[height=2cm]{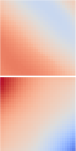}
	\caption{$p=1.0$}
	\label{xy-leptokurtic}
\end{subfigure}
~
\begin{subfigure}[t]{0.45\linewidth}
	\centering
	\includegraphics[height=2cm]{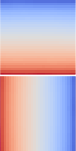}
	\includegraphics[height=2cm]{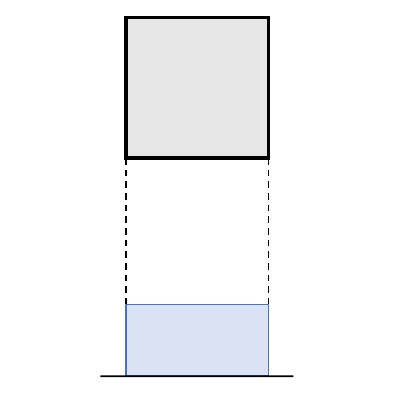}
	\includegraphics[height=2cm]{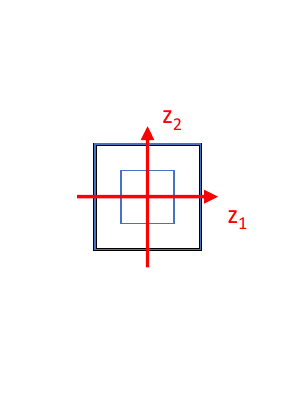}
	\caption{$p=2.4$}
	\label{xy-platykurtic}
\end{subfigure}
\end{center}
\caption{Leptokurtic and platykurtic priors encourage different orientations of the encoding of the (x,y) location of a sprite in the dSprites dataset.
A leptokurtic distribution (here the Laplace distribution) has, in two dimensions, contour lines along diagonal directions and expects most of the probability mass around $0$. Because the (x,y) locations in dSprites are distributed in a square, the projection of the coordinates onto the diagonal fits better to the Laplace prior. A platykurtic distribution however is more similar to a uniform distribution, with axis aligned contour lines in two dimensions. This fits better to an orthogonal projection of the (x,y) location. The red and blue colour coding denotes the value of the latent variable for the respective (x,y) location of a sprite.}
\label{fig:leptoplaty}
\end{figure*}

\paragraph{Generalized Gaussian Distribution} A generalized version of ICA~\citep{Lee-Lewicki-2000,Zhang-et-al-2004,Lewicki-2002,Sinz-Bethge-2010} uses a prior from the family of \emph{exponential power distributions} of the form
\begin{equation}
p_{\textrm{ICA}}(\vz) \propto \exp\left(-\tau{||\vz||}_p^p\right)
\end{equation}
also called \emph{generalized Gaussian}, \emph{generalized Laplacian} or \emph{$p$-generalized normal} distribution. 
Using $p = 2/(1+\kappa)$ the parameter $\kappa$ is a measure of kurtosis~\citep{Box-Tiao-1973}.
This family of distributions generalizes the normal ($\kappa=0$) and the Laplacian ($\kappa=1$) distribution.
In general we get for
 $\kappa>0$ \emph{leptokurtic} and for
 $\kappa<0$ \emph{platykurtic}
distributions.
The choice of a leptokurtic or platykurtic distribution has a strong influence on how a generative factor of the data is represented by a latent dimension. Fig.~\ref{fig:leptoplaty} depicts two possible prior distributions over latents that represent the (x,y) spatial location of a sprite in the dSprites dataset~\citep{dsprites17}. The leptokurtic distribution expects most of the probability mass around $0$ and therefore favours a projection of the x and y coordinates, which are distributed in a square, onto the diagonal. The platykurtic prior is closer to a uniform distribution and therefore encourages an axis-aligned representation. This example shows how the choice of the prior will effect the latent representation.

Obviously the normal distribution is a special instance of the class of $\normlp$-spherically symmetric distributions, and the normal distribution is the only $\normltwo$-spherically symmetric distribution with independent marginals.
Equivalently~\citep{Sinz-et-al-2009a} showed that this also generalizes to arbitrary values of $p$.
The marginals of the $p$-generalized normal distribution are independent, and it is the only factorial model in the class of $\normlp$-spherically symmetric distributions.


\subsection{Independent Subspace Analysis}

ICA can be further generalized to include independence between subspaces, but dependencies within them, by using a more general prior, the family of $\normlp$-nested symmetric distributions~\citep{Hyvarinen-Hoyer-2000,Hyvarinen-Koster-2007,Sinz-et-al-2009, Sinz-Bethge-2010}.

\begin{figure*}
\begin{subfigure}[t]{0.39\linewidth}
	\centering
	\includegraphics[height=2.5cm]{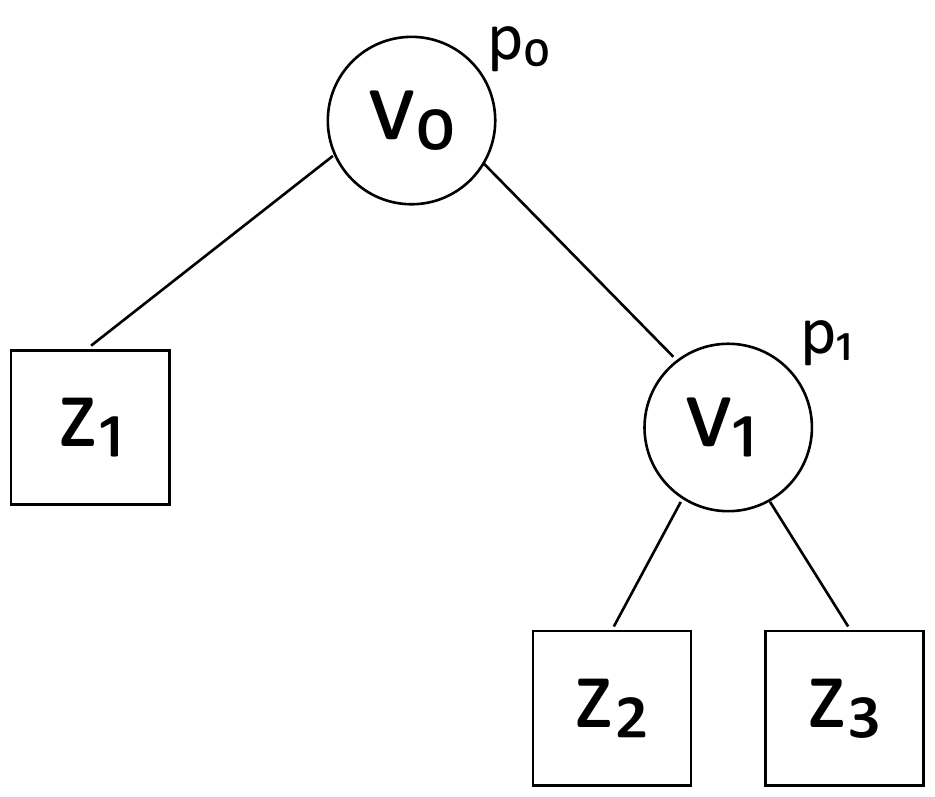}
	\caption{Tree corresponding to Eq. \ref{lpnested-example}}
	\label{fig:lpnestedtree}
\end{subfigure}
\begin{subfigure}[t]{0.59\linewidth}
	\centering
	\includegraphics[height=2.5cm]{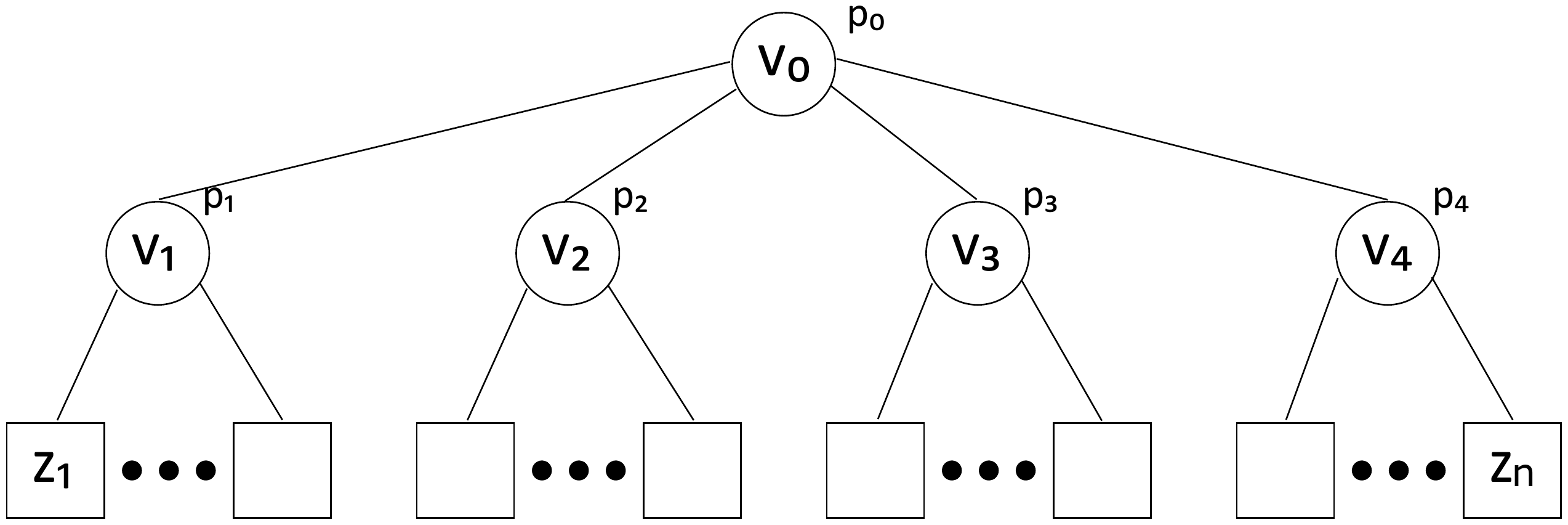}
	\caption{Tree visualization of Eq. \ref{lpnested-isa}, an $\normlp$-nested ISA model.}
	\label{fig:lpnested-ISA}
\end{subfigure}
\caption{Tree representation of $\normlp$-nested distributions. a) Tree of the example provided in Eq.~\ref{lpnested-example}. b) Tree corresponding to an $\normlp$-nested ISA model.}
\end{figure*}

\paragraph{$\normlp$-nested Function} To start, we take a look at a simple example of an $\normlp$-nested function:
\begin{equation}
\left(|z_1|^{p_0} + \left( |z_2|^{p_1} + |z_3|^{p_1} \right)^{\frac{p_0}{p_1}} \right)^{\frac{1}{p_0}} \, ,
\label{lpnested-example}
\end{equation}
with $p_0, p_1 \in \displaystyle \R$. This function is a cascade of two $\normlp$-norms. To aid intuition we provide a visualization of this distribution in \figref{fig:lpnestedtree}, which depicts (\ref{lpnested-example}) as a tree
that visualizes the nested structure of the norms. \figref{fig:isocontours} visualizes the iso-contours of this function for different values of $p_0$ and $p_1$. We call the class of functions which employ this structure \emph{$\normlp$-nested}.

\paragraph{$\normlp$-nested Distribution} Given an $\normlp$-nested function $f$ and a radial density $\psi_0 : \R \mapsto \R^+$ we define the 
\emph{$\normlp$-nested symmetric distribution} following~\citet{Fernandez-1995} as
\begin{align}
p_\textrm{ISA}(\vz) = \frac{\psi_0(f(\vz))}{f(\vz)^{n-1} \mathcal{S}_f(1)} \, ,
\label{eq:lpnesteddist}
\end{align}
where $\mathcal{S}_f(1)$ is the surface area of the $\normlp$-nested unit-sphere. This surface area can be obtained by using the gamma function:
\begin{align}
\mathcal{S}_f(R) = R^{n-1} 2^n \prod\limits_{i \in I} \frac{\prod_{k=1}^{l_1} \Gamma\left[\frac{n_{i,k}}{p_i}\right]}{p_i^{l_i-1}\Gamma\left[\frac{n_{i}}{p_i}\right]} \, ,
\label{surface}
\end{align}
where $l_i$ is the number of children of a node $i$, $n_i$ is the number of leaves in a subtree under the node $i$, and $n_{i,k}$ is the number of leaves in the subtree of the $k$-th children of node $i$.
For further details we refer the reader to the excellent work of~\citet{Sinz-Bethge-2010}.

\paragraph{Independent Subspace Analysis} The family of $\normlp$-nested distributions allows a generalization of ICA called independent subspace analysis (ISA). 
ISA uses a subclass of $\normlp$-nested distributions, which are defined by functions of the form

~\\[-16pt]
\small
\begin{align}
\nonumber
f(\vz) = & \left(\left(\sum\limits_{j=1}^{n_1}|z_j|^{p_1}\right)^{\frac{p_0}{p_1}} + \dots\right. \\
& \left. \dots + \left(\sum\limits_{j=n_1+\dots+n_{l-1}+1}^{n}|z_j|^{p_l}\right)^{\frac{p_0}{p_l}} \right)^{\frac{1}{p_0}} \, ,
\label{lpnested-isa}
\end{align}
\normalsize
and correspond to trees of depth two. The tree structure of this subclass of functions is visualized in \figref{fig:lpnested-ISA} where each $v_i, \, i=1,\dots,l_0$ denotes the function value of the $\normlp$-norm evaluated over a node's children. The components $z_j$ of $z$ that contribute to each  $v_i$ form a subspace

~\\[-16pt]
\small
\begin{align}
\mathcal{V}_i = \left\{ z_j \, \Big| \, j = a \dots b \,\, \text{with} \,\, a=\sum\limits_{k=1}^{i-1} n_k + 1, \, b = a + n_i \right\} \, .
\end{align}
\normalsize
The subspaces $\mathcal{V}_1, \dots, \mathcal{V}_{l_0}$ become independent
when using the radial distribution~\citet{Sinz-Bethge-2010}
\begin{align}
\psi_0(v_0) = \frac{p_0v_0^{n-1}}{\Gamma\left[\frac{n}{p_0}\right]s^{\frac{n}{p_0}}} \exp\left(-\frac{v_0^{p_0}}{s}\right) \, .
\end{align}
We can interpret this as a generalization of the Chi-distribution: it is the radial distribution of an $\normlp$-nested distribution that becomes equivalent to the Chi-distribution in the case of an $\normltwo$-spherically symmetric (Gaussian) distribution.

\paragraph{ISA-VAE}
We propose to choose the latent prior $p_\textrm{ISA}(\vz)$ (Eq.~\ref{eq:lpnesteddist}) with $f(\vz)$ from the family of ISA models of the form of Eq.~\ref{lpnested-isa}, which allows us to define independent subspaces in the latent space.\footnote{Independend of this work, \citet{Higgins-et-al-2018} recently proposed to use independent vector subspaces as latent representations to define a new notion of disentanglement.}
The Kulback-Leibler divergence of the ELBO-objective can be estimated by Monte-Carlo sampling. This leads to an ELBO-objective of the form
\begin{align}
\nonumber \mathcal{L}_{\textrm{ISA-VAE}} =
& \mathbb{E}_{z \sim q_\phi(\vz|\vx)} \left[ \log p_\theta(\vx|\vz) - \right. \\
& \left. - \beta \left(\log q_\phi\left(\vz|\vx\right) -  \log p_\textrm{ISA}\left(\vz\right) \right) \right]
\, ,
\end{align}
which only requires to compute the log-density of the prior that is readily accessible from the density defined in Eq.~\ref{eq:lpnesteddist}. As discussed in \cite{Roeder-et-al-2017} this form of the ELBO even has potential advantages (variance reduction) in comparison to a closed form KL-divergence.


\paragraph{Sampling and the Reparameterization Trick}
If we want to sample from the generative model we have to be able to sample from the prior distribution. \citet{Sinz-Bethge-2010} describe an exact sampling approach to sample from an $\normlp$-nested distribution, which we reproduce as Algorithm~\ref{alg:sampling} in the appendix.
Note that during training we only have to sample from the approximate posterior $q_\phi$, which we do not have to modify and which can remain a multivariate Gaussian distribution following the original VAE approach. As a consequence, the reparameterization trick can be applied~\citep{Kingma-Welling-2014}.

%

Experiments in the following section demonstrate that the proposed prior supports unsupervised learning of disentangled representation even for the unmodified ELBO objective ($\beta=1$).


\section{Experiments}
\label{sec:experiments}


\begin{figure*}
\begin{center}
\begin{subfigure}{0.175\linewidth}
\vspace{-0.4cm}
\includegraphics[width=0.97\linewidth]{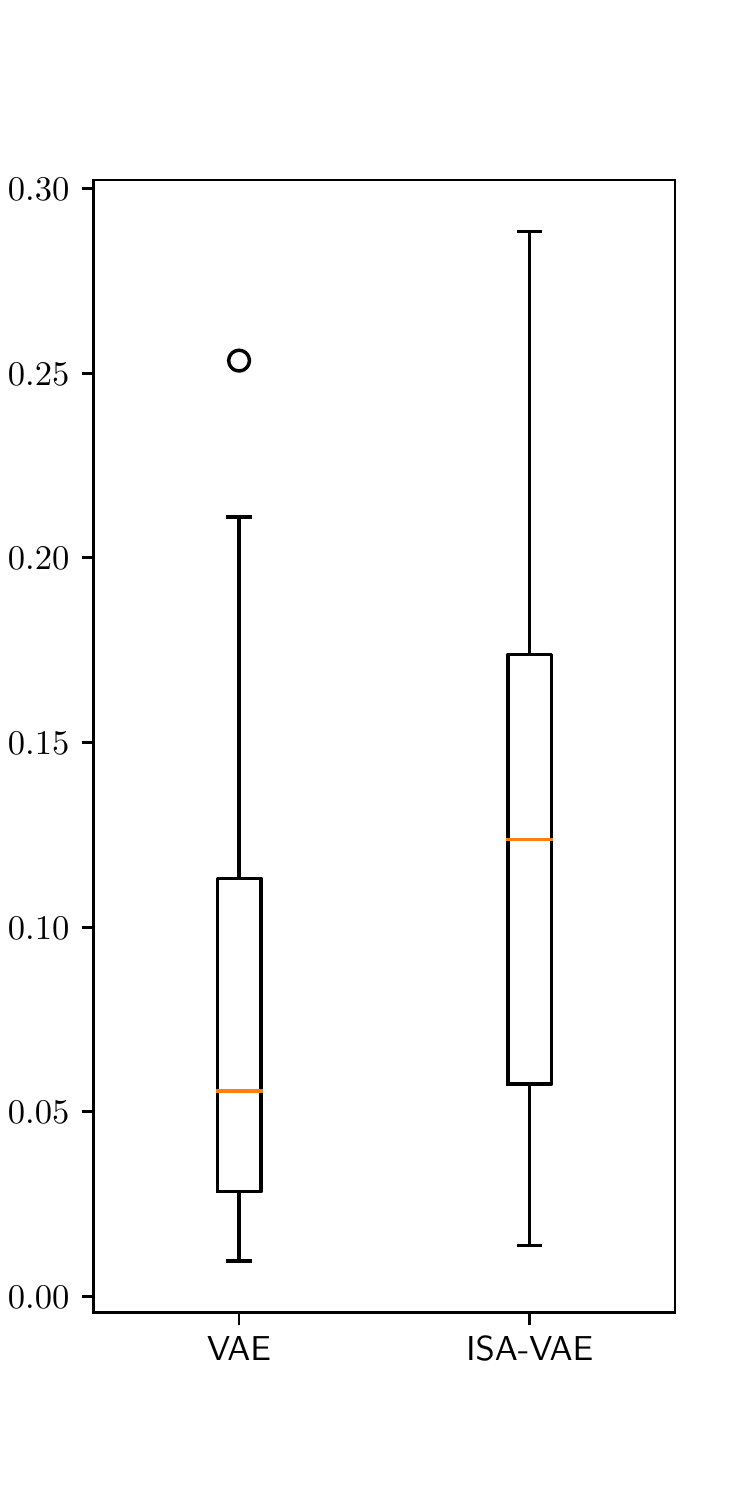}
\vspace{-0.38cm}
\caption{dSprites, $\beta=1.0$}
\label{fig:beta1-boxplot}
\end{subfigure}
\begin{subfigure}{0.4\linewidth}
\centering
\includegraphics[width=\linewidth]{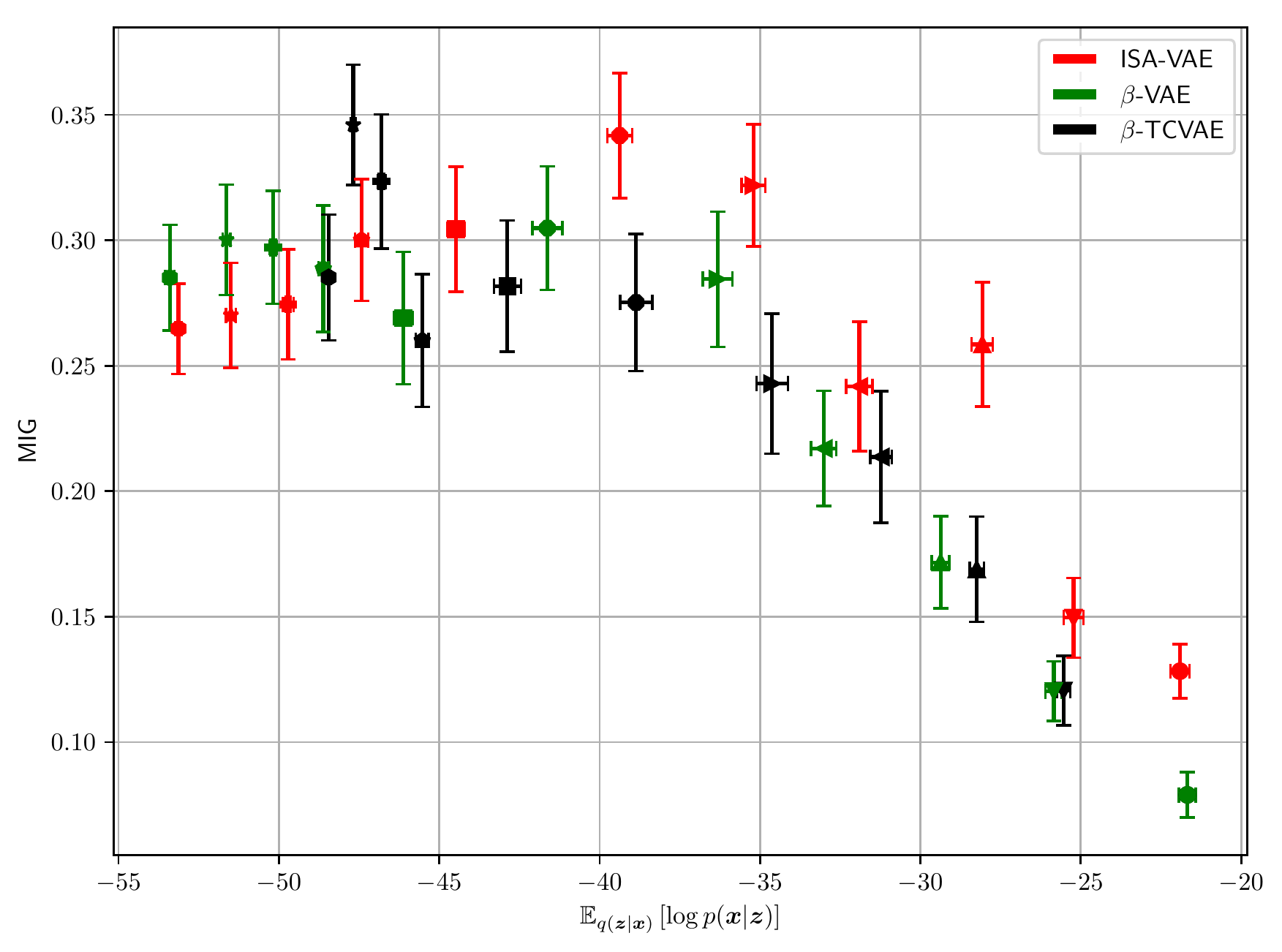}
\caption{dSprites, $\beta=1. 0, 1.5, \dots, 6.0$}
\label{fig:migvsloss-dsprites}
\end{subfigure}
\begin{subfigure}{0.4\linewidth}
\centering
\includegraphics[width=\linewidth]{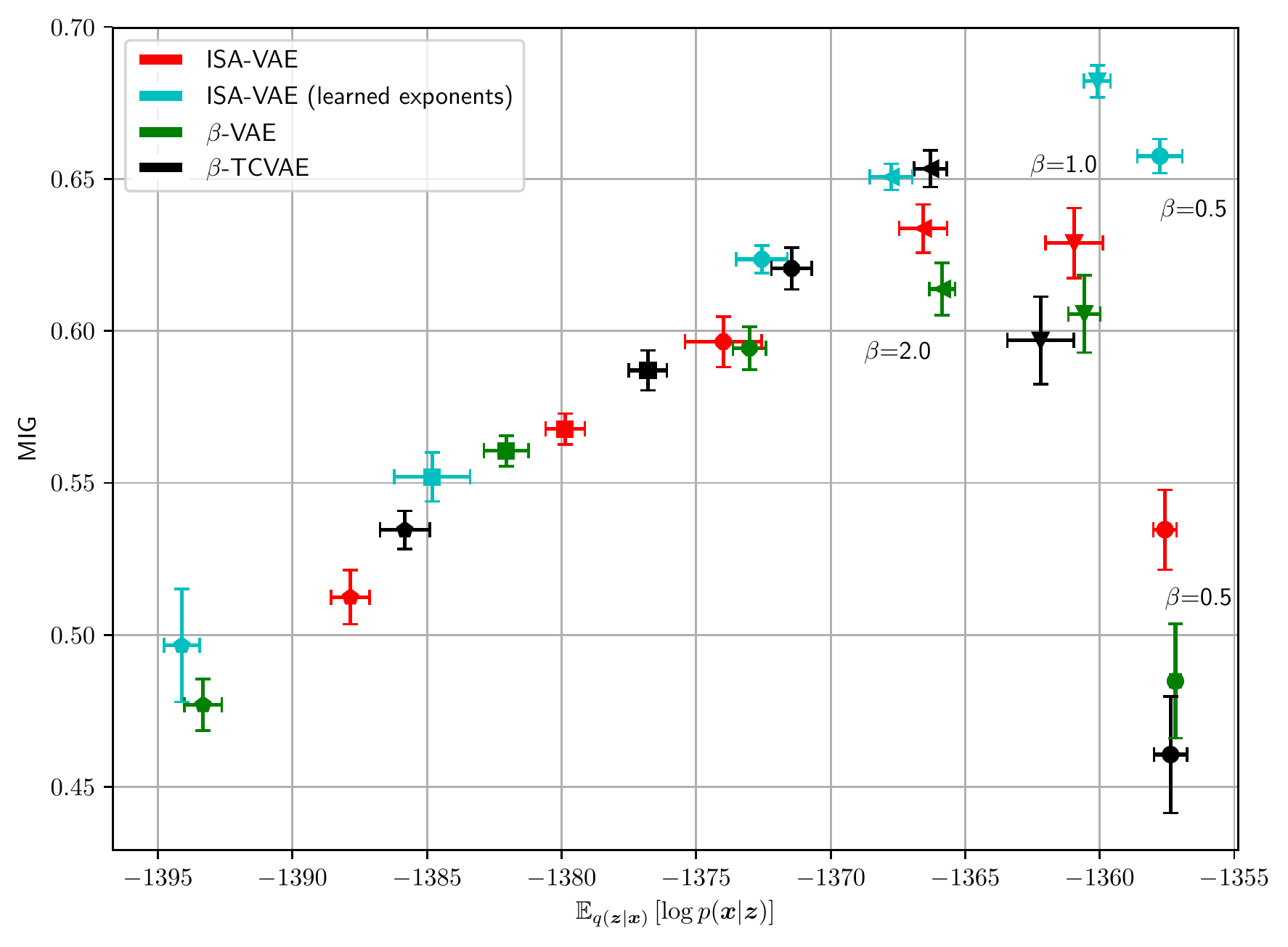}
\caption{3D faces, $\beta=0.5, 1.0, 2.0, 3.0, 4.0, 5.0$}
\label{fig:migvsloss-faces}
\end{subfigure}
\end{center}
\caption{Comparison of the different approaches for different values of $\beta$.
(a)~The proposed prior facilitates disentanglement, as demonstrated when comparing it to the standard normal prior when using the unmodified ELBO ($\beta=1.0$).
(b),~(c)~Scatter plots of MIG-score (higher score is better) and reconstruction quality (larger values to the right are better)/ Results for lower $\beta$ and better reconstruction quality are on the right. Error bars denote the standard error. The proposed approach ISA-VAE allows a better trade-off between disentanglement and reconstruction quality, often outperforming $\beta$-VAE and $\beta$-TCVAE with respect to  MIG score and reconstruction loss on both datasets. (b) On dSprites the baselines reach better MIG scores only for models with poor reconstruction quality. 
(c) Learning the exponents allows to improve the trade-off between disentanglement and reconstruction loss even further, clearly outperforming the baselines for the unmodefied objective $\beta=1.0$ and for $\beta=0.5$.
Layout of the ISA model: $l_0=5$, $l_{1,\dots,5}=4$.}
\label{fig:mig-logpx-comparison}
\end{figure*}

In our experiments, we evaluate the influence of the proposed prior distribution on disentanglement and on the quality of the reconstruction on the dSprites dataset~\citep{dsprites17}, which contains images of three different shapes undergoing transformations of their position (32 different positions in x and 32 different positions in y), scale (6) and rotation (40), and on the dataset 3D Faces~\citep{3dfaces} that was also used for evaluation in~\cite{Chen-et-al-2016}, which consists of synthetic images of faces with the latent factors azimut (21), elevation (11) and lighting (11). Further we present results on the cars3d dataset~\cite{cars3d}. We follow the same procedure as in~\citet{Locatello-et-al-2019} and perform an extensive evaluation with 50 experiments for each parameter setting, resulting in a total number of 1280 experiments. We will make the source code to repoduce our experiments available online.

\paragraph{Disentanglement Metrics}
To provide a quantitative evaluation of disentanglement we compute the disentanglement metric \emph{Mutual Information Gap} (MIG) that was proposed in~\citet{Chen-et-al-2018}.
The MIG score measures how much mutual information a latent dimension shares with the underlying factor, and how well this latent dimension is separated from the other latent factors.
Therefore the MIG measures the two desired properties usually referred to with the term \emph{disentanglement}: a factorized latent representation, and interpretability of the latent factors. \citet{Chen-et-al-2018} compare the MIG metric to existing disentanglement metrics \citep{Higgins-et-al-2016,Kim-and-Mnih-2018} and demonstrate that the MIG is more effective and that other metrics do not allow to capture both properties in a desirable way.

\paragraph{Reconstruction Quality}
To quantify the reconstruction quality, we report the expected \mbox{log-}likelihood of the reconstructed data $\mathbb{E}_{q_\phi(\vz|\vx)} \left[ \log p_\theta(\vx|\vz) \right]$. In our opinion this measure is more informative than the ELBO, frequently reported in existing work, e.g.~\citep{Chen-et-al-2018}, especially when varying the $\beta$ parameter, the weighting of the KL term, which is part of the ELBO and therefore affects its value.

\paragraph{Comparison Baselines}
\citet{Chen-et-al-2018} demonstrate that $\beta$-TCVAE, a modification of the $\beta$-VAE, enables learning of representations with higher disentanglement score than $\beta$-VAE~\citep{Higgins-et-al-2016}, InfoGAN~\citep{Chen-et-al-2016}, and FactorVAE~\citep{Kim-and-Mnih-2018}.
Therefore we choose $\beta$-TCVAE as a baseline for comparison and also compare against
$\beta$-VAE~\citep{Higgins-et-al-2016} which for $\beta=1$ includes the standard VAE with normal prior. To allow a quantitiative comparison with exisiting work we evaluate on the datasets dSprites~\cite{dsprites17} and 3D Faces~\citep{3dfaces} that were already used in~\citep{Chen-et-al-2016}.

\paragraph{Architecture of the Encoder and Decoder}
To allow a quantitative comparison with existing work and reproducible results we use the same architecture for the decoder and encoder as presented in~\citet{Chen-et-al-2018}. We reproduce the description of the encoder and decoder in appendix~\ref{sec:architecture}

\paragraph{Choosing the ISA-layout}
In our experience the layout only needs to provide sufficient independent vector spaces for learning the representations. If more than the required latent dimensions are provided, unused latent dimensions are usually pruned away.

\paragraph{Choosing the Exponents}
As we previously discussed in \secref{sec:ICA} it is important if the prior is leptokurtic or platykurtic. For $p_0$ we chose the value $p_0 = 2.1$, which results in a platykurtic prior for the distribution over the subspaces, which leads to a rotationally invariant prior. For simplicity, we choose the same exponent $p_1$ for all the subspaces. For dSprites, a platykurtic distribution fits best to the desired orientation of the $x$- and $y$-coordinate (Compare to Fig.~\ref{fig:leptoplaty}). To allow a factorized distribution the exponent has to be different to $p_0$, thus we chose $p_1=2.2$. On the 3D faces dataset a leptokyrtic distribution with $p_1=1.9$ provided better results than a platykurtic distribution.
\todo{(A comparison is provided in Fig.~\ref{fig:faces-p1-compare} in the appendix).}

\begin{figure}
\begin{center}
\begin{subfigure}[b]{\linewidth}
\centering
\includegraphics[width=\linewidth]{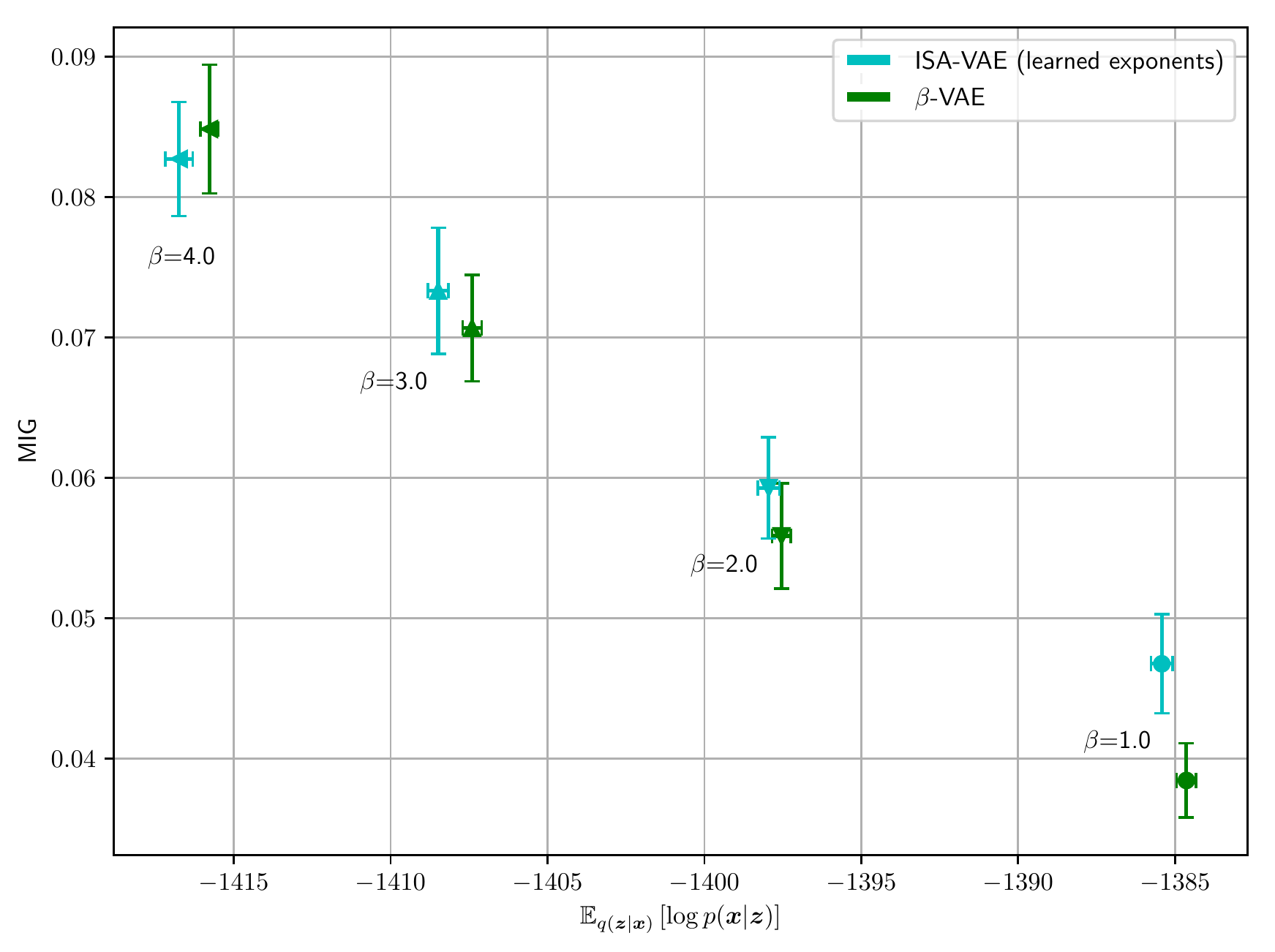}
\caption{cars3d, $\beta=1.0, 2.0, 3.0, 4.0$}
\end{subfigure}
\end{center}
\caption{Comparison on the cars3d dataset. Scatter plot of MIG-score (higher score is better) and reconstruction quality (larger values to the right are better). Error bars denote the standard error.
For the unmodified ELBO (beta=1, rightmost data points), ISA-VAE with learned exponents outperforms $\beta$-VAE with respect to the MIG score. These results confirm that the prior facilitates disentangled representations. This advantage however decreases for larger beta values.
}
\label{fig:mig-logpx-comparison-cars3d}
\end{figure}

\paragraph{Learning the Exponents}
Instead of choosing a fixed set of exponents, the exponents can also be learned during training. We use the modified ELBO as objective function and optimize encoder, decoder, and the exponents of the prior at the same time. We keep $p_0 = 2.1$ fixed and, beginning from $p_{1,\dots,k}=2.0$ optimize the exponent of each subspace individually. Interestingly, optimizing the exponents during training allows to improve the trade-off between disentanglement and reconstruction loss even further. We report results on learning the exponents on the 3D faces dataset in Fig.~\ref{fig:migvsloss-faces} (15 experiments per beta value) and on the cars3d dataset~\cite{cars3d} in Fig.~\ref{fig:mig-logpx-comparison-cars3d} (35 experiments per beta value). Histograms of the learnt exponents are shown in appendix~\ref{sec:learnedexponents}.

\paragraph{Hyperparameters}
To allow reproducibility and a comparison of our results we chose the same hyperparameters as in \citet{Chen-et-al-2018} and \citet{Locatello-et-al-2019}. We present a table with the evaluated hyperparameters in appendix~\ref{sec:hyperparams}.

\subsection{Support of the Prior to Learn Disentangled Representations}

First, we investigate the ability of the prior to support unsupervised learning of disentangled representations for the unmodified ELBO-objective ($\beta=1$) and compare the distribution of MIG scores that can be reached with ISA-VAE and the standard VAE in Fig.~\ref{fig:beta1-boxplot}. We perform $n=50$ experiments each.
We observe a higher mean, median and maximum quantile of disentanglement scores for ISA-VAE which indicates that the prior facilitates to learn interpretable representations even when using the unmodified ELBO objective with $\beta=1$.


\subsection{Trade-off between Disentanglement and Reconstruction Loss}
\todo{rewrite for n=50 and scatter plot}

Since the proposed prior facilitates learning of disentangled representations, not only a higher disentanglement score can be reached, but also higher scores are reached for smaller values of $\beta$, when compared to the original approaches. This leads to a clear improvement of the trade-off between disentanglement and reconstruction loss. The improvement of this trade-off is demonstrated in~Fig.~\ref{fig:mig-logpx-comparison}, where we plot both the disentanglement score and the reconstruction loss for varying values of $\beta$. ISA-$\beta$-VAE  reaches high values of the disentanglement score for smaller values of $\beta$ which at the same time preserves a higher quality of the reconstruction than the respective original approaches. 


The results on both datasets show that the increase of the MIG score for the baseline method $\beta$-TCVAE comes at the cost of a lower reconstruction quality. This difference in the reconstruction quality becomes visible in the quality of the reconstructed images, especially for the more complex heart shape. Please refer to the appendix where we present latent traversals in appendix~\ref{sec:traversals}.
With the proposed approach ISA-VAE the reconstruction quality can be increased while at the same time providing a higher disentanglement.
This trade-off can be even further improved when learning the exponents $p_{1,\dots,k}$ of the prior during training. Results for ISA-VAE with learned exponents on the 3d faces dataset are depicted in Fig. \ref{fig:migvsloss-faces} and on the cars 3d dataset in Fig. \ref{fig:mig-logpx-comparison-cars3d}. 

When learning the exponents, we observe the highest difference of disentanglement scores on the cars 3d dataset for low values of $\beta$. On the 3d faces cataset, the highest MIG scores among all approaches are reached with ISA-VAE with learned exponents for the unmodified ELBO objective ($\beta=1.0$), outperforming the existing approaches by a large margin.
Both results strongly support our hypothesis, that the proposed prior facilitates learning of disentangled representations even for the unmodified ELBO objective.

\section{Conclusion}

We presented a structured prior for unsupervised learning of disentangled representations in deep generative models.
We choose the prior from the family of \emph{$\normlp$-nested symmetric distributions} which enables the definition of a hierarchy of independent subspaces in the latent space. In contrast to the standard normal prior that is often used in training of deep generative models the proposed prior is not rotationally invariant and therefore enhances the interpretability of the latent space.
We demonstrate in our experiments, that a combination of the proposed prior with existing approaches for unsupervised learning of disentangled representations allows a significant improvement of the trade-off between disentanglement and reconstruction loss.
This trade-off can be improved further by learning the parameters of the prior during training.

\bibliographystyle{ieeetr}

\clearpage
\appendix
\section{Appendix}

\subsection{Sampling from $\normlp$-nested Symmetric Distributions}
We reproduce the sampling algorithm for $\normlp$-nested symmetric distributions from \citet{Sinz-Bethge-2010} in Alg.~\ref{alg:sampling}.
\begin{algorithm}[h!]
\SetAlgoLined
\label{alg:sample}
    \SetKwInOut{Input}{Input}
    \SetKwInOut{Output}{Output}

\caption{Exact sampling algorithm for $\normlp$-nested symmetric distributions\\ from \citet{Sinz-Bethge-2010}}\Input{The radial distribution $\psi_0(v_0)$ of an $\normlp$-nested symmetric distribution $p_{\normlp}$ for the $\normlp$-nested function $f$}
\Output{Sample $x$ from $p_{\normlp}$}
\begin{enumerate}
\item Sample $v_0$ from a beta distribution ${\beta}[n,1]$

\item For each inner node $i$ of the tree associated with $f$, sample the auxiliary variable $s_i$ from a Dirichlet distribution $\operatorname{Dir}\left[\frac{n_{i,1}}{p_i},\dots,\frac{n_{i,l_1}}{p_i}\right]$ where $n_{i,k}$ are the number of leaves in the subtree under node $i,k$. Obtain coordinates on the $\normlp$-nested sphere within the positive orthant by $\vs_i \mapsto \vs_i^{\frac{1}{p_i}} = \tilde{\vu}_i$ (the exponentiation is taken component-wise)

\item Transform these samples to Cartesian coordinates by
$v_i \cdot \tilde{\vu}_i = \vv_{i,1:l_i}$
for each inner node, starting from the root node and descending to lower layers.  The components of
$\vv_{i,1:l_i}$
constitute the radii for the layer direct below them. If
$i=0$, the radius had been sampled in step 1

\item Once the two previous steps have been repeated until no inner node is left, we have a sample $\vx$
from the uniform distribution in the positive quadrant. Normalize $\vx$
to get a uniform sample from the sphere $\vu = \frac{\vx}{f(\vx)}$

\item  Sample a new radius $\tilde{v}_0$
from the radial distribution of the target radial distribution
$\psi_0$
and obtain the sample via $\tilde{\vx} = \tilde{v}_0 \cdot \vu$

\item Multiply each entry $x_i$ of $\tilde{\vx}$ by and independent sample $z_i$ from the uniform distribution over $\{ -1, 1\}$.
\end{enumerate}
\label{alg:sampling}
\end{algorithm}


\subsection{Learned Exponents}
\label{sec:learnedexponents}

\begin{figure*}
\begin{center}
\begin{subfigure}[t]{0.3\linewidth}
	\centering
	\includegraphics[width=\linewidth]{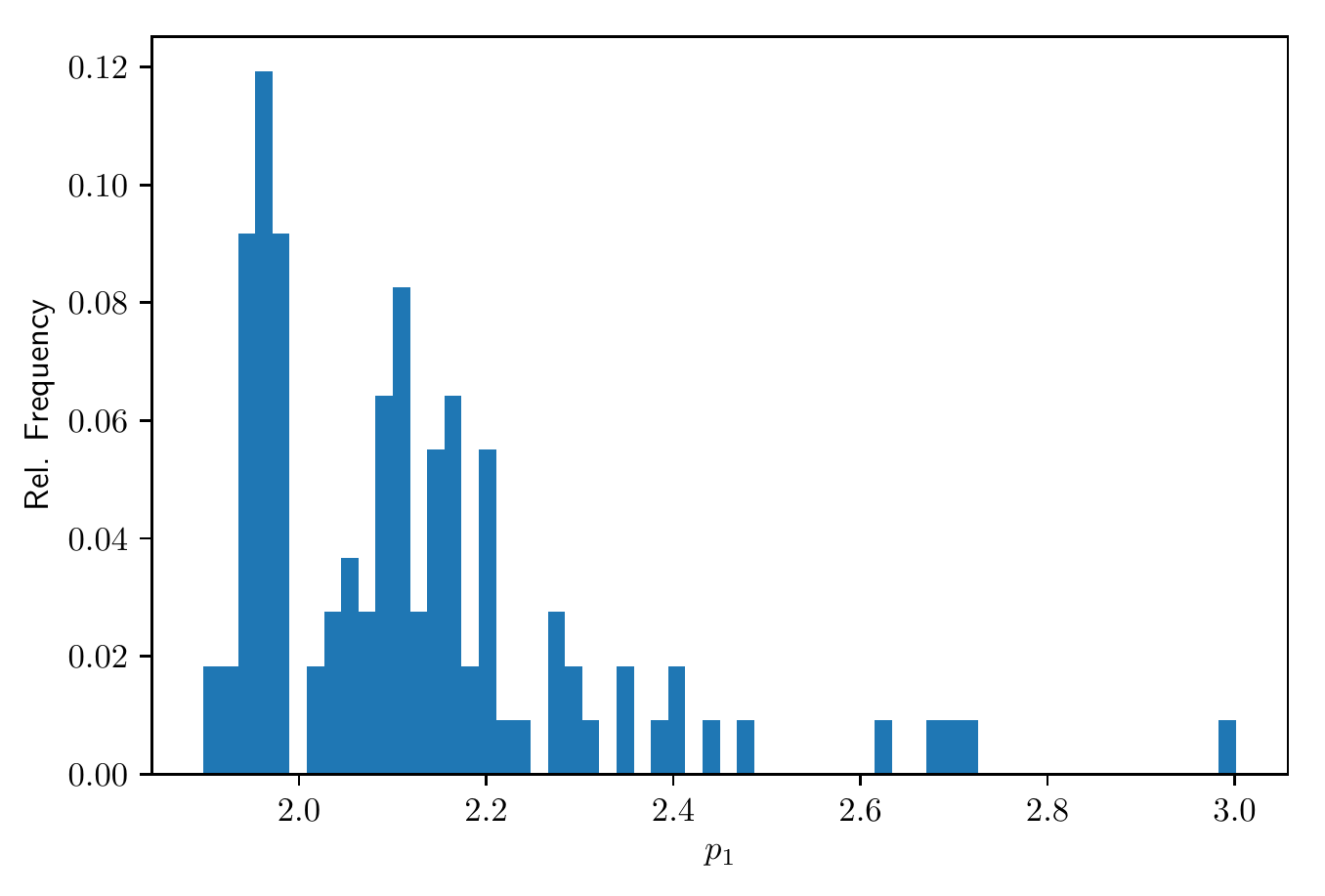}
	\caption{Histogram of $p_1$.}
\end{subfigure}
\begin{subfigure}[t]{0.3\linewidth}
	\centering
	\includegraphics[width=\linewidth]{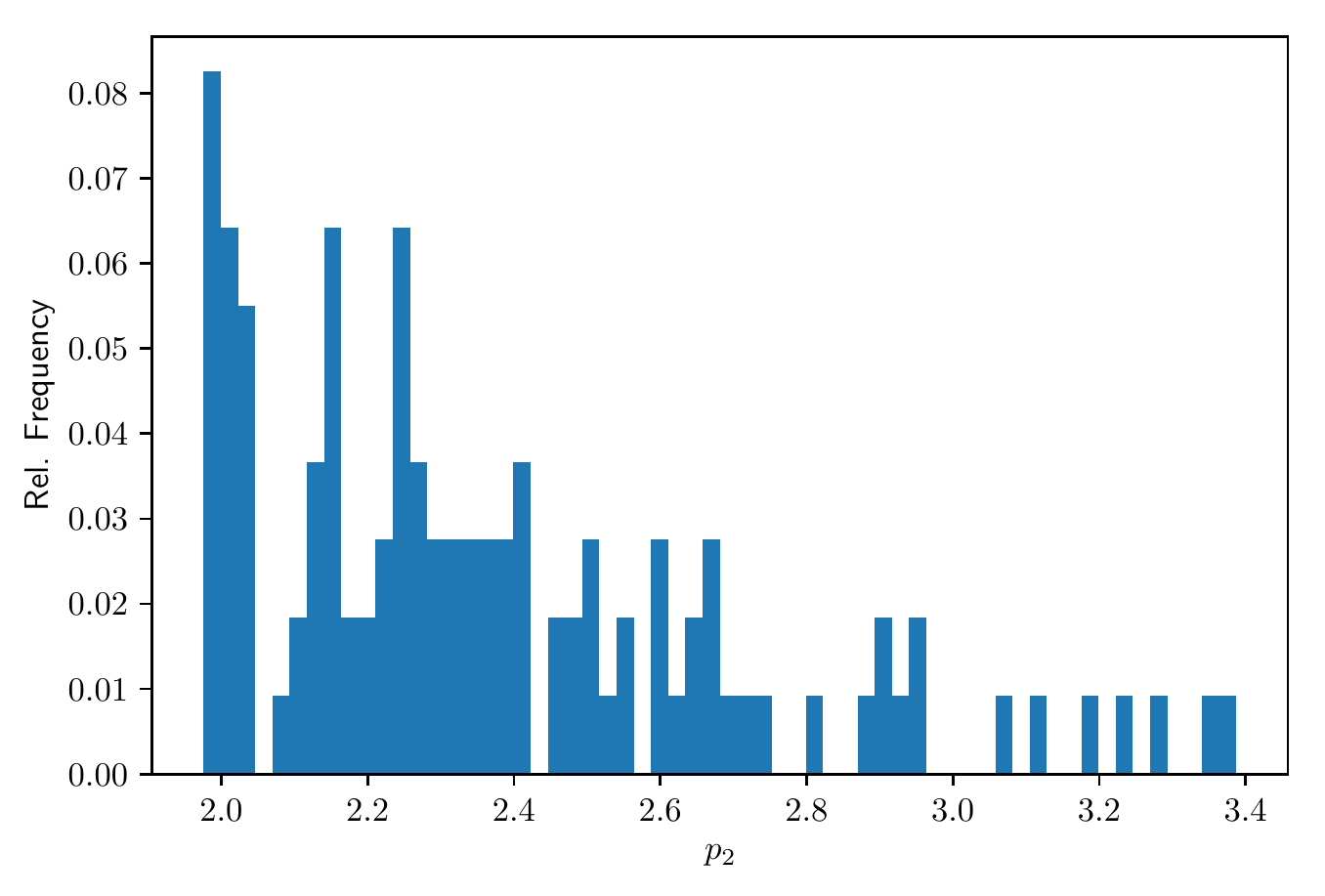}
	\caption{Histogram of $p_2$.}
\end{subfigure}
\begin{subfigure}[t]{0.3\linewidth}
	\centering
	\includegraphics[width=\linewidth]{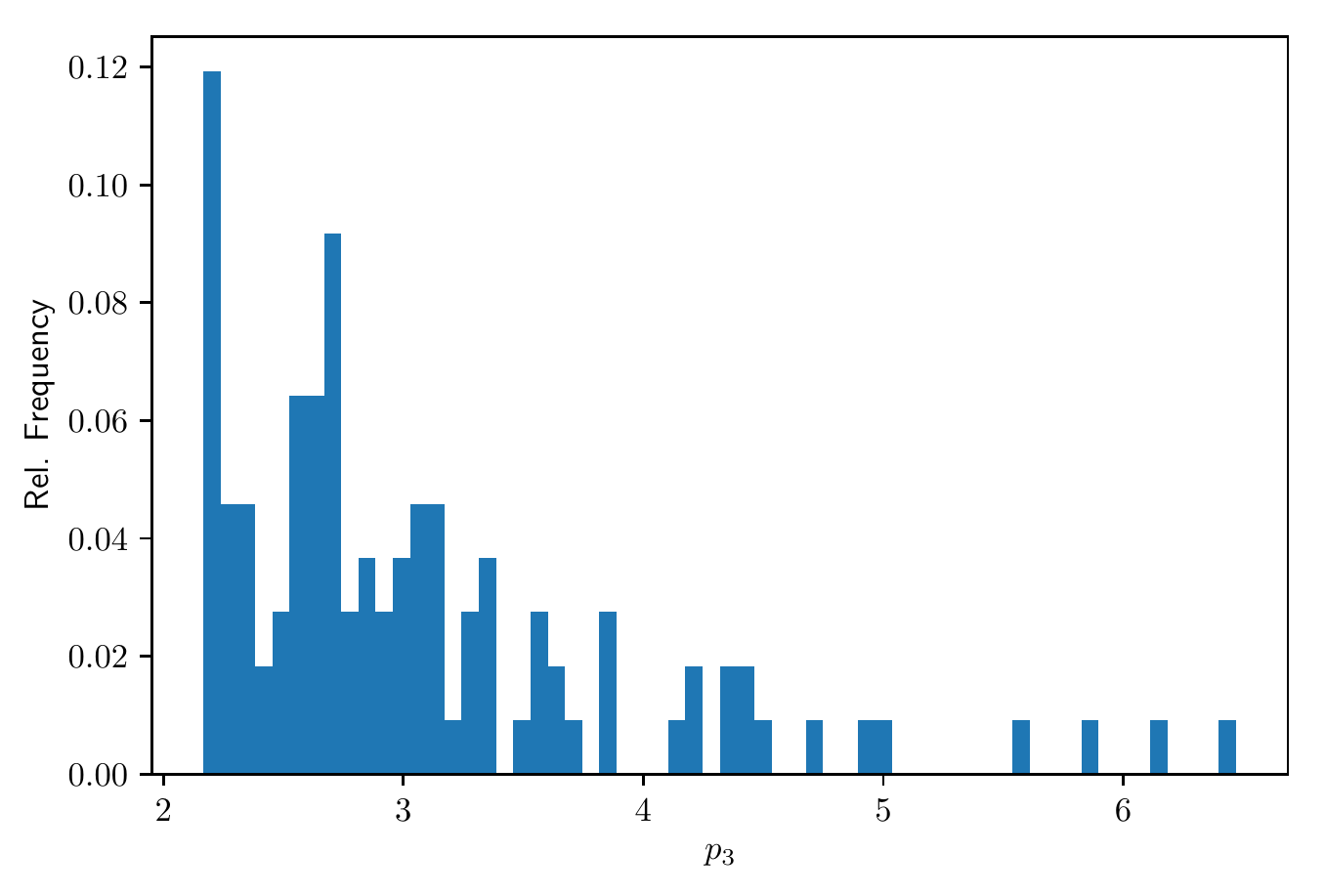}
	\caption{Histogram of $p_3$.}
\end{subfigure}
\end{center}
\caption{Histogram of learned exponents on the 3d faces dataset. To identify the different subspaces we choose the ordering $p_1 < p_2 < p_3$.}
\label{fig:exponent-histograms}
\end{figure*}

An interesting question when learning the exponents of the prior is, if the trivial case is learned, in which all the exponents become equal to $p_0$. This implies a fully factorized prior over all latent variables.
To explore this we set $p_0=2.1$ and initialize the exponents of the subspaces to $p_{1,\dots,3} = 2.0$.
We train $15$ models for each value of $\beta \in \{0.5, 1.0, 2.0, 3.0, 4.0, 5.0\}$.
Fig.~\ref{fig:exponent-histograms} depicts histograms of the learnt exponents, where we sort the exponents such that $p_1 < p_2 < p_3$.
Interestingly, the exponents with highest frequency are 1.95 for $p_1$, 1.98 for $p_2$, and 2.17 for $p_3$. Also, all values of $p_3$ are strictly larger than 2.17, meaning that these exponents are also always different from $p_0$. This small deviation from the Gaussian with $p=2.0$ seems to be sufficient to break symmetry and produce a more structured representation.  

\subsection{Hyperparameters}
\label{sec:hyperparams}

The hyperparameters that we use for the experiments on the dSprites dataset can be found in table~\ref{tab:hyperparams},
and the hyperparameters for the experiments on the 3D faces dataset in table~\ref{tab:hyperparamsfaces}.

\begin{table}
\caption{Hyperparameters of FactorVAE, $\beta$-VAE, $\beta$-TCVAE, and ISA-VAE evaluated on the dSprites dataset. We evaluated each model on the whole range of regularization strength parameters. ($\gamma$ for FactorVAE and $\beta$ for the other models)}
\newcommand{\head}[1]{\textnormal{\textbf{#1}}}
\begin{tabular}{l l l}%
  \toprule[1pt]
 \head{Parameter} & \head{Values}\\
 $\beta$ & [1, 1.5, 2, 2.5, 3, 3.5, 4, 5, 6]\\
$\gamma$ (FactorVAE) & [1, 2, 3, 4, 5, 6, 7, 8, 9, 10]\\
 Epochs & 20\\
 Learning Rate & 0.001\\
  Batch Size & 2048 \\
  Latent dimension & 20\\
  Loss function & Bernoulli\\
  Optimizer & Adam \\
  \bottomrule[1pt]
\end{tabular}
\label{tab:hyperparams}
\end{table}

\begin{table}
\caption{Hyperparameters of $\beta$-VAE, $\beta$-TCVAE, and ISA-VAE evaluated on the 3D faces dataset~\citep{3dfaces}. We evaluated each model on the whole range of regularization strength parameters.}
\newcommand{\head}[1]{\textnormal{\textbf{#1}}}
\begin{tabular}{l l l}%
  \toprule[1pt]
 \head{Parameter} & \head{Values}\\
 $\beta$ & [1, 1.5, 2, 2.5, 3, 4]\\
 Epochs & 1500\\
 Learning Rate & 0.001\\
  Batch Size & 2048 \\
  Latent dimension & 10\\
  Loss function & Bernoulli\\
  Optimizer & Adam \\
  \bottomrule[1pt]
\end{tabular}
\label{tab:hyperparamsfaces}
\end{table}

\subsection{Model Architecture (PyTorch)}
\label{sec:architecture}

The models were trained with the optimization algorithm Adam~\citep{Kingma-Ba-2015} using a learning rate parameter of 0.001

All unmentioned hyperparameters are PyTorch v0.41 defaults.

\small
\begin{lstlisting}
class MLPEncoder(nn.Module):
  def __init__(self, output_dim):
    super(MLPEncoder, self).__init__()
    self.output_dim = output_dim

    self.fc1 = nn.Linear(4096,1200)
    self.fc2 = nn.Linear(1200,1200)
    self.fc3 = nn.Linear(1200,output_dim)

    self.conv_z=nn.Conv2d(64,output_dim,4,1,0)

    self.act = nn.ReLU(inplace=True)

  def forward(self, x):
    h = x.view(-1, 64 * 64)
    h = self.act(self.fc1(h))
    h = self.act(self.fc2(h))
    h = self.fc3(h)
    z = h.view(x.size(0), self.output_dim)
    return z

class MLPDecoder(nn.Module):
  def __init__(self, input_dim):
    super(MLPDecoder, self).__init__()
    self.net = nn.Sequential(
      nn.Linear(input_dim, 1200),
      nn.Tanh(),
      nn.Linear(1200, 1200),
      nn.Tanh(),
      nn.Linear(1200, 1200),
      nn.Tanh(),
      nn.Linear(1200, 4096)
    )

  def forward(self, z):
    h = z.view(z.size(0), -1)
    h = self.net(h)
    mu_img = h.view(z.size(0), 1, 64, 64)
    return mu_img
\end{lstlisting}
\normalsize
Architecture of the encoder and decoder which is identical to the architecture in~\citet{Chen-et-al-2018}.

\small
\begin{lstlisting}
class Discriminator(nn.Module):
    def __init__(self, z_dim):
        super(Discriminator, self).__init__()
        self.net = nn.Sequential(
            nn.Linear(z_dim, 1000),
            nn.LeakyReLU(0.2, True),
            nn.Linear(1000, 1000),
            nn.LeakyReLU(0.2, True),
            nn.Linear(1000, 1000),
            nn.LeakyReLU(0.2, True),
            nn.Linear(1000, 1000),
            nn.LeakyReLU(0.2, True),
            nn.Linear(1000, 1000),
            nn.LeakyReLU(0.2, True),
            nn.Linear(1000, 2),
        )

    def forward(self, z):
        return self.net(z).squeeze()
\end{lstlisting}
\normalsize
Architecture of the discriminator of FactorVAE.

\subsection{Disentangled Representations and Latent Traversals}
\label{sec:traversals}

We use the plotting technique established in \citet{Chen-et-al-2018} for visualizing latent representations and additionally show images generated by traversals of the latent along the respective axis. The red and blue colour coding in the first column denotes the value of the latent variable for the respective x,y-coordinate of the sprite in the image. Coloured lines indicate the object shape with red for ellipse, green for square, and blue for heart. We observed that the MIG scores after training are usually bimodal: Either a model disentangles well or it does not reach a good MIG score. Therefore, to choose a representative model for each model class we take the average of the upper $50\%$ quantile of MIG scores and choose a representative model that minimizes the mahalanobis distance, defined by mean and variance of MIG score and reconstruction loss.
\todo{add violin plot}
ISA-layout:
ISA-VAE: $l_0=5$, $l_{1,\dots,5}=5$, $p_0=2.1$, $p_{1,\dots,5}=2.2$.





\input{traversal_ellipse_beta1_upperquantile.tex}
\input{traversal_square_beta1_upperquantile.tex}
\input{traversal_heart_beta1_upperquantile.tex}


\input{traversal_ellipse_beta2_upperquantile.tex}
\input{traversal_square_beta2_upperquantile.tex}
\input{traversal_heart_beta2_upperquantile.tex}


\input{traversal_ellipse_beta3_upperquantile.tex}
\input{traversal_square_beta3_upperquantile.tex}
\input{traversal_heart_beta3_upperquantile.tex}

\newpage
\input{supplement.tex}

\end{document}

%% file: math_commands.tex
\usepackage{amsmath,amsfonts,bm}

\def\figref#1{figure~\ref{#1}}

\def\secref#1{section~\ref{#1}}

\def\eqref#1{equation~\ref{#1}}

\def\1{\bm{1}}

\def\vs{{\bm{s}}}

\def\vu{{\bm{u}}}
\def\vv{{\bm{v}}}

\def\vx{{\bm{x}}}

\def\vz{{\bm{z}}}

\DeclareMathAlphabet{\mathsfit}{\encodingdefault}{\sfdefault}{m}{sl}
\SetMathAlphabet{\mathsfit}{bold}{\encodingdefault}{\sfdefault}{bx}{n}

\newcommand{\R}{\mathbb{R}}

\newcommand{\normltwo}{L^2}
\newcommand{\normlp}{L^p}


%% file: traversal_ellipse_beta1_upperquantile.tex
\begin{figure*}
\begin{center}
\begin{subfigure}{\linewidth}
\centering
\begin{minipage}{0.225\linewidth}
	\centering
	\includegraphics[width=\linewidth]{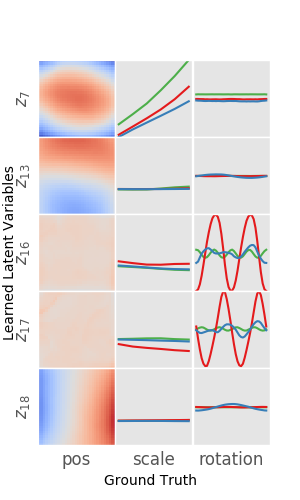}
\end{minipage}
\hspace{-1.25cm}
\begin{minipage}{0.6\linewidth}
\centering
\newcolumntype{V}{>{\centering\arraybackslash} m{.07\linewidth} }
\newcolumntype{T}{>{\centering\arraybackslash} m{.035\linewidth} }
\setlength{\tabcolsep}{0.5pt}
\renewcommand{\arraystretch}{1.0}
\begin{tabular}{T V V V V V V V V}
	\rotatebox{90}{\scalebox{.5}{scale}}&
	\includegraphics[width=\linewidth]{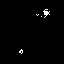}&
	\includegraphics[width=\linewidth]{figures/traversals/beta_1_upperquantile/compare_philly-vae-beta-sweep_beta_1_id_20_abm9_sampletraversal_1_ellipse_z_0_pos_1.png}&
	\includegraphics[width=\linewidth]{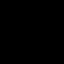}&
	\includegraphics[width=\linewidth]{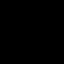}&
	\includegraphics[width=\linewidth]{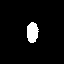}&
	\includegraphics[width=\linewidth]{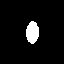}&
	\includegraphics[width=\linewidth]{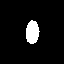}&
	\includegraphics[width=\linewidth]{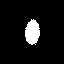}\\
	\rotatebox{90}{\scalebox{.5}{y-pos}}&
	\includegraphics[width=\linewidth]{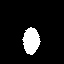}&
	\includegraphics[width=\linewidth]{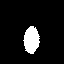}&
	\includegraphics[width=\linewidth]{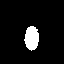}&
	\includegraphics[width=\linewidth]{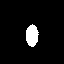}&
	\includegraphics[width=\linewidth]{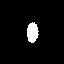}&
	\includegraphics[width=\linewidth]{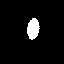}&
	\includegraphics[width=\linewidth]{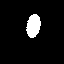}&
	\includegraphics[width=\linewidth]{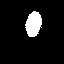}\\
	\rotatebox{90}{\scalebox{.5}{rotation}}&
	\includegraphics[width=\linewidth]{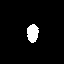}&
	\includegraphics[width=\linewidth]{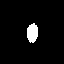}&
	\includegraphics[width=\linewidth]{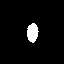}&
	\includegraphics[width=\linewidth]{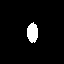}&
	\includegraphics[width=\linewidth]{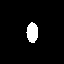}&
	\includegraphics[width=\linewidth]{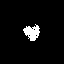}&
	\includegraphics[width=\linewidth]{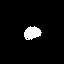}&
	\includegraphics[width=\linewidth]{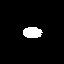}\\
	\rotatebox{90}{\scalebox{.5}{rotation}}&
	\includegraphics[width=\linewidth]{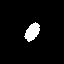}&
	\includegraphics[width=\linewidth]{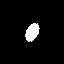}&
	\includegraphics[width=\linewidth]{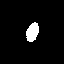}&
	\includegraphics[width=\linewidth]{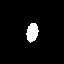}&
	\includegraphics[width=\linewidth]{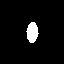}&
	\includegraphics[width=\linewidth]{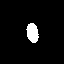}&
	\includegraphics[width=\linewidth]{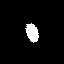}&
	\includegraphics[width=\linewidth]{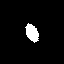}\\
	\rotatebox{90}{\scalebox{.5}{x-pos}}&
	\includegraphics[width=\linewidth]{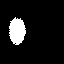}&
	\includegraphics[width=\linewidth]{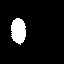}&
	\includegraphics[width=\linewidth]{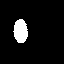}&
	\includegraphics[width=\linewidth]{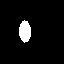}&
	\includegraphics[width=\linewidth]{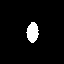}&
	\includegraphics[width=\linewidth]{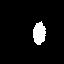}&
	\includegraphics[width=\linewidth]{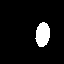}&
	\includegraphics[width=\linewidth]{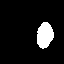}
\end{tabular}
\end{minipage}
	\caption{$\beta$-VAE, $\beta=1.0$, MIG: $0.14, logpx: -21.99$}
\end{subfigure}
\begin{subfigure}{\linewidth}
\centering
\begin{minipage}{0.225\linewidth}
	\centering
	\includegraphics[width=\linewidth]{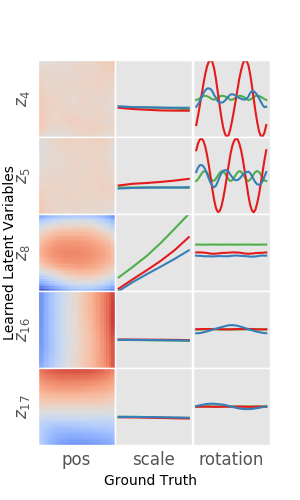}
\end{minipage}
\hspace{-1.25cm}
\begin{minipage}{0.6\linewidth}
\centering
\newcolumntype{V}{>{\centering\arraybackslash} m{.07\linewidth} }
\newcolumntype{T}{>{\centering\arraybackslash} m{.035\linewidth} }
\setlength{\tabcolsep}{0.5pt}
\renewcommand{\arraystretch}{1.0}
\begin{tabular}{T V V V V V V V V}
	\rotatebox{90}{\scalebox{.5}{rotation}}&
	\includegraphics[width=\linewidth]{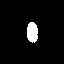}&
	\includegraphics[width=\linewidth]{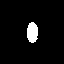}&
	\includegraphics[width=\linewidth]{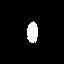}&
	\includegraphics[width=\linewidth]{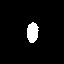}&
	\includegraphics[width=\linewidth]{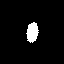}&
	\includegraphics[width=\linewidth]{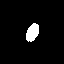}&
	\includegraphics[width=\linewidth]{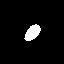}&
	\includegraphics[width=\linewidth]{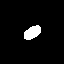}\\
	\rotatebox{90}{\scalebox{.5}{rotation}}&
	\includegraphics[width=\linewidth]{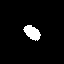}&
	\includegraphics[width=\linewidth]{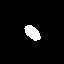}&
	\includegraphics[width=\linewidth]{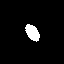}&
	\includegraphics[width=\linewidth]{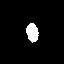}&
	\includegraphics[width=\linewidth]{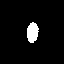}&
	\includegraphics[width=\linewidth]{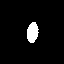}&
	\includegraphics[width=\linewidth]{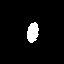}&
	\includegraphics[width=\linewidth]{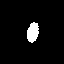}\\
	\rotatebox{90}{\scalebox{.5}{scale}}&
	\includegraphics[width=\linewidth]{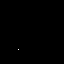}&
	\includegraphics[width=\linewidth]{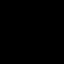}&
	\includegraphics[width=\linewidth]{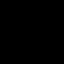}&
	\includegraphics[width=\linewidth]{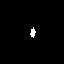}&
	\includegraphics[width=\linewidth]{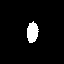}&
	\includegraphics[width=\linewidth]{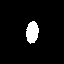}&
	\includegraphics[width=\linewidth]{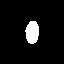}&
	\includegraphics[width=\linewidth]{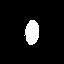}\\
	\rotatebox{90}{\scalebox{.5}{x-pos}}&
	\includegraphics[width=\linewidth]{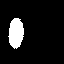}&
	\includegraphics[width=\linewidth]{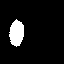}&
	\includegraphics[width=\linewidth]{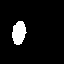}&
	\includegraphics[width=\linewidth]{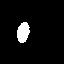}&
	\includegraphics[width=\linewidth]{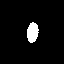}&
	\includegraphics[width=\linewidth]{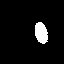}&
	\includegraphics[width=\linewidth]{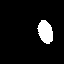}&
	\includegraphics[width=\linewidth]{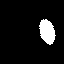}\\
	\rotatebox{90}{\scalebox{.5}{y-pos}}&
	\includegraphics[width=\linewidth]{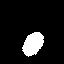}&
	\includegraphics[width=\linewidth]{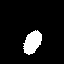}&
	\includegraphics[width=\linewidth]{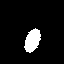}&
	\includegraphics[width=\linewidth]{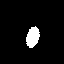}&
	\includegraphics[width=\linewidth]{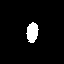}&
	\includegraphics[width=\linewidth]{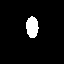}&
	\includegraphics[width=\linewidth]{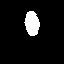}&
	\includegraphics[width=\linewidth]{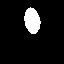}
\end{tabular}
\end{minipage}
	\caption{ISA-VAE, $\beta=1.0$, MIG: $0.20$, logpx: $-20.93$}
\end{subfigure}
\end{center}
\caption{Disentangled representations for representative models of the upper quantile of MIG scores for $\beta$-VAE (identical with $\beta$-TCVAE for $\beta=0$) and ISA-VAE (ISA-TCVAE identical for $\beta=0$) and latent traversals for the ellipse shape.}
\end{figure*}

%% file: traversal_square_beta1_upperquantile.tex
\begin{figure*}
\begin{center}
\begin{subfigure}{\linewidth}
\centering
\begin{minipage}{0.225\linewidth}
	\centering
	\includegraphics[width=\linewidth]{figures/traversals/beta_1_upperquantile/gt_vs_latent_vae_beta_1.png}
\end{minipage}
\hspace{-1.25cm}
\begin{minipage}{0.6\linewidth}
\centering
\newcolumntype{V}{>{\centering\arraybackslash} m{.07\linewidth} }
\newcolumntype{T}{>{\centering\arraybackslash} m{.035\linewidth} }
\setlength{\tabcolsep}{0.5pt}
\renewcommand{\arraystretch}{1.0}
\begin{tabular}{T V V V V V V V V}
	\rotatebox{90}{\scalebox{.5}{scale}}&
	\includegraphics[width=\linewidth]{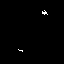}&
	\includegraphics[width=\linewidth]{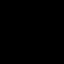}&
	\includegraphics[width=\linewidth]{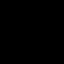}&
	\includegraphics[width=\linewidth]{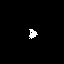}&
	\includegraphics[width=\linewidth]{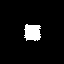}&
	\includegraphics[width=\linewidth]{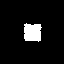}&
	\includegraphics[width=\linewidth]{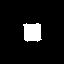}&
	\includegraphics[width=\linewidth]{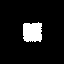}\\
	\rotatebox{90}{\scalebox{.5}{y-pos}}&
	\includegraphics[width=\linewidth]{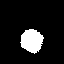}&
	\includegraphics[width=\linewidth]{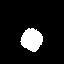}&
	\includegraphics[width=\linewidth]{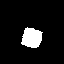}&
	\includegraphics[width=\linewidth]{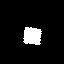}&
	\includegraphics[width=\linewidth]{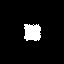}&
	\includegraphics[width=\linewidth]{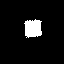}&
	\includegraphics[width=\linewidth]{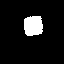}&
	\includegraphics[width=\linewidth]{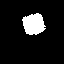}\\
	\rotatebox{90}{\scalebox{.5}{rotation}}&
	\includegraphics[width=\linewidth]{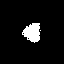}&
	\includegraphics[width=\linewidth]{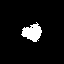}&
	\includegraphics[width=\linewidth]{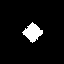}&
	\includegraphics[width=\linewidth]{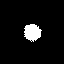}&
	\includegraphics[width=\linewidth]{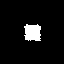}&
	\includegraphics[width=\linewidth]{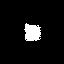}&
	\includegraphics[width=\linewidth]{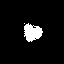}&
	\includegraphics[width=\linewidth]{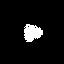}\\
	\rotatebox{90}{\scalebox{.5}{rotation}}&
	\includegraphics[width=\linewidth]{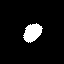}&
	\includegraphics[width=\linewidth]{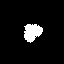}&
	\includegraphics[width=\linewidth]{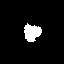}&
	\includegraphics[width=\linewidth]{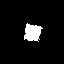}&
	\includegraphics[width=\linewidth]{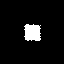}&
	\includegraphics[width=\linewidth]{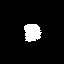}&
	\includegraphics[width=\linewidth]{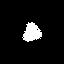}&
	\includegraphics[width=\linewidth]{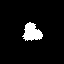}\\
	\rotatebox{90}{\scalebox{.5}{x-pos}}&
	\includegraphics[width=\linewidth]{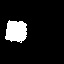}&
	\includegraphics[width=\linewidth]{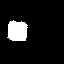}&
	\includegraphics[width=\linewidth]{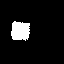}&
	\includegraphics[width=\linewidth]{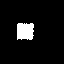}&
	\includegraphics[width=\linewidth]{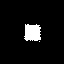}&
	\includegraphics[width=\linewidth]{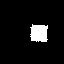}&
	\includegraphics[width=\linewidth]{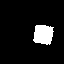}&
	\includegraphics[width=\linewidth]{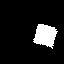}
\end{tabular}
\end{minipage}
	\caption{$\beta$-VAE, $\beta=1.0$, MIG: $0.14, logpx: -21.99$}
\end{subfigure}
\begin{subfigure}{\linewidth}
\centering
\begin{minipage}{0.225\linewidth}
	\centering
	\includegraphics[width=\linewidth]{figures/traversals/beta_1_upperquantile/gt_vs_latent_isa_vae_beta_1.png}
\end{minipage}
\hspace{-1.25cm}
\begin{minipage}{0.6\linewidth}
\centering
\newcolumntype{V}{>{\centering\arraybackslash} m{.07\linewidth} }
\newcolumntype{T}{>{\centering\arraybackslash} m{.035\linewidth} }
\setlength{\tabcolsep}{0.5pt}
\renewcommand{\arraystretch}{1.0}
\begin{tabular}{T V V V V V V V V}
	\rotatebox{90}{\scalebox{.5}{rotation}}&
	\includegraphics[width=\linewidth]{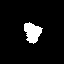}&
	\includegraphics[width=\linewidth]{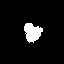}&
	\includegraphics[width=\linewidth]{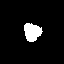}&
	\includegraphics[width=\linewidth]{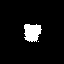}&
	\includegraphics[width=\linewidth]{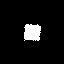}&
	\includegraphics[width=\linewidth]{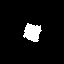}&
	\includegraphics[width=\linewidth]{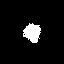}&
	\includegraphics[width=\linewidth]{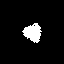}\\
	\rotatebox{90}{\scalebox{.5}{rotation}}&
	\includegraphics[width=\linewidth]{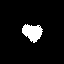}&
	\includegraphics[width=\linewidth]{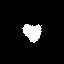}&
	\includegraphics[width=\linewidth]{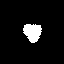}&
	\includegraphics[width=\linewidth]{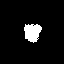}&
	\includegraphics[width=\linewidth]{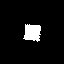}&
	\includegraphics[width=\linewidth]{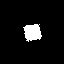}&
	\includegraphics[width=\linewidth]{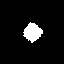}&
	\includegraphics[width=\linewidth]{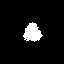}\\
	\rotatebox{90}{\scalebox{.5}{scale}}&
	\includegraphics[width=\linewidth]{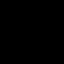}&
	\includegraphics[width=\linewidth]{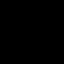}&
	\includegraphics[width=\linewidth]{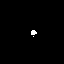}&
	\includegraphics[width=\linewidth]{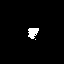}&
	\includegraphics[width=\linewidth]{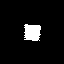}&
	\includegraphics[width=\linewidth]{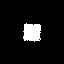}&
	\includegraphics[width=\linewidth]{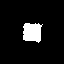}&
	\includegraphics[width=\linewidth]{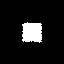}\\
	\rotatebox{90}{\scalebox{.5}{x-pos}}&
	\includegraphics[width=\linewidth]{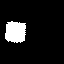}&
	\includegraphics[width=\linewidth]{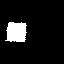}&
	\includegraphics[width=\linewidth]{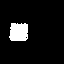}&
	\includegraphics[width=\linewidth]{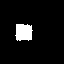}&
	\includegraphics[width=\linewidth]{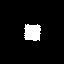}&
	\includegraphics[width=\linewidth]{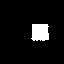}&
	\includegraphics[width=\linewidth]{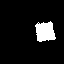}&
	\includegraphics[width=\linewidth]{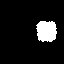}\\
	\rotatebox{90}{\scalebox{.5}{y-pos}}&
	\includegraphics[width=\linewidth]{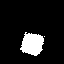}&
	\includegraphics[width=\linewidth]{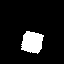}&
	\includegraphics[width=\linewidth]{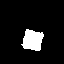}&
	\includegraphics[width=\linewidth]{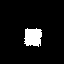}&
	\includegraphics[width=\linewidth]{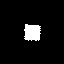}&
	\includegraphics[width=\linewidth]{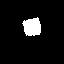}&
	\includegraphics[width=\linewidth]{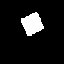}&
	\includegraphics[width=\linewidth]{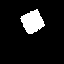}
\end{tabular}
\end{minipage}
	\caption{ISA-VAE, $\beta=1.0$, MIG: $0.20$, logpx: $-20.93$}
\end{subfigure}
\end{center}
\caption{Disentangled representations for representative models of the upper quantile of MIG scores for $\beta$-VAE (identical with $\beta$-TCVAE for $\beta=0$) and ISA-VAE (ISA-TCVAE identical for $\beta=0$) and latent traversals for the square shape.}
\end{figure*}

%% file: traversal_heart_beta1_upperquantile.tex
\begin{figure*}
\begin{center}
\begin{subfigure}{\linewidth}
\centering
\begin{minipage}{0.225\linewidth}
	\centering
	\includegraphics[width=\linewidth]{figures/traversals/beta_1_upperquantile/gt_vs_latent_vae_beta_1.png}
\end{minipage}
\hspace{-1.25cm}
\begin{minipage}{0.6\linewidth}
\centering
\newcolumntype{V}{>{\centering\arraybackslash} m{.07\linewidth} }
\newcolumntype{T}{>{\centering\arraybackslash} m{.035\linewidth} }
\setlength{\tabcolsep}{0.5pt}
\renewcommand{\arraystretch}{1.0}
\begin{tabular}{T V V V V V V V V}
	\rotatebox{90}{\scalebox{.5}{scale}}&
	\includegraphics[width=\linewidth]{figures/traversals/beta_1_upperquantile/compare_philly-vae-beta-sweep_beta_1_id_20_abm9_sampletraversal_1_heart_z_0_pos_0.png}&
	\includegraphics[width=\linewidth]{figures/traversals/beta_1_upperquantile/compare_philly-vae-beta-sweep_beta_1_id_20_abm9_sampletraversal_1_heart_z_0_pos_1.png}&
	\includegraphics[width=\linewidth]{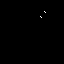}&
	\includegraphics[width=\linewidth]{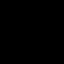}&
	\includegraphics[width=\linewidth]{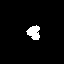}&
	\includegraphics[width=\linewidth]{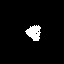}&
	\includegraphics[width=\linewidth]{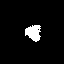}&
	\includegraphics[width=\linewidth]{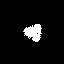}\\
	\rotatebox{90}{\scalebox{.5}{y-pos}}&
	\includegraphics[width=\linewidth]{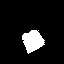}&
	\includegraphics[width=\linewidth]{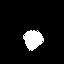}&
	\includegraphics[width=\linewidth]{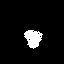}&
	\includegraphics[width=\linewidth]{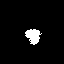}&
	\includegraphics[width=\linewidth]{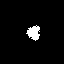}&
	\includegraphics[width=\linewidth]{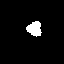}&
	\includegraphics[width=\linewidth]{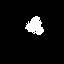}&
	\includegraphics[width=\linewidth]{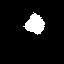}\\
	\rotatebox{90}{\scalebox{.5}{rotation}}&
	\includegraphics[width=\linewidth]{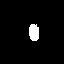}&
	\includegraphics[width=\linewidth]{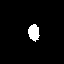}&
	\includegraphics[width=\linewidth]{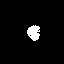}&
	\includegraphics[width=\linewidth]{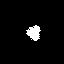}&
	\includegraphics[width=\linewidth]{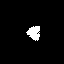}&
	\includegraphics[width=\linewidth]{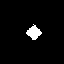}&
	\includegraphics[width=\linewidth]{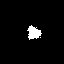}&
	\includegraphics[width=\linewidth]{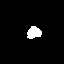}\\
	\rotatebox{90}{\scalebox{.5}{rotation}}&
	\includegraphics[width=\linewidth]{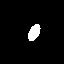}&
	\includegraphics[width=\linewidth]{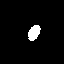}&
	\includegraphics[width=\linewidth]{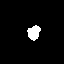}&
	\includegraphics[width=\linewidth]{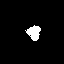}&
	\includegraphics[width=\linewidth]{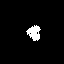}&
	\includegraphics[width=\linewidth]{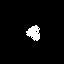}&
	\includegraphics[width=\linewidth]{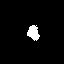}&
	\includegraphics[width=\linewidth]{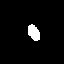}\\
	\rotatebox{90}{\scalebox{.5}{x-pos}}&
	\includegraphics[width=\linewidth]{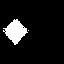}&
	\includegraphics[width=\linewidth]{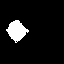}&
	\includegraphics[width=\linewidth]{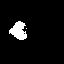}&
	\includegraphics[width=\linewidth]{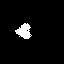}&
	\includegraphics[width=\linewidth]{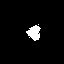}&
	\includegraphics[width=\linewidth]{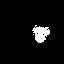}&
	\includegraphics[width=\linewidth]{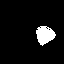}&
	\includegraphics[width=\linewidth]{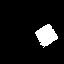}
\end{tabular}
\end{minipage}
	\caption{$\beta$-VAE, $\beta=1.0$, MIG: $0.14, logpx: -21.99$}
\end{subfigure}
\begin{subfigure}{\linewidth}
\centering
\begin{minipage}{0.225\linewidth}
	\centering
	\includegraphics[width=\linewidth]{figures/traversals/beta_1_upperquantile/gt_vs_latent_isa_vae_beta_1.png}
\end{minipage}
\hspace{-1.25cm}
\begin{minipage}{0.6\linewidth}
\centering
\newcolumntype{V}{>{\centering\arraybackslash} m{.07\linewidth} }
\newcolumntype{T}{>{\centering\arraybackslash} m{.035\linewidth} }
\setlength{\tabcolsep}{0.5pt}
\renewcommand{\arraystretch}{1.0}
\begin{tabular}{T V V V V V V V V}
	\rotatebox{90}{\scalebox{.5}{rotation}}&
	\includegraphics[width=\linewidth]{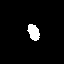}&
	\includegraphics[width=\linewidth]{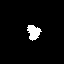}&
	\includegraphics[width=\linewidth]{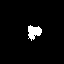}&
	\includegraphics[width=\linewidth]{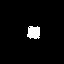}&
	\includegraphics[width=\linewidth]{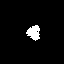}&
	\includegraphics[width=\linewidth]{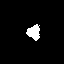}&
	\includegraphics[width=\linewidth]{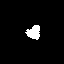}&
	\includegraphics[width=\linewidth]{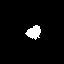}\\
	\rotatebox{90}{\scalebox{.5}{rotation}}&
	\includegraphics[width=\linewidth]{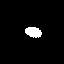}&
	\includegraphics[width=\linewidth]{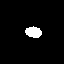}&
	\includegraphics[width=\linewidth]{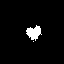}&
	\includegraphics[width=\linewidth]{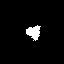}&
	\includegraphics[width=\linewidth]{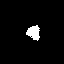}&
	\includegraphics[width=\linewidth]{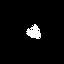}&
	\includegraphics[width=\linewidth]{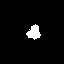}&
	\includegraphics[width=\linewidth]{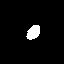}\\
	\rotatebox{90}{\scalebox{.5}{scale}}&
	\includegraphics[width=\linewidth]{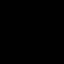}&
	\includegraphics[width=\linewidth]{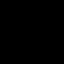}&
	\includegraphics[width=\linewidth]{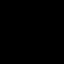}&
	\includegraphics[width=\linewidth]{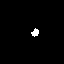}&
	\includegraphics[width=\linewidth]{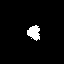}&
	\includegraphics[width=\linewidth]{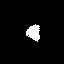}&
	\includegraphics[width=\linewidth]{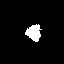}&
	\includegraphics[width=\linewidth]{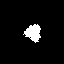}\\
	\rotatebox{90}{\scalebox{.5}{x-pos}}&
	\includegraphics[width=\linewidth]{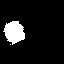}&
	\includegraphics[width=\linewidth]{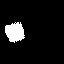}&
	\includegraphics[width=\linewidth]{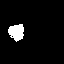}&
	\includegraphics[width=\linewidth]{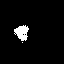}&
	\includegraphics[width=\linewidth]{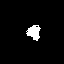}&
	\includegraphics[width=\linewidth]{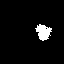}&
	\includegraphics[width=\linewidth]{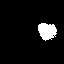}&
	\includegraphics[width=\linewidth]{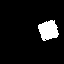}\\
	\rotatebox{90}{\scalebox{.5}{y-pos}}&
	\includegraphics[width=\linewidth]{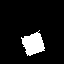}&
	\includegraphics[width=\linewidth]{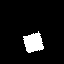}&
	\includegraphics[width=\linewidth]{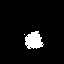}&
	\includegraphics[width=\linewidth]{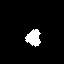}&
	\includegraphics[width=\linewidth]{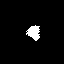}&
	\includegraphics[width=\linewidth]{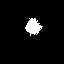}&
	\includegraphics[width=\linewidth]{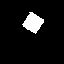}&
	\includegraphics[width=\linewidth]{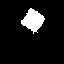}
\end{tabular}
\end{minipage}
	\caption{ISA-VAE, $\beta=1.0$, MIG: $0.20$, logpx: $-20.93$}
\end{subfigure}
\end{center}
\caption{Disentangled representations for representative models of the upper quantile of MIG scores for $\beta$-VAE (identical with $\beta$-TCVAE for $\beta=0$) and ISA-VAE (ISA-TCVAE identical for $\beta=0$) and latent traversals for the heart shape.}

\end{figure*}

%% file: traversal_ellipse_beta2_upperquantile.tex
\begin{figure*}
\begin{center}
\begin{subfigure}{\linewidth}
\centering
\begin{minipage}{0.225\linewidth}
	\centering
	\includegraphics[width=\linewidth]{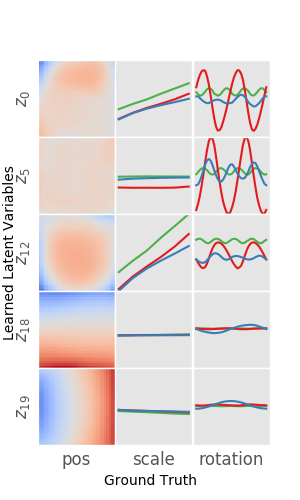}
\end{minipage}
\hspace{-1.25cm}
\begin{minipage}{0.6\linewidth}
\centering
\newcolumntype{V}{>{\centering\arraybackslash} m{.07\linewidth} }
\newcolumntype{T}{>{\centering\arraybackslash} m{.035\linewidth} }
\setlength{\tabcolsep}{0.5pt}
\renewcommand{\arraystretch}{1.0}
\begin{tabular}{T V V V V V V V V}
	\rotatebox{90}{\scalebox{.5}{rotation}}&
	\includegraphics[width=\linewidth]{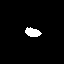}&
	\includegraphics[width=\linewidth]{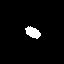}&
	\includegraphics[width=\linewidth]{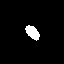}&
	\includegraphics[width=\linewidth]{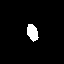}&
	\includegraphics[width=\linewidth]{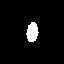}&
	\includegraphics[width=\linewidth]{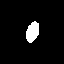}&
	\includegraphics[width=\linewidth]{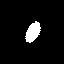}&
	\includegraphics[width=\linewidth]{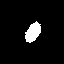}\\
	\rotatebox{90}{\scalebox{.5}{rotation}}&
	\includegraphics[width=\linewidth]{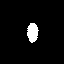}&
	\includegraphics[width=\linewidth]{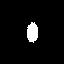}&
	\includegraphics[width=\linewidth]{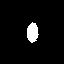}&
	\includegraphics[width=\linewidth]{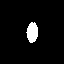}&
	\includegraphics[width=\linewidth]{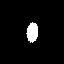}&
	\includegraphics[width=\linewidth]{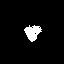}&
	\includegraphics[width=\linewidth]{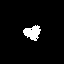}&
	\includegraphics[width=\linewidth]{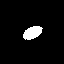}\\
	\rotatebox{90}{\scalebox{.5}{scale}}&
	\includegraphics[width=\linewidth]{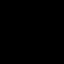}&
	\includegraphics[width=\linewidth]{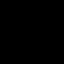}&
	\includegraphics[width=\linewidth]{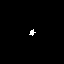}&
	\includegraphics[width=\linewidth]{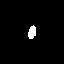}&
	\includegraphics[width=\linewidth]{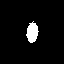}&
	\includegraphics[width=\linewidth]{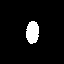}&
	\includegraphics[width=\linewidth]{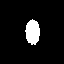}&
	\includegraphics[width=\linewidth]{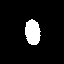}\\
	\rotatebox{90}{\scalebox{.5}{y-pos}}&
	\includegraphics[width=\linewidth]{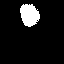}&
	\includegraphics[width=\linewidth]{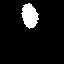}&
	\includegraphics[width=\linewidth]{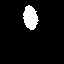}&
	\includegraphics[width=\linewidth]{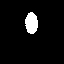}&
	\includegraphics[width=\linewidth]{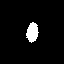}&
	\includegraphics[width=\linewidth]{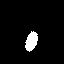}&
	\includegraphics[width=\linewidth]{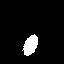}&
	\includegraphics[width=\linewidth]{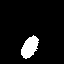}\\
	\rotatebox{90}{\scalebox{.5}{x-pos}}&
	\includegraphics[width=\linewidth]{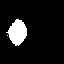}&
	\includegraphics[width=\linewidth]{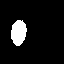}&
	\includegraphics[width=\linewidth]{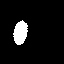}&
	\includegraphics[width=\linewidth]{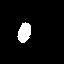}&
	\includegraphics[width=\linewidth]{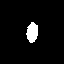}&
	\includegraphics[width=\linewidth]{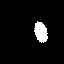}&
	\includegraphics[width=\linewidth]{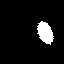}&
	\includegraphics[width=\linewidth]{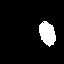}
\end{tabular}
\end{minipage}
	\caption{$\beta$-VAE, $\beta=2.0$, MIG: $0.28$, logpx: $-29.40$}
\end{subfigure}
\begin{subfigure}{\linewidth}
\centering
\begin{minipage}{0.225\linewidth}
	\centering
	\includegraphics[width=\linewidth]{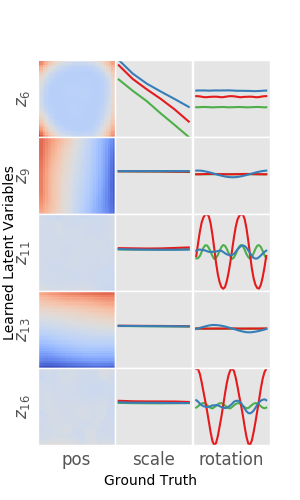}
\end{minipage}
\hspace{-1.25cm}
\begin{minipage}{0.6\linewidth}
\centering
\newcolumntype{V}{>{\centering\arraybackslash} m{.07\linewidth} }
\newcolumntype{T}{>{\centering\arraybackslash} m{.035\linewidth} }
\setlength{\tabcolsep}{0.5pt}
\renewcommand{\arraystretch}{1.0}
\begin{tabular}{T V V V V V V V V}
	\rotatebox{90}{\scalebox{.5}{scale}}&
	\includegraphics[width=\linewidth]{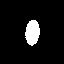}&
	\includegraphics[width=\linewidth]{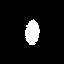}&
	\includegraphics[width=\linewidth]{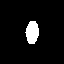}&
	\includegraphics[width=\linewidth]{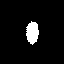}&
	\includegraphics[width=\linewidth]{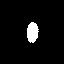}&
	\includegraphics[width=\linewidth]{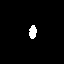}&
	\includegraphics[width=\linewidth]{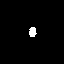}&
	\includegraphics[width=\linewidth]{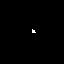}\\
	\rotatebox{90}{\scalebox{.5}{x-pos}}&
	\includegraphics[width=\linewidth]{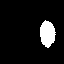}&
	\includegraphics[width=\linewidth]{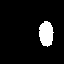}&
	\includegraphics[width=\linewidth]{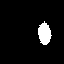}&
	\includegraphics[width=\linewidth]{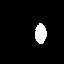}&
	\includegraphics[width=\linewidth]{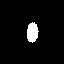}&
	\includegraphics[width=\linewidth]{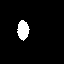}&
	\includegraphics[width=\linewidth]{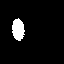}&
	\includegraphics[width=\linewidth]{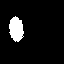}\\
	\rotatebox{90}{\scalebox{.5}{rotation}}&
	\includegraphics[width=\linewidth]{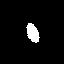}&
	\includegraphics[width=\linewidth]{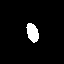}&
	\includegraphics[width=\linewidth]{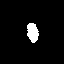}&
	\includegraphics[width=\linewidth]{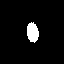}&
	\includegraphics[width=\linewidth]{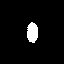}&
	\includegraphics[width=\linewidth]{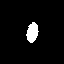}&
	\includegraphics[width=\linewidth]{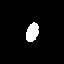}&
	\includegraphics[width=\linewidth]{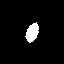}\\
	\rotatebox{90}{\scalebox{.5}{y-pos}}&
	\includegraphics[width=\linewidth]{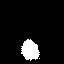}&
	\includegraphics[width=\linewidth]{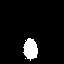}&
	\includegraphics[width=\linewidth]{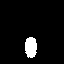}&
	\includegraphics[width=\linewidth]{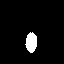}&
	\includegraphics[width=\linewidth]{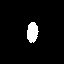}&
	\includegraphics[width=\linewidth]{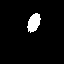}&
	\includegraphics[width=\linewidth]{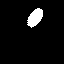}&
	\includegraphics[width=\linewidth]{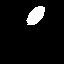}\\
	\rotatebox{90}{\scalebox{.5}{rotation}}&
	\includegraphics[width=\linewidth]{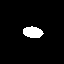}&
	\includegraphics[width=\linewidth]{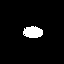}&
	\includegraphics[width=\linewidth]{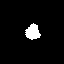}&
	\includegraphics[width=\linewidth]{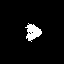}&
	\includegraphics[width=\linewidth]{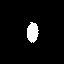}&
	\includegraphics[width=\linewidth]{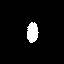}&
	\includegraphics[width=\linewidth]{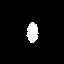}&
	\includegraphics[width=\linewidth]{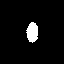}
\end{tabular}
\end{minipage}
	\caption{$\beta$-TCVAE, $\beta=2.0$, MIG: $0.30$, logpx: $-27.15$}
\end{subfigure}\\
\begin{subfigure}{\linewidth}
\centering
\begin{minipage}{0.225\linewidth}
	\centering
	\includegraphics[width=\linewidth]{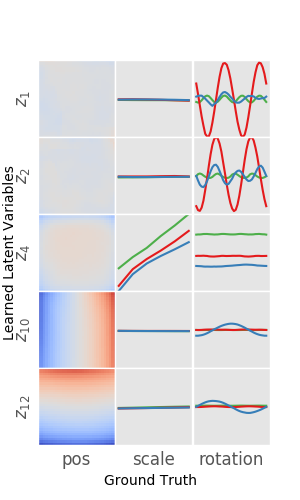}
\end{minipage}
\hspace{-1.25cm}
\begin{minipage}{0.6\linewidth}
\centering
\newcolumntype{V}{>{\centering\arraybackslash} m{.07\linewidth} }
\newcolumntype{T}{>{\centering\arraybackslash} m{.035\linewidth} }
\setlength{\tabcolsep}{0.5pt}
\renewcommand{\arraystretch}{1.0}
\begin{tabular}{T V V V V V V V V}
	\rotatebox{90}{\scalebox{.5}{rotation}}&
	\includegraphics[width=\linewidth]{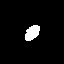}&
	\includegraphics[width=\linewidth]{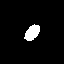}&
	\includegraphics[width=\linewidth]{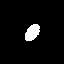}&
	\includegraphics[width=\linewidth]{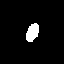}&
	\includegraphics[width=\linewidth]{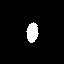}&
	\includegraphics[width=\linewidth]{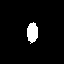}&
	\includegraphics[width=\linewidth]{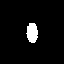}&
	\includegraphics[width=\linewidth]{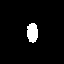}\\
	\rotatebox{90}{\scalebox{.5}{rotation}}&
	\includegraphics[width=\linewidth]{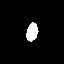}&
	\includegraphics[width=\linewidth]{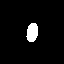}&
	\includegraphics[width=\linewidth]{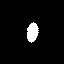}&
	\includegraphics[width=\linewidth]{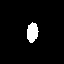}&
	\includegraphics[width=\linewidth]{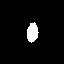}&
	\includegraphics[width=\linewidth]{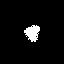}&
	\includegraphics[width=\linewidth]{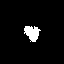}&
	\includegraphics[width=\linewidth]{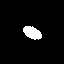}\\
	\rotatebox{90}{\scalebox{.5}{scale}}&
	\includegraphics[width=\linewidth]{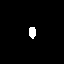}&
	\includegraphics[width=\linewidth]{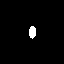}&
	\includegraphics[width=\linewidth]{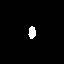}&
	\includegraphics[width=\linewidth]{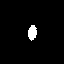}&
	\includegraphics[width=\linewidth]{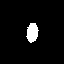}&
	\includegraphics[width=\linewidth]{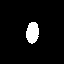}&
	\includegraphics[width=\linewidth]{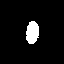}&
	\includegraphics[width=\linewidth]{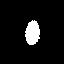}\\
	\rotatebox{90}{\scalebox{.5}{x-pos}}&
	\includegraphics[width=\linewidth]{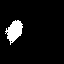}&
	\includegraphics[width=\linewidth]{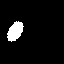}&
	\includegraphics[width=\linewidth]{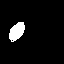}&
	\includegraphics[width=\linewidth]{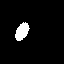}&
	\includegraphics[width=\linewidth]{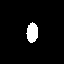}&
	\includegraphics[width=\linewidth]{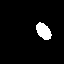}&
	\includegraphics[width=\linewidth]{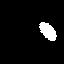}&
	\includegraphics[width=\linewidth]{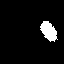}\\
	\rotatebox{90}{\scalebox{.5}{y-pos}}&
	\includegraphics[width=\linewidth]{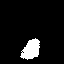}&
	\includegraphics[width=\linewidth]{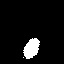}&
	\includegraphics[width=\linewidth]{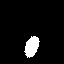}&
	\includegraphics[width=\linewidth]{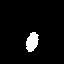}&
	\includegraphics[width=\linewidth]{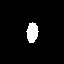}&
	\includegraphics[width=\linewidth]{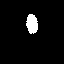}&
	\includegraphics[width=\linewidth]{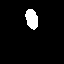}&
	\includegraphics[width=\linewidth]{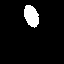}
\end{tabular}
\end{minipage}
	\caption{ISA-VAE, $\beta=2.0$, MIG: $0.41$, logpx: $-25.86$}
\end{subfigure}
\end{center}
\caption{Disentangled representations for models representative for the upper quantile of MIG scores for $\beta=2.0$ for $\beta$-VAE, $\beta$-TCVAE, ISA-VAE and ISA-TCVAE and latent traversals for the ellipse shape.}
\end{figure*}

%% file: traversal_square_beta2_upperquantile.tex
\begin{figure*}
\begin{center}
\begin{subfigure}{\linewidth}
\centering
\begin{minipage}{0.225\linewidth}
	\centering
	\includegraphics[width=\linewidth]{figures/traversals/beta_2_upperquantile/gt_vs_latent_vae_beta_2.png}
\end{minipage}
\hspace{-1.25cm}
\begin{minipage}{0.6\linewidth}
\centering
\newcolumntype{V}{>{\centering\arraybackslash} m{.07\linewidth} }
\newcolumntype{T}{>{\centering\arraybackslash} m{.035\linewidth} }
\setlength{\tabcolsep}{0.5pt}
\renewcommand{\arraystretch}{1.0}
\begin{tabular}{T V V V V V V V V}
	\rotatebox{90}{\scalebox{.5}{rotation}}&
	\includegraphics[width=\linewidth]{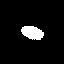}&
	\includegraphics[width=\linewidth]{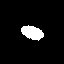}&
	\includegraphics[width=\linewidth]{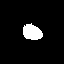}&
	\includegraphics[width=\linewidth]{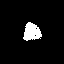}&
	\includegraphics[width=\linewidth]{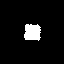}&
	\includegraphics[width=\linewidth]{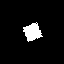}&
	\includegraphics[width=\linewidth]{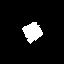}&
	\includegraphics[width=\linewidth]{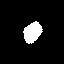}\\
	\rotatebox{90}{\scalebox{.5}{rotation}}&
	\includegraphics[width=\linewidth]{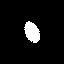}&
	\includegraphics[width=\linewidth]{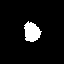}&
	\includegraphics[width=\linewidth]{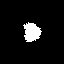}&
	\includegraphics[width=\linewidth]{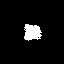}&
	\includegraphics[width=\linewidth]{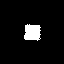}&
	\includegraphics[width=\linewidth]{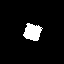}&
	\includegraphics[width=\linewidth]{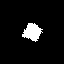}&
	\includegraphics[width=\linewidth]{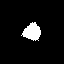}\\
	\rotatebox{90}{\scalebox{.5}{scale}}&
	\includegraphics[width=\linewidth]{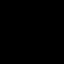}&
	\includegraphics[width=\linewidth]{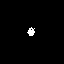}&
	\includegraphics[width=\linewidth]{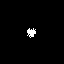}&
	\includegraphics[width=\linewidth]{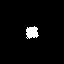}&
	\includegraphics[width=\linewidth]{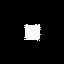}&
	\includegraphics[width=\linewidth]{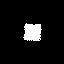}&
	\includegraphics[width=\linewidth]{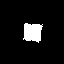}&
	\includegraphics[width=\linewidth]{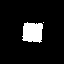}\\
	\rotatebox{90}{\scalebox{.5}{y-pos}}&
	\includegraphics[width=\linewidth]{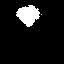}&
	\includegraphics[width=\linewidth]{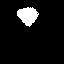}&
	\includegraphics[width=\linewidth]{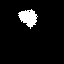}&
	\includegraphics[width=\linewidth]{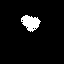}&
	\includegraphics[width=\linewidth]{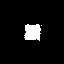}&
	\includegraphics[width=\linewidth]{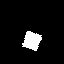}&
	\includegraphics[width=\linewidth]{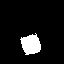}&
	\includegraphics[width=\linewidth]{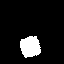}\\
	\rotatebox{90}{\scalebox{.5}{x-pos}}&
	\includegraphics[width=\linewidth]{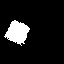}&
	\includegraphics[width=\linewidth]{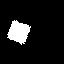}&
	\includegraphics[width=\linewidth]{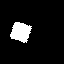}&
	\includegraphics[width=\linewidth]{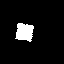}&
	\includegraphics[width=\linewidth]{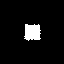}&
	\includegraphics[width=\linewidth]{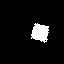}&
	\includegraphics[width=\linewidth]{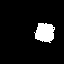}&
	\includegraphics[width=\linewidth]{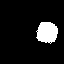}
\end{tabular}
\end{minipage}
	\caption{$\beta$-VAE, $\beta=2.0$, MIG: $0.28$, logpx: $-29.40$}
\end{subfigure}
\begin{subfigure}{\linewidth}
\centering
\begin{minipage}{0.225\linewidth}
	\centering
	\includegraphics[width=\linewidth]{figures/traversals/beta_2_upperquantile/gt_vs_latent_tcvae_beta_2.png}
\end{minipage}
\hspace{-1.25cm}
\begin{minipage}{0.6\linewidth}
\centering
\newcolumntype{V}{>{\centering\arraybackslash} m{.07\linewidth} }
\newcolumntype{T}{>{\centering\arraybackslash} m{.035\linewidth} }
\setlength{\tabcolsep}{0.5pt}
\renewcommand{\arraystretch}{1.0}
\begin{tabular}{T V V V V V V V V}
	\rotatebox{90}{\scalebox{.5}{scale}}&
	\includegraphics[width=\linewidth]{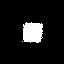}&
	\includegraphics[width=\linewidth]{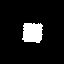}&
	\includegraphics[width=\linewidth]{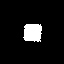}&
	\includegraphics[width=\linewidth]{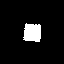}&
	\includegraphics[width=\linewidth]{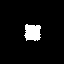}&
	\includegraphics[width=\linewidth]{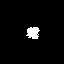}&
	\includegraphics[width=\linewidth]{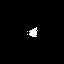}&
	\includegraphics[width=\linewidth]{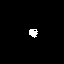}\\
	\rotatebox{90}{\scalebox{.5}{x-pos}}&
	\includegraphics[width=\linewidth]{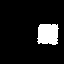}&
	\includegraphics[width=\linewidth]{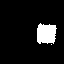}&
	\includegraphics[width=\linewidth]{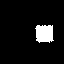}&
	\includegraphics[width=\linewidth]{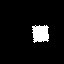}&
	\includegraphics[width=\linewidth]{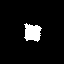}&
	\includegraphics[width=\linewidth]{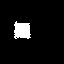}&
	\includegraphics[width=\linewidth]{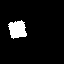}&
	\includegraphics[width=\linewidth]{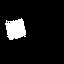}\\
	\rotatebox{90}{\scalebox{.5}{rotation}}&
	\includegraphics[width=\linewidth]{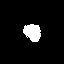}&
	\includegraphics[width=\linewidth]{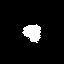}&
	\includegraphics[width=\linewidth]{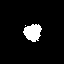}&
	\includegraphics[width=\linewidth]{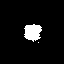}&
	\includegraphics[width=\linewidth]{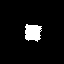}&
	\includegraphics[width=\linewidth]{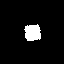}&
	\includegraphics[width=\linewidth]{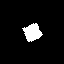}&
	\includegraphics[width=\linewidth]{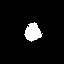}\\
	\rotatebox{90}{\scalebox{.5}{y-pos}}&
	\includegraphics[width=\linewidth]{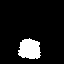}&
	\includegraphics[width=\linewidth]{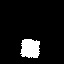}&
	\includegraphics[width=\linewidth]{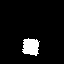}&
	\includegraphics[width=\linewidth]{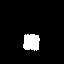}&
	\includegraphics[width=\linewidth]{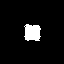}&
	\includegraphics[width=\linewidth]{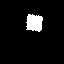}&
	\includegraphics[width=\linewidth]{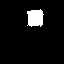}&
	\includegraphics[width=\linewidth]{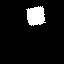}\\
	\rotatebox{90}{\scalebox{.5}{rotation}}&
	\includegraphics[width=\linewidth]{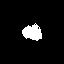}&
	\includegraphics[width=\linewidth]{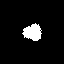}&
	\includegraphics[width=\linewidth]{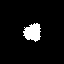}&
	\includegraphics[width=\linewidth]{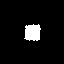}&
	\includegraphics[width=\linewidth]{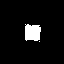}&
	\includegraphics[width=\linewidth]{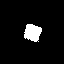}&
	\includegraphics[width=\linewidth]{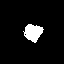}&
	\includegraphics[width=\linewidth]{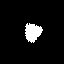}
\end{tabular}
\end{minipage}
	\caption{$\beta$-TCVAE, $\beta=2.0$, MIG: $0.30$, logpx: $-27.15$}
\end{subfigure}\\
\begin{subfigure}{\linewidth}
\centering
\begin{minipage}{0.225\linewidth}
	\centering
	\includegraphics[width=\linewidth]{figures/traversals/beta_2_upperquantile/gt_vs_latent_isa_vae_beta_2.png}
\end{minipage}
\hspace{-1.25cm}
\begin{minipage}{0.6\linewidth}
\centering
\newcolumntype{V}{>{\centering\arraybackslash} m{.07\linewidth} }
\newcolumntype{T}{>{\centering\arraybackslash} m{.035\linewidth} }
\setlength{\tabcolsep}{0.5pt}
\renewcommand{\arraystretch}{1.0}
\begin{tabular}{T V V V V V V V V}
	\rotatebox{90}{\scalebox{.5}{rotation}}&
	\includegraphics[width=\linewidth]{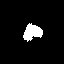}&
	\includegraphics[width=\linewidth]{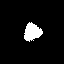}&
	\includegraphics[width=\linewidth]{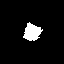}&
	\includegraphics[width=\linewidth]{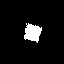}&
	\includegraphics[width=\linewidth]{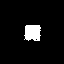}&
	\includegraphics[width=\linewidth]{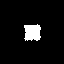}&
	\includegraphics[width=\linewidth]{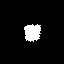}&
	\includegraphics[width=\linewidth]{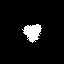}\\
	\rotatebox{90}{\scalebox{.5}{rotation}}&
	\includegraphics[width=\linewidth]{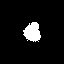}&
	\includegraphics[width=\linewidth]{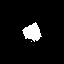}&
	\includegraphics[width=\linewidth]{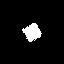}&
	\includegraphics[width=\linewidth]{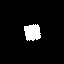}&
	\includegraphics[width=\linewidth]{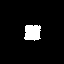}&
	\includegraphics[width=\linewidth]{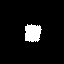}&
	\includegraphics[width=\linewidth]{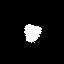}&
	\includegraphics[width=\linewidth]{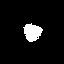}\\
	\rotatebox{90}{\scalebox{.5}{scale}}&
	\includegraphics[width=\linewidth]{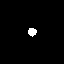}&
	\includegraphics[width=\linewidth]{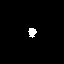}&
	\includegraphics[width=\linewidth]{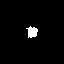}&
	\includegraphics[width=\linewidth]{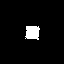}&
	\includegraphics[width=\linewidth]{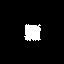}&
	\includegraphics[width=\linewidth]{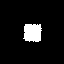}&
	\includegraphics[width=\linewidth]{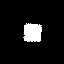}&
	\includegraphics[width=\linewidth]{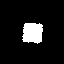}\\
	\rotatebox{90}{\scalebox{.5}{x-pos}}&
	\includegraphics[width=\linewidth]{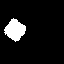}&
	\includegraphics[width=\linewidth]{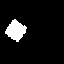}&
	\includegraphics[width=\linewidth]{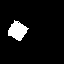}&
	\includegraphics[width=\linewidth]{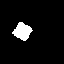}&
	\includegraphics[width=\linewidth]{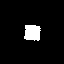}&
	\includegraphics[width=\linewidth]{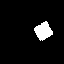}&
	\includegraphics[width=\linewidth]{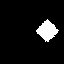}&
	\includegraphics[width=\linewidth]{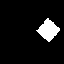}\\
	\rotatebox{90}{\scalebox{.5}{y-pos}}&
	\includegraphics[width=\linewidth]{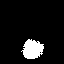}&
	\includegraphics[width=\linewidth]{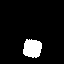}&
	\includegraphics[width=\linewidth]{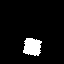}&
	\includegraphics[width=\linewidth]{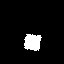}&
	\includegraphics[width=\linewidth]{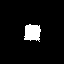}&
	\includegraphics[width=\linewidth]{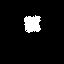}&
	\includegraphics[width=\linewidth]{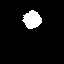}&
	\includegraphics[width=\linewidth]{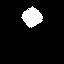}
\end{tabular}
\end{minipage}
	\caption{ISA-VAE, $\beta=2.0$, MIG: $0.41$, logpx: $-25.86$}
\end{subfigure}
\end{center}
\caption{Disentangled representations for models representative for the upper quantile of MIG scores for $\beta=2.0$ for $\beta$-VAE, $\beta$-TCVAE, ISA-VAE and ISA-TCVAE and latent traversals for the square shape.}
\end{figure*}

%% file: traversal_heart_beta2_upperquantile.tex
\begin{figure*}
\begin{center}
\begin{subfigure}{\linewidth}
\centering
\begin{minipage}{0.225\linewidth}
	\centering
	\includegraphics[width=\linewidth]{figures/traversals/beta_2_upperquantile/gt_vs_latent_vae_beta_2.png}
\end{minipage}
\hspace{-1.25cm}
\begin{minipage}{0.6\linewidth}
\centering
\newcolumntype{V}{>{\centering\arraybackslash} m{.07\linewidth} }
\newcolumntype{T}{>{\centering\arraybackslash} m{.035\linewidth} }
\setlength{\tabcolsep}{0.5pt}
\renewcommand{\arraystretch}{1.0}
\begin{tabular}{T V V V V V V V V}
	\rotatebox{90}{\scalebox{.5}{rotation}}&
	\includegraphics[width=\linewidth]{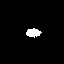}&
	\includegraphics[width=\linewidth]{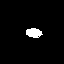}&
	\includegraphics[width=\linewidth]{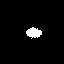}&
	\includegraphics[width=\linewidth]{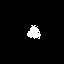}&
	\includegraphics[width=\linewidth]{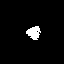}&
	\includegraphics[width=\linewidth]{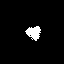}&
	\includegraphics[width=\linewidth]{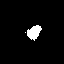}&
	\includegraphics[width=\linewidth]{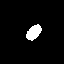}\\
	\rotatebox{90}{\scalebox{.5}{rotation}}&
	\includegraphics[width=\linewidth]{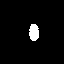}&
	\includegraphics[width=\linewidth]{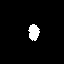}&
	\includegraphics[width=\linewidth]{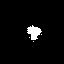}&
	\includegraphics[width=\linewidth]{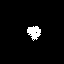}&
	\includegraphics[width=\linewidth]{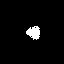}&
	\includegraphics[width=\linewidth]{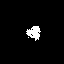}&
	\includegraphics[width=\linewidth]{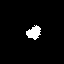}&
	\includegraphics[width=\linewidth]{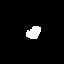}\\
	\rotatebox{90}{\scalebox{.5}{scale}}&
	\includegraphics[width=\linewidth]{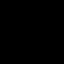}&
	\includegraphics[width=\linewidth]{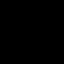}&
	\includegraphics[width=\linewidth]{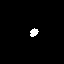}&
	\includegraphics[width=\linewidth]{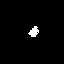}&
	\includegraphics[width=\linewidth]{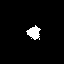}&
	\includegraphics[width=\linewidth]{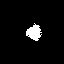}&
	\includegraphics[width=\linewidth]{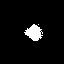}&
	\includegraphics[width=\linewidth]{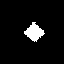}\\
	\rotatebox{90}{\scalebox{.5}{y-pos}}&
	\includegraphics[width=\linewidth]{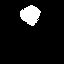}&
	\includegraphics[width=\linewidth]{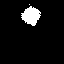}&
	\includegraphics[width=\linewidth]{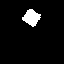}&
	\includegraphics[width=\linewidth]{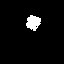}&
	\includegraphics[width=\linewidth]{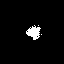}&
	\includegraphics[width=\linewidth]{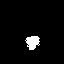}&
	\includegraphics[width=\linewidth]{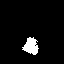}&
	\includegraphics[width=\linewidth]{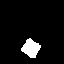}\\
	\rotatebox{90}{\scalebox{.5}{x-pos}}&
	\includegraphics[width=\linewidth]{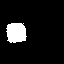}&
	\includegraphics[width=\linewidth]{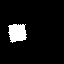}&
	\includegraphics[width=\linewidth]{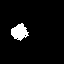}&
	\includegraphics[width=\linewidth]{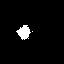}&
	\includegraphics[width=\linewidth]{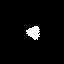}&
	\includegraphics[width=\linewidth]{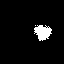}&
	\includegraphics[width=\linewidth]{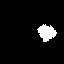}&
	\includegraphics[width=\linewidth]{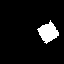}
\end{tabular}
\end{minipage}
	\caption{$\beta$-VAE, $\beta=2.0$, MIG: $0.28$, logpx: $-29.40$}
\end{subfigure}
\begin{subfigure}{\linewidth}
\centering
\begin{minipage}{0.225\linewidth}
	\centering
	\includegraphics[width=\linewidth]{figures/traversals/beta_2_upperquantile/gt_vs_latent_tcvae_beta_2.png}
\end{minipage}
\hspace{-1.25cm}
\begin{minipage}{0.6\linewidth}
\centering
\newcolumntype{V}{>{\centering\arraybackslash} m{.07\linewidth} }
\newcolumntype{T}{>{\centering\arraybackslash} m{.035\linewidth} }
\setlength{\tabcolsep}{0.5pt}
\renewcommand{\arraystretch}{1.0}
\begin{tabular}{T V V V V V V V V}
	\rotatebox{90}{\scalebox{.5}{scale}}&
	\includegraphics[width=\linewidth]{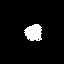}&
	\includegraphics[width=\linewidth]{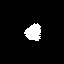}&
	\includegraphics[width=\linewidth]{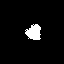}&
	\includegraphics[width=\linewidth]{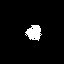}&
	\includegraphics[width=\linewidth]{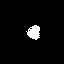}&
	\includegraphics[width=\linewidth]{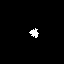}&
	\includegraphics[width=\linewidth]{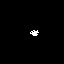}&
	\includegraphics[width=\linewidth]{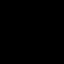}\\
	\rotatebox{90}{\scalebox{.5}{x-pos}}&
	\includegraphics[width=\linewidth]{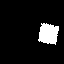}&
	\includegraphics[width=\linewidth]{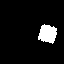}&
	\includegraphics[width=\linewidth]{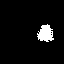}&
	\includegraphics[width=\linewidth]{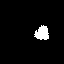}&
	\includegraphics[width=\linewidth]{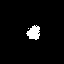}&
	\includegraphics[width=\linewidth]{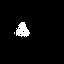}&
	\includegraphics[width=\linewidth]{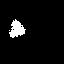}&
	\includegraphics[width=\linewidth]{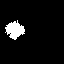}\\
	\rotatebox{90}{\scalebox{.5}{rotation}}&
	\includegraphics[width=\linewidth]{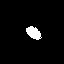}&
	\includegraphics[width=\linewidth]{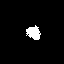}&
	\includegraphics[width=\linewidth]{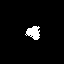}&
	\includegraphics[width=\linewidth]{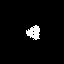}&
	\includegraphics[width=\linewidth]{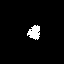}&
	\includegraphics[width=\linewidth]{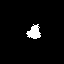}&
	\includegraphics[width=\linewidth]{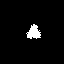}&
	\includegraphics[width=\linewidth]{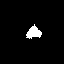}\\
	\rotatebox{90}{\scalebox{.5}{y-pos}}&
	\includegraphics[width=\linewidth]{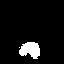}&
	\includegraphics[width=\linewidth]{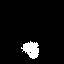}&
	\includegraphics[width=\linewidth]{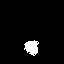}&
	\includegraphics[width=\linewidth]{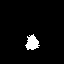}&
	\includegraphics[width=\linewidth]{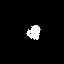}&
	\includegraphics[width=\linewidth]{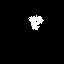}&
	\includegraphics[width=\linewidth]{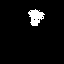}&
	\includegraphics[width=\linewidth]{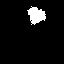}\\
	\rotatebox{90}{\scalebox{.5}{rotation}}&
	\includegraphics[width=\linewidth]{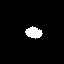}&
	\includegraphics[width=\linewidth]{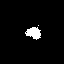}&
	\includegraphics[width=\linewidth]{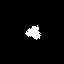}&
	\includegraphics[width=\linewidth]{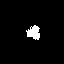}&
	\includegraphics[width=\linewidth]{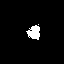}&
	\includegraphics[width=\linewidth]{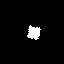}&
	\includegraphics[width=\linewidth]{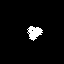}&
	\includegraphics[width=\linewidth]{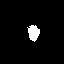}
\end{tabular}
\end{minipage}
	\caption{$\beta$-TCVAE, $\beta=2.0$, MIG: $0.30$, logpx: $-27.15$}
\end{subfigure}\\
\begin{subfigure}{\linewidth}
\centering
\begin{minipage}{0.225\linewidth}
	\centering
	\includegraphics[width=\linewidth]{figures/traversals/beta_2_upperquantile/gt_vs_latent_isa_vae_beta_2.png}
\end{minipage}
\hspace{-1.25cm}
\begin{minipage}{0.6\linewidth}
\centering
\newcolumntype{V}{>{\centering\arraybackslash} m{.07\linewidth} }
\newcolumntype{T}{>{\centering\arraybackslash} m{.035\linewidth} }
\setlength{\tabcolsep}{0.5pt}
\renewcommand{\arraystretch}{1.0}
\begin{tabular}{T V V V V V V V V}
	\rotatebox{90}{\scalebox{.5}{rotation}}&
	\includegraphics[width=\linewidth]{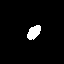}&
	\includegraphics[width=\linewidth]{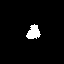}&
	\includegraphics[width=\linewidth]{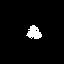}&
	\includegraphics[width=\linewidth]{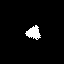}&
	\includegraphics[width=\linewidth]{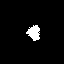}&
	\includegraphics[width=\linewidth]{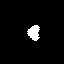}&
	\includegraphics[width=\linewidth]{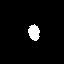}&
	\includegraphics[width=\linewidth]{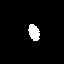}\\
	\rotatebox{90}{\scalebox{.5}{rotation}}&
	\includegraphics[width=\linewidth]{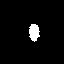}&
	\includegraphics[width=\linewidth]{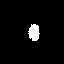}&
	\includegraphics[width=\linewidth]{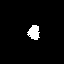}&
	\includegraphics[width=\linewidth]{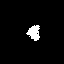}&
	\includegraphics[width=\linewidth]{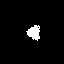}&
	\includegraphics[width=\linewidth]{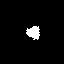}&
	\includegraphics[width=\linewidth]{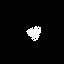}&
	\includegraphics[width=\linewidth]{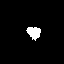}\\
	\rotatebox{90}{\scalebox{.5}{scale}}&
	\includegraphics[width=\linewidth]{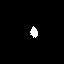}&
	\includegraphics[width=\linewidth]{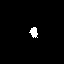}&
	\includegraphics[width=\linewidth]{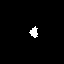}&
	\includegraphics[width=\linewidth]{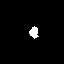}&
	\includegraphics[width=\linewidth]{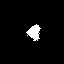}&
	\includegraphics[width=\linewidth]{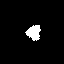}&
	\includegraphics[width=\linewidth]{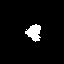}&
	\includegraphics[width=\linewidth]{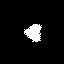}\\
	\rotatebox{90}{\scalebox{.5}{x-pos}}&
	\includegraphics[width=\linewidth]{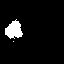}&
	\includegraphics[width=\linewidth]{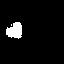}&
	\includegraphics[width=\linewidth]{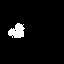}&
	\includegraphics[width=\linewidth]{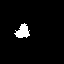}&
	\includegraphics[width=\linewidth]{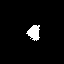}&
	\includegraphics[width=\linewidth]{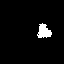}&
	\includegraphics[width=\linewidth]{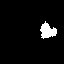}&
	\includegraphics[width=\linewidth]{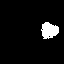}\\
	\rotatebox{90}{\scalebox{.5}{y-pos}}&
	\includegraphics[width=\linewidth]{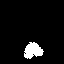}&
	\includegraphics[width=\linewidth]{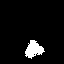}&
	\includegraphics[width=\linewidth]{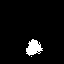}&
	\includegraphics[width=\linewidth]{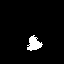}&
	\includegraphics[width=\linewidth]{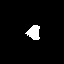}&
	\includegraphics[width=\linewidth]{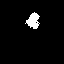}&
	\includegraphics[width=\linewidth]{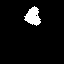}&
	\includegraphics[width=\linewidth]{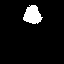}
\end{tabular}
\end{minipage}
	\caption{ISA-VAE, $\beta=2.0$, MIG: $0.41$, logpx: $-25.86$}
\end{subfigure}
\end{center}
\caption{Disentangled representations for models representative for the upper quantile of MIG scores for $\beta=2.0$ for $\beta$-VAE, $\beta$-TCVAE, ISA-VAE and ISA-TCVAE and latent traversals for the heart shape.}
\end{figure*}

%% file: traversal_ellipse_beta3_upperquantile.tex
\begin{figure*}
\begin{center}
\begin{subfigure}{\linewidth}
\centering
\begin{minipage}{0.225\linewidth}
	\centering
	\includegraphics[width=\linewidth]{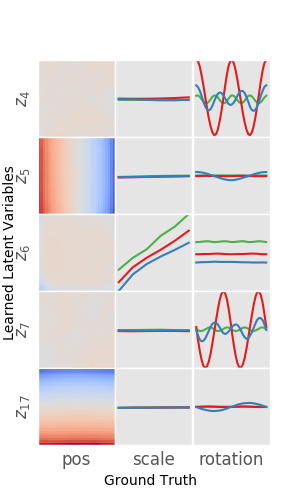}
\end{minipage}
\hspace{-1.25cm}
\begin{minipage}{0.6\linewidth}
\centering
\newcolumntype{V}{>{\centering\arraybackslash} m{.07\linewidth} }
\newcolumntype{T}{>{\centering\arraybackslash} m{.035\linewidth} }
\setlength{\tabcolsep}{0.5pt}
\renewcommand{\arraystretch}{1.0}
\begin{tabular}{T V V V V V V V V}
	\rotatebox{90}{\scalebox{.5}{rotation}}&
	\includegraphics[width=\linewidth]{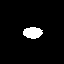}&
	\includegraphics[width=\linewidth]{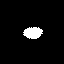}&
	\includegraphics[width=\linewidth]{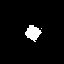}&
	\includegraphics[width=\linewidth]{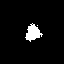}&
	\includegraphics[width=\linewidth]{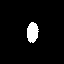}&
	\includegraphics[width=\linewidth]{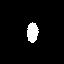}&
	\includegraphics[width=\linewidth]{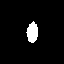}&
	\includegraphics[width=\linewidth]{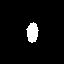}\\
	\rotatebox{90}{\scalebox{.5}{x-pos}}&
	\includegraphics[width=\linewidth]{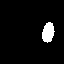}&
	\includegraphics[width=\linewidth]{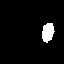}&
	\includegraphics[width=\linewidth]{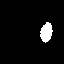}&
	\includegraphics[width=\linewidth]{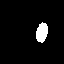}&
	\includegraphics[width=\linewidth]{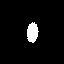}&
	\includegraphics[width=\linewidth]{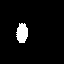}&
	\includegraphics[width=\linewidth]{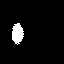}&
	\includegraphics[width=\linewidth]{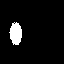}\\
	\rotatebox{90}{\scalebox{.5}{scale}}&
	\includegraphics[width=\linewidth]{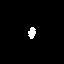}&
	\includegraphics[width=\linewidth]{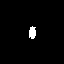}&
	\includegraphics[width=\linewidth]{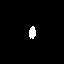}&
	\includegraphics[width=\linewidth]{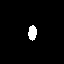}&
	\includegraphics[width=\linewidth]{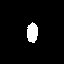}&
	\includegraphics[width=\linewidth]{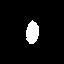}&
	\includegraphics[width=\linewidth]{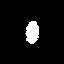}&
	\includegraphics[width=\linewidth]{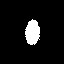}\\
	\rotatebox{90}{\scalebox{.5}{rotation}}&
	\includegraphics[width=\linewidth]{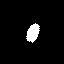}&
	\includegraphics[width=\linewidth]{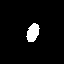}&
	\includegraphics[width=\linewidth]{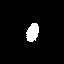}&
	\includegraphics[width=\linewidth]{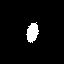}&
	\includegraphics[width=\linewidth]{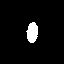}&
	\includegraphics[width=\linewidth]{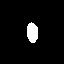}&
	\includegraphics[width=\linewidth]{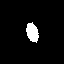}&
	\includegraphics[width=\linewidth]{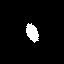}\\
	\rotatebox{90}{\scalebox{.5}{y-pos}}&
	\includegraphics[width=\linewidth]{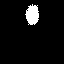}&
	\includegraphics[width=\linewidth]{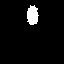}&
	\includegraphics[width=\linewidth]{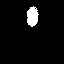}&
	\includegraphics[width=\linewidth]{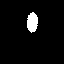}&
	\includegraphics[width=\linewidth]{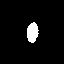}&
	\includegraphics[width=\linewidth]{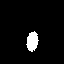}&
	\includegraphics[width=\linewidth]{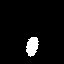}&
	\includegraphics[width=\linewidth]{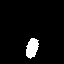}
\end{tabular}
\end{minipage}
	\caption{$\beta$-VAE, $\beta=3.0$, MIG: $0.47, logpx: -33.44$}
\end{subfigure}
\begin{subfigure}{\linewidth}
\centering
\begin{minipage}{0.225\linewidth}
	\centering
	\includegraphics[width=\linewidth]{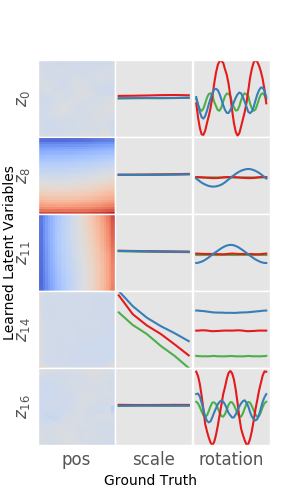}
\end{minipage}
\hspace{-1.25cm}
\begin{minipage}{0.6\linewidth}
\centering
\newcolumntype{V}{>{\centering\arraybackslash} m{.07\linewidth} }
\newcolumntype{T}{>{\centering\arraybackslash} m{.035\linewidth} }
\setlength{\tabcolsep}{0.5pt}
\renewcommand{\arraystretch}{1.0}
\begin{tabular}{T V V V V V V V V}
	\rotatebox{90}{\scalebox{.5}{rotation}}&
	\includegraphics[width=\linewidth]{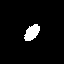}&
	\includegraphics[width=\linewidth]{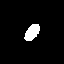}&
	\includegraphics[width=\linewidth]{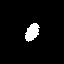}&
	\includegraphics[width=\linewidth]{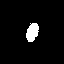}&
	\includegraphics[width=\linewidth]{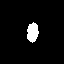}&
	\includegraphics[width=\linewidth]{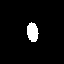}&
	\includegraphics[width=\linewidth]{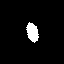}&
	\includegraphics[width=\linewidth]{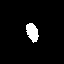}\\
	\rotatebox{90}{\scalebox{.5}{y-pos}}&
	\includegraphics[width=\linewidth]{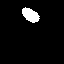}&
	\includegraphics[width=\linewidth]{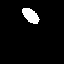}&
	\includegraphics[width=\linewidth]{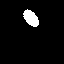}&
	\includegraphics[width=\linewidth]{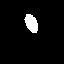}&
	\includegraphics[width=\linewidth]{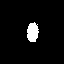}&
	\includegraphics[width=\linewidth]{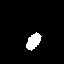}&
	\includegraphics[width=\linewidth]{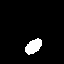}&
	\includegraphics[width=\linewidth]{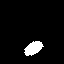}\\
	\rotatebox{90}{\scalebox{.5}{x-pos}}&
	\includegraphics[width=\linewidth]{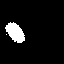}&
	\includegraphics[width=\linewidth]{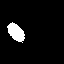}&
	\includegraphics[width=\linewidth]{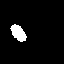}&
	\includegraphics[width=\linewidth]{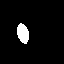}&
	\includegraphics[width=\linewidth]{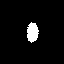}&
	\includegraphics[width=\linewidth]{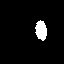}&
	\includegraphics[width=\linewidth]{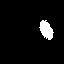}&
	\includegraphics[width=\linewidth]{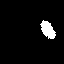}\\
	\rotatebox{90}{\scalebox{.5}{scale}}&
	\includegraphics[width=\linewidth]{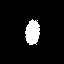}&
	\includegraphics[width=\linewidth]{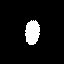}&
	\includegraphics[width=\linewidth]{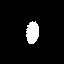}&
	\includegraphics[width=\linewidth]{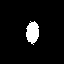}&
	\includegraphics[width=\linewidth]{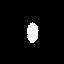}&
	\includegraphics[width=\linewidth]{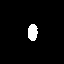}&
	\includegraphics[width=\linewidth]{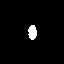}&
	\includegraphics[width=\linewidth]{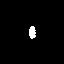}\\
	\rotatebox{90}{\scalebox{.5}{rotation}}&
	\includegraphics[width=\linewidth]{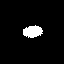}&
	\includegraphics[width=\linewidth]{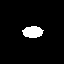}&
	\includegraphics[width=\linewidth]{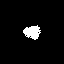}&
	\includegraphics[width=\linewidth]{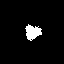}&
	\includegraphics[width=\linewidth]{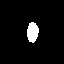}&
	\includegraphics[width=\linewidth]{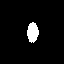}&
	\includegraphics[width=\linewidth]{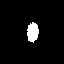}&
	\includegraphics[width=\linewidth]{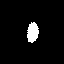}
\end{tabular}
\end{minipage}
	\caption{$\beta$-TCVAE, $\beta=3.0$, MIG: $0.43$, logpx: $-33.40$}
\end{subfigure}\\
\begin{subfigure}{\linewidth}
\centering
\begin{minipage}{0.225\linewidth}
	\centering
	\includegraphics[width=\linewidth]{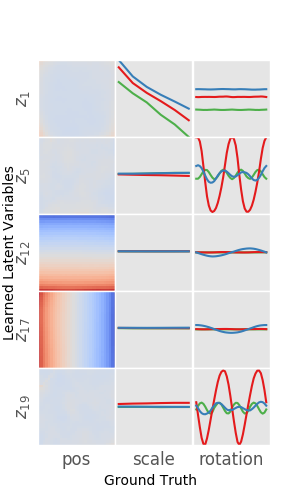}
\end{minipage}
\hspace{-1.25cm}
\begin{minipage}{0.6\linewidth}
\centering
\newcolumntype{V}{>{\centering\arraybackslash} m{.07\linewidth} }
\newcolumntype{T}{>{\centering\arraybackslash} m{.035\linewidth} }
\setlength{\tabcolsep}{0.5pt}
\renewcommand{\arraystretch}{1.0}
\begin{tabular}{T V V V V V V V V}
	\rotatebox{90}{\scalebox{.5}{scale}}&
	\includegraphics[width=\linewidth]{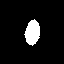}&
	\includegraphics[width=\linewidth]{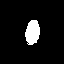}&
	\includegraphics[width=\linewidth]{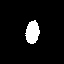}&
	\includegraphics[width=\linewidth]{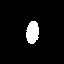}&
	\includegraphics[width=\linewidth]{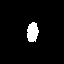}&
	\includegraphics[width=\linewidth]{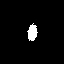}&
	\includegraphics[width=\linewidth]{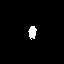}&
	\includegraphics[width=\linewidth]{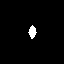}\\
	\rotatebox{90}{\scalebox{.5}{rotation}}&
	\includegraphics[width=\linewidth]{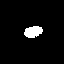}&
	\includegraphics[width=\linewidth]{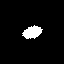}&
	\includegraphics[width=\linewidth]{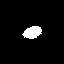}&
	\includegraphics[width=\linewidth]{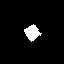}&
	\includegraphics[width=\linewidth]{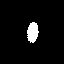}&
	\includegraphics[width=\linewidth]{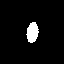}&
	\includegraphics[width=\linewidth]{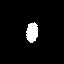}&
	\includegraphics[width=\linewidth]{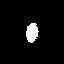}\\
	\rotatebox{90}{\scalebox{.5}{y-pos}}&
	\includegraphics[width=\linewidth]{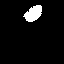}&
	\includegraphics[width=\linewidth]{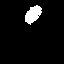}&
	\includegraphics[width=\linewidth]{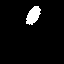}&
	\includegraphics[width=\linewidth]{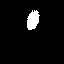}&
	\includegraphics[width=\linewidth]{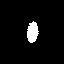}&
	\includegraphics[width=\linewidth]{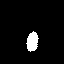}&
	\includegraphics[width=\linewidth]{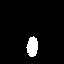}&
	\includegraphics[width=\linewidth]{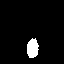}\\
	\rotatebox{90}{\scalebox{.5}{x-pos}}&
	\includegraphics[width=\linewidth]{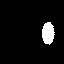}&
	\includegraphics[width=\linewidth]{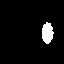}&
	\includegraphics[width=\linewidth]{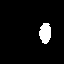}&
	\includegraphics[width=\linewidth]{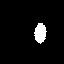}&
	\includegraphics[width=\linewidth]{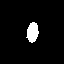}&
	\includegraphics[width=\linewidth]{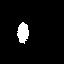}&
	\includegraphics[width=\linewidth]{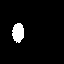}&
	\includegraphics[width=\linewidth]{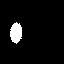}\\
	\rotatebox{90}{\scalebox{.5}{rotation}}&
	\includegraphics[width=\linewidth]{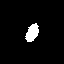}&
	\includegraphics[width=\linewidth]{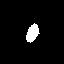}&
	\includegraphics[width=\linewidth]{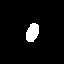}&
	\includegraphics[width=\linewidth]{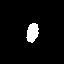}&
	\includegraphics[width=\linewidth]{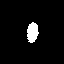}&
	\includegraphics[width=\linewidth]{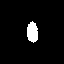}&
	\includegraphics[width=\linewidth]{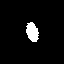}&
	\includegraphics[width=\linewidth]{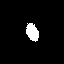}
\end{tabular}
\end{minipage}
	\caption{ISA-VAE, $\beta=3.0$, MIG: $0.48$, logpx: $-32.42$}
\end{subfigure}
\end{center}
\caption{Disentangled representations for models representative for the upper quantile of MIG scores for $\beta=3.0$ for $\beta$-VAE, $\beta$-TCVAE, ISA-VAE and ISA-TCVAE and latent traversals for the ellipse shape.}
\end{figure*}

%% file: traversal_square_beta3_upperquantile.tex
\begin{figure*}
\begin{center}
\begin{subfigure}{\linewidth}
\centering
\begin{minipage}{0.225\linewidth}
	\centering
	\includegraphics[width=\linewidth]{figures/traversals/beta_3_upperquantile/gt_vs_latent_vae_beta_3.png}
\end{minipage}
\hspace{-1.25cm}
\begin{minipage}{0.6\linewidth}
\centering
\newcolumntype{V}{>{\centering\arraybackslash} m{.07\linewidth} }
\newcolumntype{T}{>{\centering\arraybackslash} m{.035\linewidth} }
\setlength{\tabcolsep}{0.5pt}
\renewcommand{\arraystretch}{1.0}
\begin{tabular}{T V V V V V V V V}
	\rotatebox{90}{\scalebox{.5}{rotation}}&
	\includegraphics[width=\linewidth]{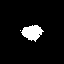}&
	\includegraphics[width=\linewidth]{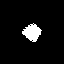}&
	\includegraphics[width=\linewidth]{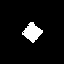}&
	\includegraphics[width=\linewidth]{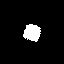}&
	\includegraphics[width=\linewidth]{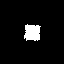}&
	\includegraphics[width=\linewidth]{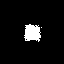}&
	\includegraphics[width=\linewidth]{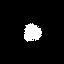}&
	\includegraphics[width=\linewidth]{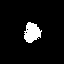}\\
	\rotatebox{90}{\scalebox{.5}{x-pos}}&
	\includegraphics[width=\linewidth]{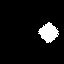}&
	\includegraphics[width=\linewidth]{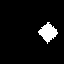}&
	\includegraphics[width=\linewidth]{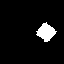}&
	\includegraphics[width=\linewidth]{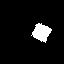}&
	\includegraphics[width=\linewidth]{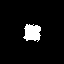}&
	\includegraphics[width=\linewidth]{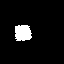}&
	\includegraphics[width=\linewidth]{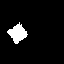}&
	\includegraphics[width=\linewidth]{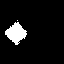}\\
	\rotatebox{90}{\scalebox{.5}{scale}}&
	\includegraphics[width=\linewidth]{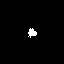}&
	\includegraphics[width=\linewidth]{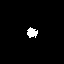}&
	\includegraphics[width=\linewidth]{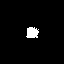}&
	\includegraphics[width=\linewidth]{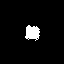}&
	\includegraphics[width=\linewidth]{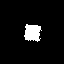}&
	\includegraphics[width=\linewidth]{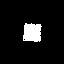}&
	\includegraphics[width=\linewidth]{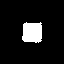}&
	\includegraphics[width=\linewidth]{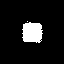}\\
	\rotatebox{90}{\scalebox{.5}{rotation}}&
	\includegraphics[width=\linewidth]{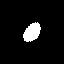}&
	\includegraphics[width=\linewidth]{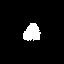}&
	\includegraphics[width=\linewidth]{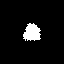}&
	\includegraphics[width=\linewidth]{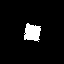}&
	\includegraphics[width=\linewidth]{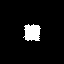}&
	\includegraphics[width=\linewidth]{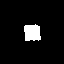}&
	\includegraphics[width=\linewidth]{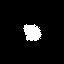}&
	\includegraphics[width=\linewidth]{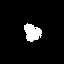}\\
	\rotatebox{90}{\scalebox{.5}{y-pos}}&
	\includegraphics[width=\linewidth]{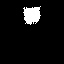}&
	\includegraphics[width=\linewidth]{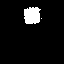}&
	\includegraphics[width=\linewidth]{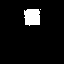}&
	\includegraphics[width=\linewidth]{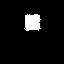}&
	\includegraphics[width=\linewidth]{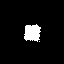}&
	\includegraphics[width=\linewidth]{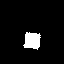}&
	\includegraphics[width=\linewidth]{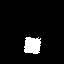}&
	\includegraphics[width=\linewidth]{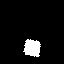}
\end{tabular}
\end{minipage}
	\caption{$\beta$-VAE, $\beta=3.0$, MIG: $0.47, logpx: -33.44$}
\end{subfigure}
\begin{subfigure}{\linewidth}
\centering
\begin{minipage}{0.225\linewidth}
	\centering
	\includegraphics[width=\linewidth]{figures/traversals/beta_3_upperquantile/gt_vs_latent_tcvae_beta_3.png}
\end{minipage}
\hspace{-1.25cm}
\begin{minipage}{0.6\linewidth}
\centering
\newcolumntype{V}{>{\centering\arraybackslash} m{.07\linewidth} }
\newcolumntype{T}{>{\centering\arraybackslash} m{.035\linewidth} }
\setlength{\tabcolsep}{0.5pt}
\renewcommand{\arraystretch}{1.0}
\begin{tabular}{T V V V V V V V V}
	\rotatebox{90}{\scalebox{.5}{rotation}}&
	\includegraphics[width=\linewidth]{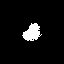}&
	\includegraphics[width=\linewidth]{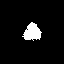}&
	\includegraphics[width=\linewidth]{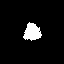}&
	\includegraphics[width=\linewidth]{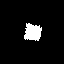}&
	\includegraphics[width=\linewidth]{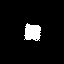}&
	\includegraphics[width=\linewidth]{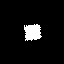}&
	\includegraphics[width=\linewidth]{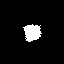}&
	\includegraphics[width=\linewidth]{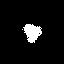}\\
	\rotatebox{90}{\scalebox{.5}{y-pos}}&
	\includegraphics[width=\linewidth]{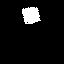}&
	\includegraphics[width=\linewidth]{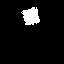}&
	\includegraphics[width=\linewidth]{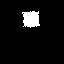}&
	\includegraphics[width=\linewidth]{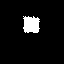}&
	\includegraphics[width=\linewidth]{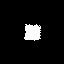}&
	\includegraphics[width=\linewidth]{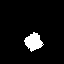}&
	\includegraphics[width=\linewidth]{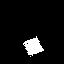}&
	\includegraphics[width=\linewidth]{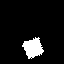}\\
	\rotatebox{90}{\scalebox{.5}{x-pos}}&
	\includegraphics[width=\linewidth]{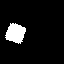}&
	\includegraphics[width=\linewidth]{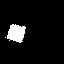}&
	\includegraphics[width=\linewidth]{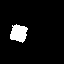}&
	\includegraphics[width=\linewidth]{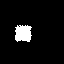}&
	\includegraphics[width=\linewidth]{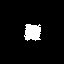}&
	\includegraphics[width=\linewidth]{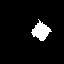}&
	\includegraphics[width=\linewidth]{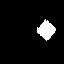}&
	\includegraphics[width=\linewidth]{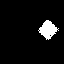}\\
	\rotatebox{90}{\scalebox{.5}{scale}}&
	\includegraphics[width=\linewidth]{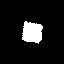}&
	\includegraphics[width=\linewidth]{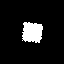}&
	\includegraphics[width=\linewidth]{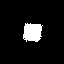}&
	\includegraphics[width=\linewidth]{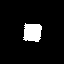}&
	\includegraphics[width=\linewidth]{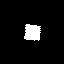}&
	\includegraphics[width=\linewidth]{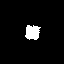}&
	\includegraphics[width=\linewidth]{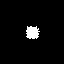}&
	\includegraphics[width=\linewidth]{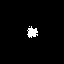}\\
	\rotatebox{90}{\scalebox{.5}{rotation}}&
	\includegraphics[width=\linewidth]{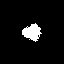}&
	\includegraphics[width=\linewidth]{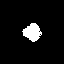}&
	\includegraphics[width=\linewidth]{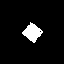}&
	\includegraphics[width=\linewidth]{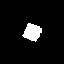}&
	\includegraphics[width=\linewidth]{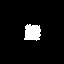}&
	\includegraphics[width=\linewidth]{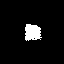}&
	\includegraphics[width=\linewidth]{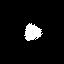}&
	\includegraphics[width=\linewidth]{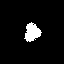}
\end{tabular}
\end{minipage}
	\caption{$\beta$-TCVAE, $\beta=3.0$, MIG: $0.43$, logpx: $-33.40$}
\end{subfigure}\\
\begin{subfigure}{\linewidth}
\centering
\begin{minipage}{0.225\linewidth}
	\centering
	\includegraphics[width=\linewidth]{figures/traversals/beta_3_upperquantile/gt_vs_latent_isa_vae_beta_3.png}
\end{minipage}
\hspace{-1.25cm}
\begin{minipage}{0.6\linewidth}
\centering
\newcolumntype{V}{>{\centering\arraybackslash} m{.07\linewidth} }
\newcolumntype{T}{>{\centering\arraybackslash} m{.035\linewidth} }
\setlength{\tabcolsep}{0.5pt}
\renewcommand{\arraystretch}{1.0}
\begin{tabular}{T V V V V V V V V}
	\rotatebox{90}{\scalebox{.5}{scale}}&
	\includegraphics[width=\linewidth]{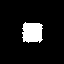}&
	\includegraphics[width=\linewidth]{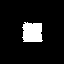}&
	\includegraphics[width=\linewidth]{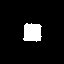}&
	\includegraphics[width=\linewidth]{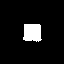}&
	\includegraphics[width=\linewidth]{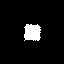}&
	\includegraphics[width=\linewidth]{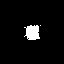}&
	\includegraphics[width=\linewidth]{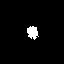}&
	\includegraphics[width=\linewidth]{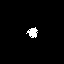}\\
	\rotatebox{90}{\scalebox{.5}{rotation}}&
	\includegraphics[width=\linewidth]{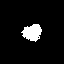}&
	\includegraphics[width=\linewidth]{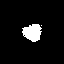}&
	\includegraphics[width=\linewidth]{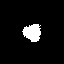}&
	\includegraphics[width=\linewidth]{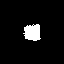}&
	\includegraphics[width=\linewidth]{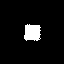}&
	\includegraphics[width=\linewidth]{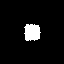}&
	\includegraphics[width=\linewidth]{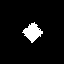}&
	\includegraphics[width=\linewidth]{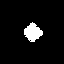}\\
	\rotatebox{90}{\scalebox{.5}{y-pos}}&
	\includegraphics[width=\linewidth]{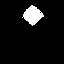}&
	\includegraphics[width=\linewidth]{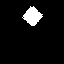}&
	\includegraphics[width=\linewidth]{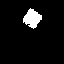}&
	\includegraphics[width=\linewidth]{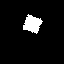}&
	\includegraphics[width=\linewidth]{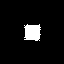}&
	\includegraphics[width=\linewidth]{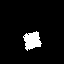}&
	\includegraphics[width=\linewidth]{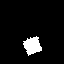}&
	\includegraphics[width=\linewidth]{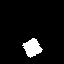}\\
	\rotatebox{90}{\scalebox{.5}{x-pos}}&
	\includegraphics[width=\linewidth]{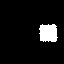}&
	\includegraphics[width=\linewidth]{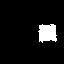}&
	\includegraphics[width=\linewidth]{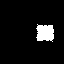}&
	\includegraphics[width=\linewidth]{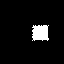}&
	\includegraphics[width=\linewidth]{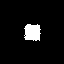}&
	\includegraphics[width=\linewidth]{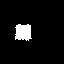}&
	\includegraphics[width=\linewidth]{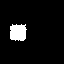}&
	\includegraphics[width=\linewidth]{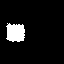}\\
	\rotatebox{90}{\scalebox{.5}{rotation}}&
	\includegraphics[width=\linewidth]{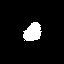}&
	\includegraphics[width=\linewidth]{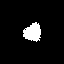}&
	\includegraphics[width=\linewidth]{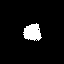}&
	\includegraphics[width=\linewidth]{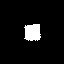}&
	\includegraphics[width=\linewidth]{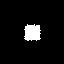}&
	\includegraphics[width=\linewidth]{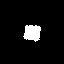}&
	\includegraphics[width=\linewidth]{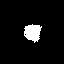}&
	\includegraphics[width=\linewidth]{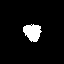}
\end{tabular}
\end{minipage}
	\caption{ISA-VAE, $\beta=3.0$, MIG: $0.48$, logpx: $-32.42$}
\end{subfigure}
\end{center}
\caption{Disentangled representations for models representative for the upper quantile of MIG scores for $\beta=3.0$ for $\beta$-VAE, $\beta$-TCVAE, ISA-VAE and ISA-TCVAE and latent traversals for the square shape.}
\end{figure*}

%% file: traversal_heart_beta3_upperquantile.tex
\begin{figure*}
\begin{center}
\begin{subfigure}{\linewidth}
\centering
\begin{minipage}{0.225\linewidth}
	\centering
	\includegraphics[width=\linewidth]{figures/traversals/beta_3_upperquantile/gt_vs_latent_vae_beta_3.png}
\end{minipage}
\hspace{-1.25cm}
\begin{minipage}{0.6\linewidth}
\centering
\newcolumntype{V}{>{\centering\arraybackslash} m{.07\linewidth} }
\newcolumntype{T}{>{\centering\arraybackslash} m{.035\linewidth} }
\setlength{\tabcolsep}{0.5pt}
\renewcommand{\arraystretch}{1.0}
\begin{tabular}{T V V V V V V V V}
	\rotatebox{90}{\scalebox{.5}{rotation}}&
	\includegraphics[width=\linewidth]{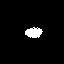}&
	\includegraphics[width=\linewidth]{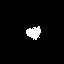}&
	\includegraphics[width=\linewidth]{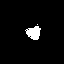}&
	\includegraphics[width=\linewidth]{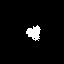}&
	\includegraphics[width=\linewidth]{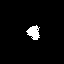}&
	\includegraphics[width=\linewidth]{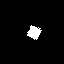}&
	\includegraphics[width=\linewidth]{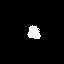}&
	\includegraphics[width=\linewidth]{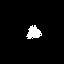}\\
	\rotatebox{90}{\scalebox{.5}{x-pos}}&
	\includegraphics[width=\linewidth]{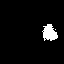}&
	\includegraphics[width=\linewidth]{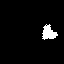}&
	\includegraphics[width=\linewidth]{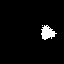}&
	\includegraphics[width=\linewidth]{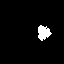}&
	\includegraphics[width=\linewidth]{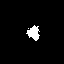}&
	\includegraphics[width=\linewidth]{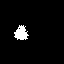}&
	\includegraphics[width=\linewidth]{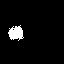}&
	\includegraphics[width=\linewidth]{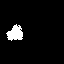}\\
	\rotatebox{90}{\scalebox{.5}{scale}}&
	\includegraphics[width=\linewidth]{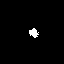}&
	\includegraphics[width=\linewidth]{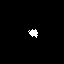}&
	\includegraphics[width=\linewidth]{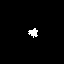}&
	\includegraphics[width=\linewidth]{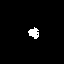}&
	\includegraphics[width=\linewidth]{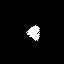}&
	\includegraphics[width=\linewidth]{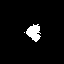}&
	\includegraphics[width=\linewidth]{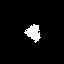}&
	\includegraphics[width=\linewidth]{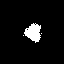}\\
	\rotatebox{90}{\scalebox{.5}{rotation}}&
	\includegraphics[width=\linewidth]{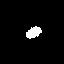}&
	\includegraphics[width=\linewidth]{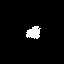}&
	\includegraphics[width=\linewidth]{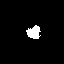}&
	\includegraphics[width=\linewidth]{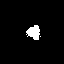}&
	\includegraphics[width=\linewidth]{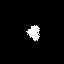}&
	\includegraphics[width=\linewidth]{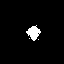}&
	\includegraphics[width=\linewidth]{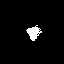}&
	\includegraphics[width=\linewidth]{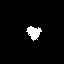}\\
	\rotatebox{90}{\scalebox{.5}{y-pos}}&
	\includegraphics[width=\linewidth]{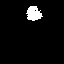}&
	\includegraphics[width=\linewidth]{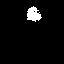}&
	\includegraphics[width=\linewidth]{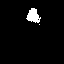}&
	\includegraphics[width=\linewidth]{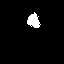}&
	\includegraphics[width=\linewidth]{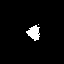}&
	\includegraphics[width=\linewidth]{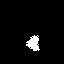}&
	\includegraphics[width=\linewidth]{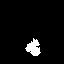}&
	\includegraphics[width=\linewidth]{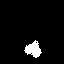}
\end{tabular}
\end{minipage}
	\caption{$\beta$-VAE, $\beta=3.0$, MIG: $0.47, logpx: -33.44$}
\end{subfigure}
\begin{subfigure}{\linewidth}
\centering
\begin{minipage}{0.225\linewidth}
	\centering
	\includegraphics[width=\linewidth]{figures/traversals/beta_3_upperquantile/gt_vs_latent_tcvae_beta_3.png}
\end{minipage}
\hspace{-1.25cm}
\begin{minipage}{0.6\linewidth}
\centering
\newcolumntype{V}{>{\centering\arraybackslash} m{.07\linewidth} }
\newcolumntype{T}{>{\centering\arraybackslash} m{.035\linewidth} }
\setlength{\tabcolsep}{0.5pt}
\renewcommand{\arraystretch}{1.0}
\begin{tabular}{T V V V V V V V V}
	\rotatebox{90}{\scalebox{.5}{rotation}}&
	\includegraphics[width=\linewidth]{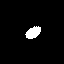}&
	\includegraphics[width=\linewidth]{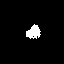}&
	\includegraphics[width=\linewidth]{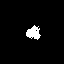}&
	\includegraphics[width=\linewidth]{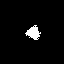}&
	\includegraphics[width=\linewidth]{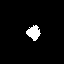}&
	\includegraphics[width=\linewidth]{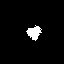}&
	\includegraphics[width=\linewidth]{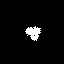}&
	\includegraphics[width=\linewidth]{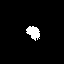}\\
	\rotatebox{90}{\scalebox{.5}{y-pos}}&
	\includegraphics[width=\linewidth]{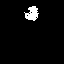}&
	\includegraphics[width=\linewidth]{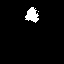}&
	\includegraphics[width=\linewidth]{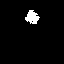}&
	\includegraphics[width=\linewidth]{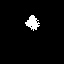}&
	\includegraphics[width=\linewidth]{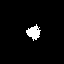}&
	\includegraphics[width=\linewidth]{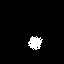}&
	\includegraphics[width=\linewidth]{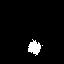}&
	\includegraphics[width=\linewidth]{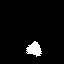}\\
	\rotatebox{90}{\scalebox{.5}{x-pos}}&
	\includegraphics[width=\linewidth]{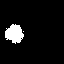}&
	\includegraphics[width=\linewidth]{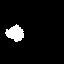}&
	\includegraphics[width=\linewidth]{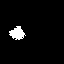}&
	\includegraphics[width=\linewidth]{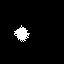}&
	\includegraphics[width=\linewidth]{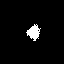}&
	\includegraphics[width=\linewidth]{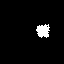}&
	\includegraphics[width=\linewidth]{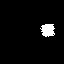}&
	\includegraphics[width=\linewidth]{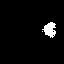}\\
	\rotatebox{90}{\scalebox{.5}{scale}}&
	\includegraphics[width=\linewidth]{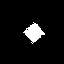}&
	\includegraphics[width=\linewidth]{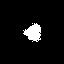}&
	\includegraphics[width=\linewidth]{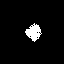}&
	\includegraphics[width=\linewidth]{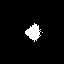}&
	\includegraphics[width=\linewidth]{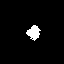}&
	\includegraphics[width=\linewidth]{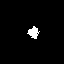}&
	\includegraphics[width=\linewidth]{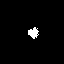}&
	\includegraphics[width=\linewidth]{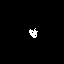}\\
	\rotatebox{90}{\scalebox{.5}{rotation}}&
	\includegraphics[width=\linewidth]{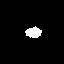}&
	\includegraphics[width=\linewidth]{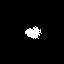}&
	\includegraphics[width=\linewidth]{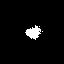}&
	\includegraphics[width=\linewidth]{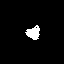}&
	\includegraphics[width=\linewidth]{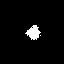}&
	\includegraphics[width=\linewidth]{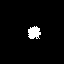}&
	\includegraphics[width=\linewidth]{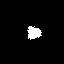}&
	\includegraphics[width=\linewidth]{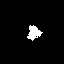}
\end{tabular}
\end{minipage}
	\caption{$\beta$-TCVAE, $\beta=3.0$, MIG: $0.43$, logpx: $-33.40$}
\end{subfigure}\\
\begin{subfigure}{\linewidth}
\centering
\begin{minipage}{0.225\linewidth}
	\centering
	\includegraphics[width=\linewidth]{figures/traversals/beta_3_upperquantile/gt_vs_latent_isa_vae_beta_3.png}
\end{minipage}
\hspace{-1.25cm}
\begin{minipage}{0.6\linewidth}
\centering
\newcolumntype{V}{>{\centering\arraybackslash} m{.07\linewidth} }
\newcolumntype{T}{>{\centering\arraybackslash} m{.035\linewidth} }
\setlength{\tabcolsep}{0.5pt}
\renewcommand{\arraystretch}{1.0}
\begin{tabular}{T V V V V V V V V}
	\rotatebox{90}{\scalebox{.5}{scale}}&
	\includegraphics[width=\linewidth]{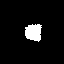}&
	\includegraphics[width=\linewidth]{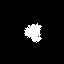}&
	\includegraphics[width=\linewidth]{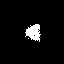}&
	\includegraphics[width=\linewidth]{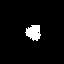}&
	\includegraphics[width=\linewidth]{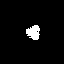}&
	\includegraphics[width=\linewidth]{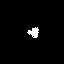}&
	\includegraphics[width=\linewidth]{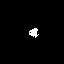}&
	\includegraphics[width=\linewidth]{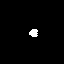}\\
	\rotatebox{90}{\scalebox{.5}{rotation}}&
	\includegraphics[width=\linewidth]{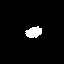}&
	\includegraphics[width=\linewidth]{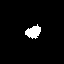}&
	\includegraphics[width=\linewidth]{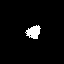}&
	\includegraphics[width=\linewidth]{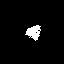}&
	\includegraphics[width=\linewidth]{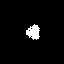}&
	\includegraphics[width=\linewidth]{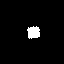}&
	\includegraphics[width=\linewidth]{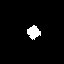}&
	\includegraphics[width=\linewidth]{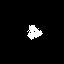}\\
	\rotatebox{90}{\scalebox{.5}{y-pos}}&
	\includegraphics[width=\linewidth]{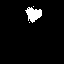}&
	\includegraphics[width=\linewidth]{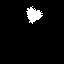}&
	\includegraphics[width=\linewidth]{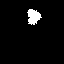}&
	\includegraphics[width=\linewidth]{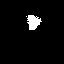}&
	\includegraphics[width=\linewidth]{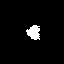}&
	\includegraphics[width=\linewidth]{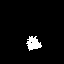}&
	\includegraphics[width=\linewidth]{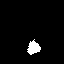}&
	\includegraphics[width=\linewidth]{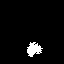}\\
	\rotatebox{90}{\scalebox{.5}{x-pos}}&
	\includegraphics[width=\linewidth]{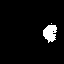}&
	\includegraphics[width=\linewidth]{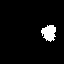}&
	\includegraphics[width=\linewidth]{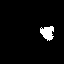}&
	\includegraphics[width=\linewidth]{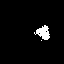}&
	\includegraphics[width=\linewidth]{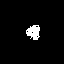}&
	\includegraphics[width=\linewidth]{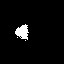}&
	\includegraphics[width=\linewidth]{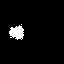}&
	\includegraphics[width=\linewidth]{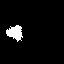}\\
	\rotatebox{90}{\scalebox{.5}{rotation}}&
	\includegraphics[width=\linewidth]{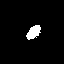}&
	\includegraphics[width=\linewidth]{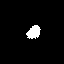}&
	\includegraphics[width=\linewidth]{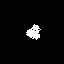}&
	\includegraphics[width=\linewidth]{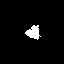}&
	\includegraphics[width=\linewidth]{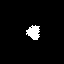}&
	\includegraphics[width=\linewidth]{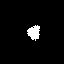}&
	\includegraphics[width=\linewidth]{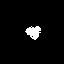}&
	\includegraphics[width=\linewidth]{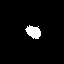}
\end{tabular}
\end{minipage}
	\caption{ISA-VAE, $\beta=3.0$, MIG: $0.48$, logpx: $-32.42$}
\end{subfigure}
\end{center}
\caption{Disentangled representations for models representative for the upper quantile of MIG scores for $\beta=3.0$ for $\beta$-VAE, $\beta$-TCVAE, ISA-VAE and ISA-TCVAE and latent traversals for the heart shape.}
\end{figure*}

%% file: supplement.tex
\section{Toy examples showing biases in Variational Inference and $\beta$-Variational Inference}

This section provides the details of the toy examples that reveal the biases in variational methods.

First we will consider the factor analysis model showing that Variational Inference (VI) breaks the degeneracy of the maximum-likelihood solution to 1) discover orthogonal weights that lie in the PCA directions, 2) prune out extra components in over-complete factor analysis models, even though there are solutions with the same likelihood that preserve all components. We also show that in these examples the $\beta$-VI returns identical model fits to VI regardless of the setting of $\beta$.

Second, we consider an over-complete ICA model and initialize using the true model. We show that 1) VI is biased away from the true component directions towards more orthogonal directions, and 2) $\beta$-VI with a modest setting of $\beta=5$ prunes away one of the components and finds orthogonal directions for the other two. That is, it finds a disentangled representation, but one which does not reflect the underlying components.  

\subsection{Background}

The $\beta$-VAE optimizes the modified free-energy, $\mathcal{F}_{\beta}(q(z_{1:N}),\theta)$, with respect to the parameters $\theta$ and the variational approximation $q(z_{1:N})$,
\begin{align}
\mathcal{F}_{\beta}(q(z_{1:N}),\theta) =&\mathbb{E}_{q(z_{1:N})}(\log p(x_{1:N}|z_{1:N},\theta)) \nonumber\\
& - \beta \mathrm{KL}(q(z_{1:N})||p(z_{1:N})).
\end{align}
Consider the case where $M = \frac{1}{\beta}$ is a positive integer, $M \in \mathbb{N}$, we then have
\begin{align}
\mathcal{F}_{\beta}(q(z_{1:N}),\theta) =& \sum_{n=1}^N \Big[ \mathbb{E}_{q(z_{n})}( M(\beta) \log p(x_{n}|z_{n},\theta))\nonumber\\
&- \mathrm{KL}(q(z_{n})||p(z_{n})) \Big]\nonumber
\end{align}
In this case, the $\beta$-VAE can be thought of as attaching $M$ replicated observations to each latent variable $z_n$ and then running standard variational inference on the new replicated dataset. This can equivalently be thought of as raising each likelihood $p(x_n|z_n,\theta)$ to the power $M$.

Now in real applications $\beta$ will be set to a value that is greater than one. In this case, the effect of $\beta$ is the opposite: it is to reduce the number of effective data points per latent variable to be less than one $M<1$. Or equivalently we raise each likelihood term to a power $M$ that is less than one. Standard VI is then run on these modified data (e.g.~via joint optimization of $q$ and $\theta$). 

Although this view is mathematically straightforward, the perspective of the $\beta$-VAE i) modifying the dataset, and ii) applying standard VI, is useful as it will allow us to derive optimal solutions for the variational distribution $q(z)$ in simple cases like the factor analysis model considered next. 

\subsection{Factor analysis}

Consider the factor analysis generative model. Let $\mathbf{x} \in \mathbb{R}^L$ and $\mathbf{z} \in \mathbb{R}^K$.  
\begin{align}
& \text{for} \;\; n=1...N \nonumber \\
&\;\; \;\;\;\mathbf{z}_n \sim \mathcal{N}(\mathbf{0},\mathrm{I}),\\
& \;\; \;\;\;\mathbf{x}_n \sim \mathcal{N}(W \mathbf{z}_n,D) \;\; \text{where} \;\; D = \text{diag}(\mathrm{[\sigma^2_1,...,\sigma^2_D]})\nonumber
\end{align}

The true posterior is a Gaussian $p(\mathbf{z}_n | \mathbf{x}_n,\theta) = \mathcal{N}(\mathbf{z}; \mu_{\mathbf{z}|\mathbf{x}},\Sigma_{\mathbf{z}|\mathbf{x}})$ where
\begin{align}
\mu_{\mathbf{z}|\mathbf{x}}  &= \Sigma_{\mathbf{z}|\mathbf{x}} W^{\top}D^{-1} \mathbf{x} \\
\;\; \text{and} \;\;
\Sigma_{\mathbf{z}|\mathbf{x}}  &= (W^{\top}D^{-1} W + \mathrm{I})^{-1}. \nonumber
\end{align}

The true log-likelihood of the parameters is
\begin{align}
\log p(\mathbf{x}_{1:N} | \theta) &= &&\sum_{n=1}^N \log \mathcal{N}(\mathbf{x}_n,\mathbf{0},WW^{\top} + D)\nonumber\\
& = && - \frac{N}{2} \log \mathrm{det} (2 \pi (WW^{\top} + D)) \nonumber\\
&&& - \frac{1}{2} \sum_{n=1}^N \mathbf{x}^{\top}_n (WW^{\top} + D)^{-1} \mathbf{x}_n \nonumber \\
& = && - \frac{1}{2} N \Big [ \log \mathrm{det} (2 \pi (W W^{\top} + D)) \nonumber\\
&&&+ \mathrm{trace} ( (W W^{\top} + D)^{-1} (\mu_{\mathbf{x}} \mu_\mathbf{x}^{\top} + \Sigma_\mathbf{x})) \Big]\nonumber
\end{align}

Here we have defined the empirical mean and covariance of the observations $\mu_\mathbf{x} = \frac{1}{N} \sum_{n=1}^N \mathbf{x}_n$ and $\Sigma_\mathbf{x} = \frac{1}{N} \sum_{n=1}^N (\mathbf{x}_n-\mu_\mathbf{x})(\mathbf{x}_n-\mu_\mathbf{x})^{\top}$ i.e. the sufficient statistics. 

The likelihood is invariant under orthogonal transformations of the latent variables: $\mathbf{z}' = R \mathbf{z}$ where $RR^\top = \mathrm{I}$. 

Interpreting $\beta$-VI as running VI in a modified generative model  (see previous section) we have the new generative process
\begin{align}
& \text{for} \;\; n=1...N \nonumber \\
&\;\; \;\;\;\mathbf{z}_n \sim \mathcal{N}(\mathbf{z}_n;\mathbf{0},\mathrm{I}),\nonumber\\
& \;\; \;\;\;  \text{for} \;\; m=1...M(\beta)\nonumber \\
& \;\; \;\;\; \;\; \;\;\; \mathbf{x}_{n,m} \sim \mathcal{N}(W \mathbf{z}_n,D) \;\; \text{where} \;\; D = \text{diag}(\mathrm{[\sigma^2_1,...,\sigma^2_D]})\nonumber
\end{align}
We now observe data and set $\mathbf{x}_{n,m} = \mathbf{x}_{n}$.

The  posterior is again Gaussian $p(\mathbf{z}_n | \mathbf{x}_n,\theta,M(\beta)) = \mathcal{N}(\mathbf{z}_n; \tilde{\mu}_{\mathbf{z}|\mathbf{x}}(\beta,n),\tilde{\Sigma}_{\mathbf{z}|\mathbf{x}}(\beta))$ where
\begin{align}
\tilde{\mu}_{\mathbf{z}|\mathbf{x}}(\beta,n)  = \tilde{\Sigma}_{\mathbf{z}|\mathbf{x}}^{-1}(\beta) M(\beta) W^{\top}D^{-1} \mathbf{x}_n\nonumber\\
\text{and} \;\;
\tilde{\Sigma}_{\mathbf{z}|\mathbf{x}}(\beta)  = (M(\beta) W^{\top}D^{-1} W + \mathrm{I})^{-1} \nonumber
\end{align}
Here we have taken care to explicitly reveal all of the direct dependencies on $\beta$.

Mean-field variational inference, $q(\mathbf{z}_n) = \prod_k q_{n,k}(z_{k,d})$, will return a diagonal Gaussian approximation to the true posterior with the same mean and matching diagonal precision,
\begin{align}
q(\mathbf{z}_n | \mathbf{x}_n,\theta,M(\beta)) = \mathcal{N}\left(\mathbf{z}_n; \tilde{\mu}_{\mathbf{z}|\mathbf{x}}(\beta,n),\Sigma_{q}(\beta)\right) \, ,\nonumber\\
\text{where} \;\; \Sigma_{q}^{-1}(\beta) = \mathrm{diag} \left( \tilde{\Sigma}_{\mathbf{z}|\mathbf{x}}^{-1}(\beta) \right)\nonumber
\end{align}
We notice that the posterior mean is a linear combination of the observations $\tilde{\mu}_{\mathbf{z}|\mathbf{x}}(\beta,n)  = R(\beta) \mathbf{x}_n$ where $R(\beta) = \tilde{\Sigma}_{\mathbf{z}|\mathbf{x}}(\beta) M(\beta) W^{\top}D^{-1}$ are recognition weights. Notice that the recognition weights and the posterior variances are the same for all data points: they do not depend on $n$. The free-energy is then
\begin{align}
\mathcal{F}(q,\theta,\beta) & =  \mathbb{E}_{q(z)}(\log p(x|z)) - \mathrm{KL}(q(z)|p(z)) \nonumber
\end{align}
with the reconstruction term being 
\begin{align}
&\mathbb{E}_{q(z)}(\log p(x|z)) = \nonumber\\
= & - \frac{1}{2 \beta} \sum_{n=1}^N \mathbf{x}^{\top}_n (D^{-1} - 2 R^{\top}W^\top D^{-1} \nonumber\\
&+ R^{\top}W^\top D^{-1} W R ) \mathbf{x}_n \nonumber \\ 
&- \frac{N}{2 \beta} \log \det(2 \pi D) - \frac{N}{2 \beta} \mathrm{trace}(W^\top D^{-1}\Sigma_q) \nonumber \\
= &  - \frac{N}{2 \beta} \biggr ( \mathrm{trace} \Big ( (D^{-1} - 2 R^{\top}W^\top D^{-1} \nonumber\\
&+ R^{\top}W^\top D^{-1} W R ) (\Sigma_\mathbf{x} + \mu_\mathbf{x}\mu_\mathbf{x}^\top) \Big)\nonumber\\
& + \log \det(2 \pi D) + \mathrm{trace}(W^\top D^{-1}W\Sigma_q) \biggr)
\end{align}
and the KL or regularization term being
\begin{align}
&\mathrm{KL}(q(z)|p(z)) = \nonumber\\
= & - \frac{NK}{2} -\frac{N}{2} \log \det (\Sigma_q) + \frac{N}{2} \mathrm{trace} (\Sigma_q) \nonumber\\
&+ \frac{1}{2} \sum_{n=1}^N \mathbf{x}_n^\top R^\top R \mathbf{x}_n \nonumber\\
= & -\frac{N}{2} \Big ( K + \log \det (\Sigma_q) - \mathrm{trace} (\Sigma_q) \nonumber\\
&- \mathrm{trace}(  R^\top R (\Sigma_\mathbf{x} + \mu_\mathbf{x}\mu_\mathbf{x}^\top) ) \Big).\nonumber
\end{align}

We will now consider the objective functions and the posterior distributions in several cases to reason about the parameter estimates arising from the methods above.

\subsection{Experiment 1: mean field VI applied to factor analysis yields the PCA directions}

Consider the situation where we know a maximum likelihood solution of the weights $W_{\text{ML}}$. For simplicity we select the solution $W_{\text{ML}}$ which has orthogonal weights in the observation space. We then rotate this solution by an amount $\theta$ so that $W'_{\text{ML}} = R(\theta)  W_{\text{ML}}$. The resulting weights are no longer orthogonal (assuming the rotation is not an integer multiple of $\pi/2$). We compute the log-likelihood (which will not change) and the free-energy (which will change) and plot the true and approximate posterior covariance (which does not depend on the datapoint value $x_n$). 

First here are the weights are aligned with the true ones. The log-likelihood  and the free-energy take the same value of -17.82 nats.

\begin{center}
\includegraphics[width=\linewidth]{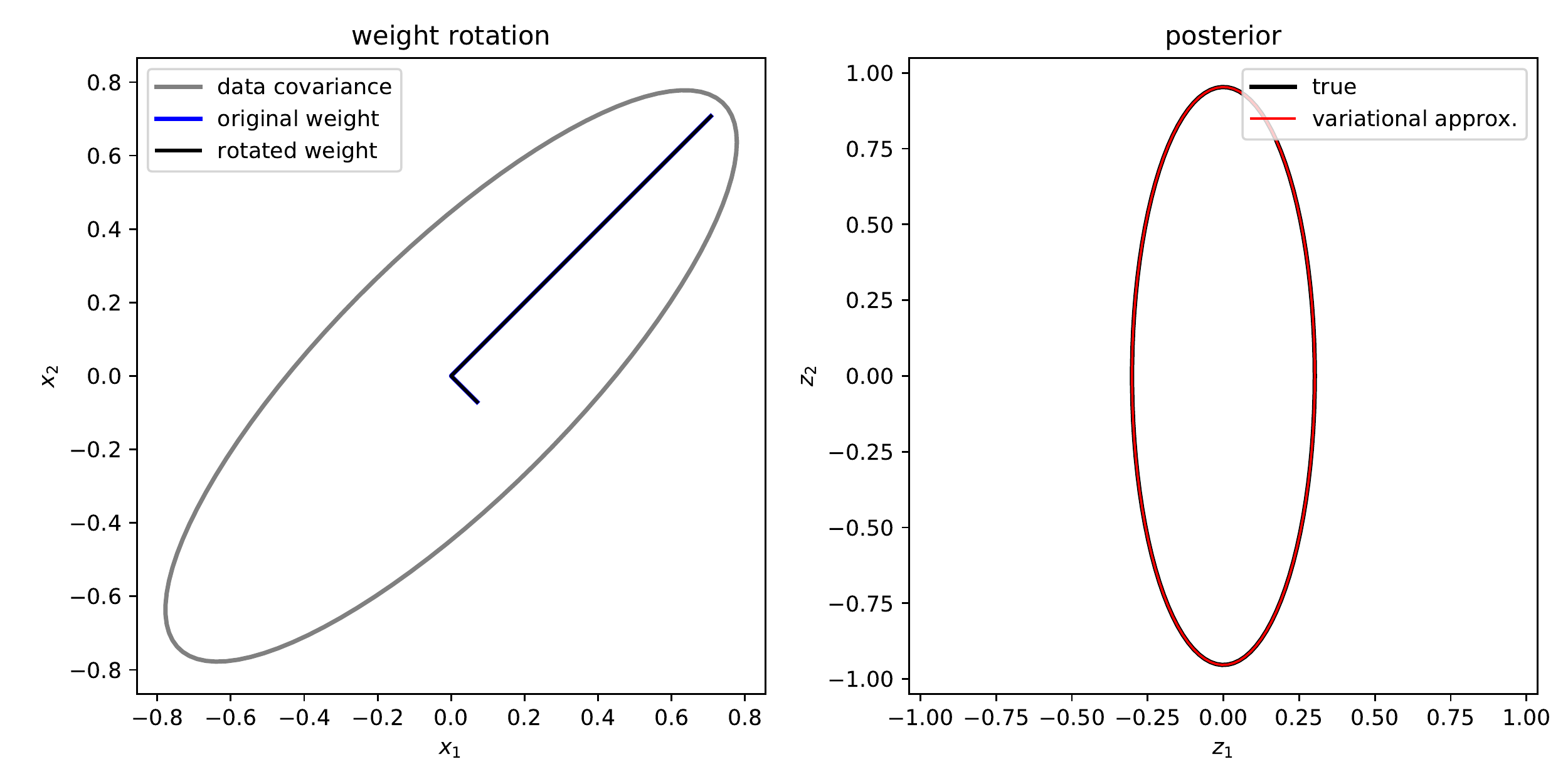}
\end{center}

Second, here are the weights rotated $\pi/4$ and the log-likelihood is -17.82 nats and the free-energy -57.16 nats.

\begin{center}
\includegraphics[width=\linewidth]{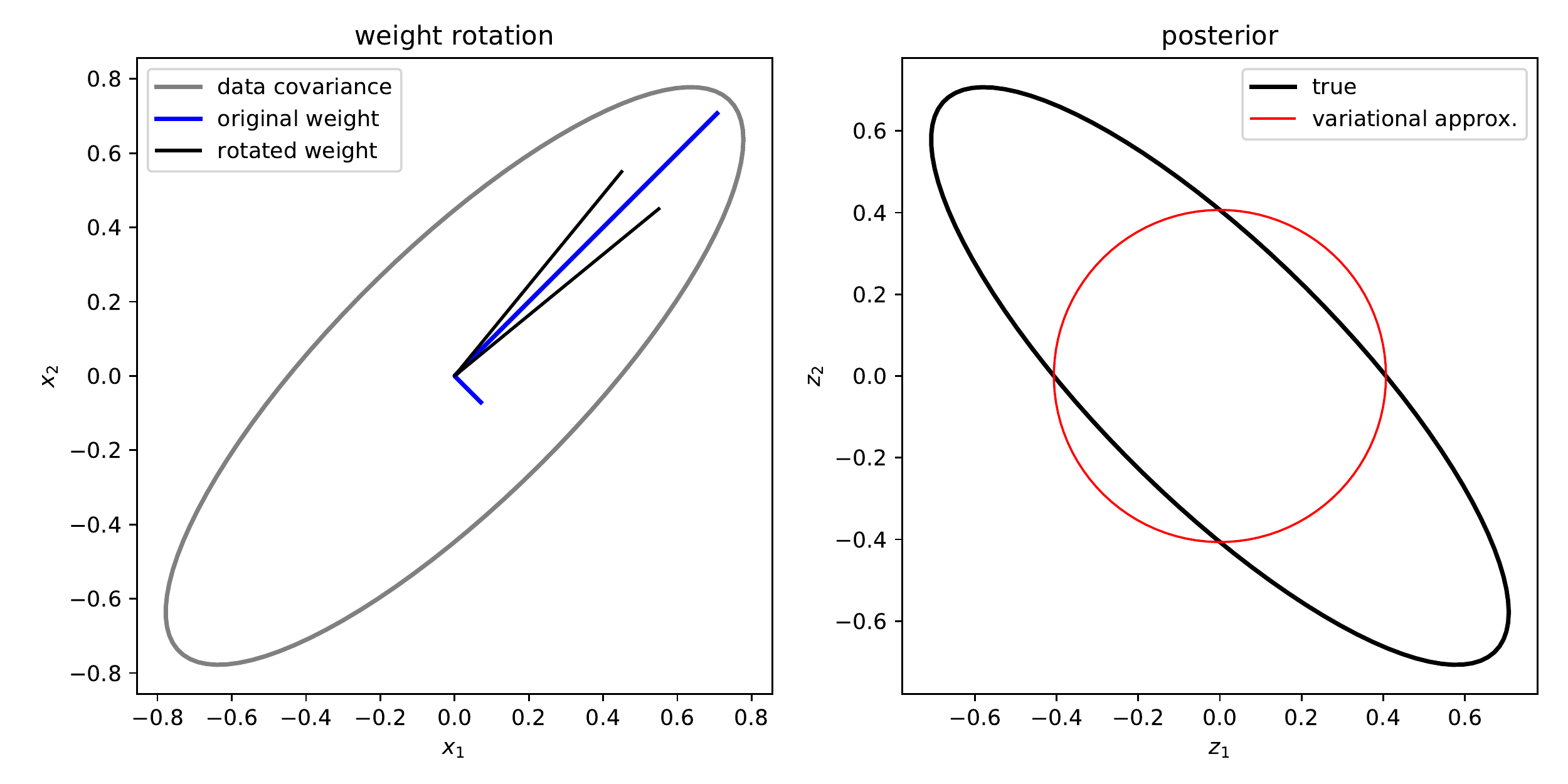}
\end{center}

When varying the rotation away fom the orthogonal setting, $\theta$, the plots above indicate that orthogonal settings of the weights ($\theta = m \pi/2$ where $m = 0,1,2,...$) lead to factorized posteriors. In these cases the KL between the approximate posterior and the true posterior is zero and the free-energy is equal to the log-likelihood. This will be the optimal free-energy for any weight setting (due to the fact that it is equal to the true log-likelihood which is maximal, and the free-energy is a lower bound of this quantity.) For intermediate values of $\theta$ the posterior is correlated and the free-energy is not tight to the log likelihood. 

Now let's plot the free-energy and the log-likelihood as $\theta$ is varied. This shows that the free-energy prefers orthogonal settings of the weights as this leads to factorized posteriors, even though the log-likelihood is insensitive to $\theta$. So, variational inference recovers the same weight directions as the PCA solution.

\begin{center}
\includegraphics[width=0.66\linewidth]{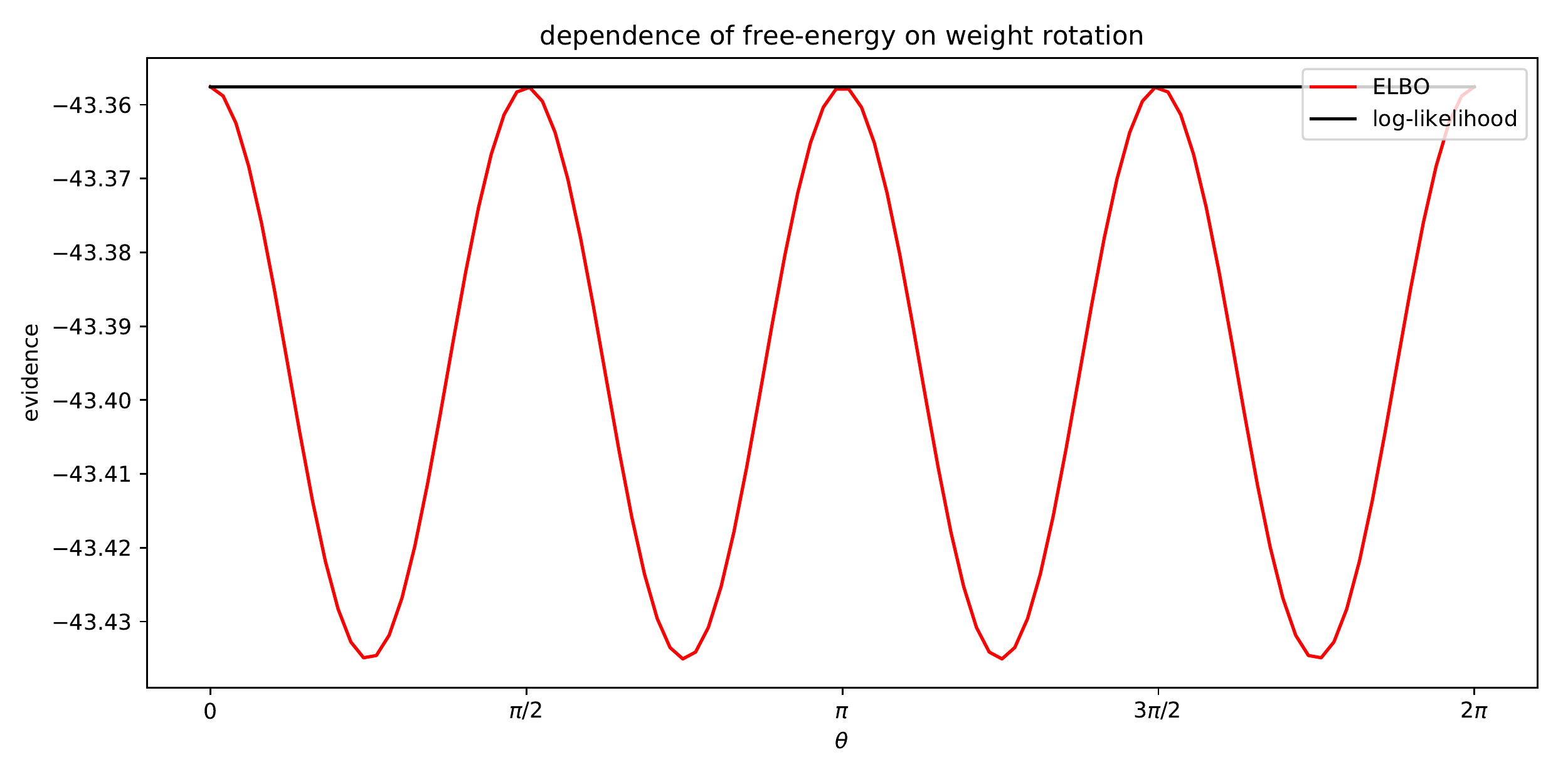}
\end{center}

The above shows that the bias inherent in variational methods will cause them to break the symmetry in the log-likelihood and find orthorgonal latent components. This occurs because orthoginal components result in posterior distributions that are factorized. These are then well-modelled by the variational approximation and result in a small KL between the approximate and true posteriors. 

\subsection{Experiment 2: mean field VI applied to over-complete factor analysis prunes out the additional latent dimensions}

A similar effect occurs if we model 2D data with a 3D latent space. Many settings of the weights attain the maximum of the likelihood, including solutions which use all three latent variables. However, the optimal solution for VI is to retain two orthogonal components and to set the magnitude of the third component to zero. This solution a) returns weights that maximise the likelihood, and b) has a factorised posterior distribution (the pruned component having a posterior equal to its prior) that therefore incurs no cost $\mathrm{KL}(q(\mathbf{z}) || p(\mathbf{z}| \mathbf{x},\theta)) = 0$. In this way the bound becomes tight.

Here's an example of this effect. We consider a model of the form:

\begin{align}
\mathbf{x} = \frac{\alpha}{\sqrt{2}} \left[\begin{array}{c}1\\1 \end{array} \right] z_1 + \frac{\beta}{\sqrt{2}} \left[\begin{array}{c}1\\1 \end{array} \right] z_2 + \frac{\rho}{\sqrt{2}} \left[\begin{array}{c}1\\-1 \end{array} \right] z_3 + \epsilon
\end{align}

We set $\alpha^2+\beta^2=1$ so that all models imply the same covariance and set this to be the maximum likelihood covariance by construction. We then consider varying $\alpha$ from $0$ to $1/2$. The setting equal to $0$ attains the maximum of the free-energy, even though it has the same likelihood as any other setting. 

\begin{center}
\includegraphics[width=0.66\linewidth]{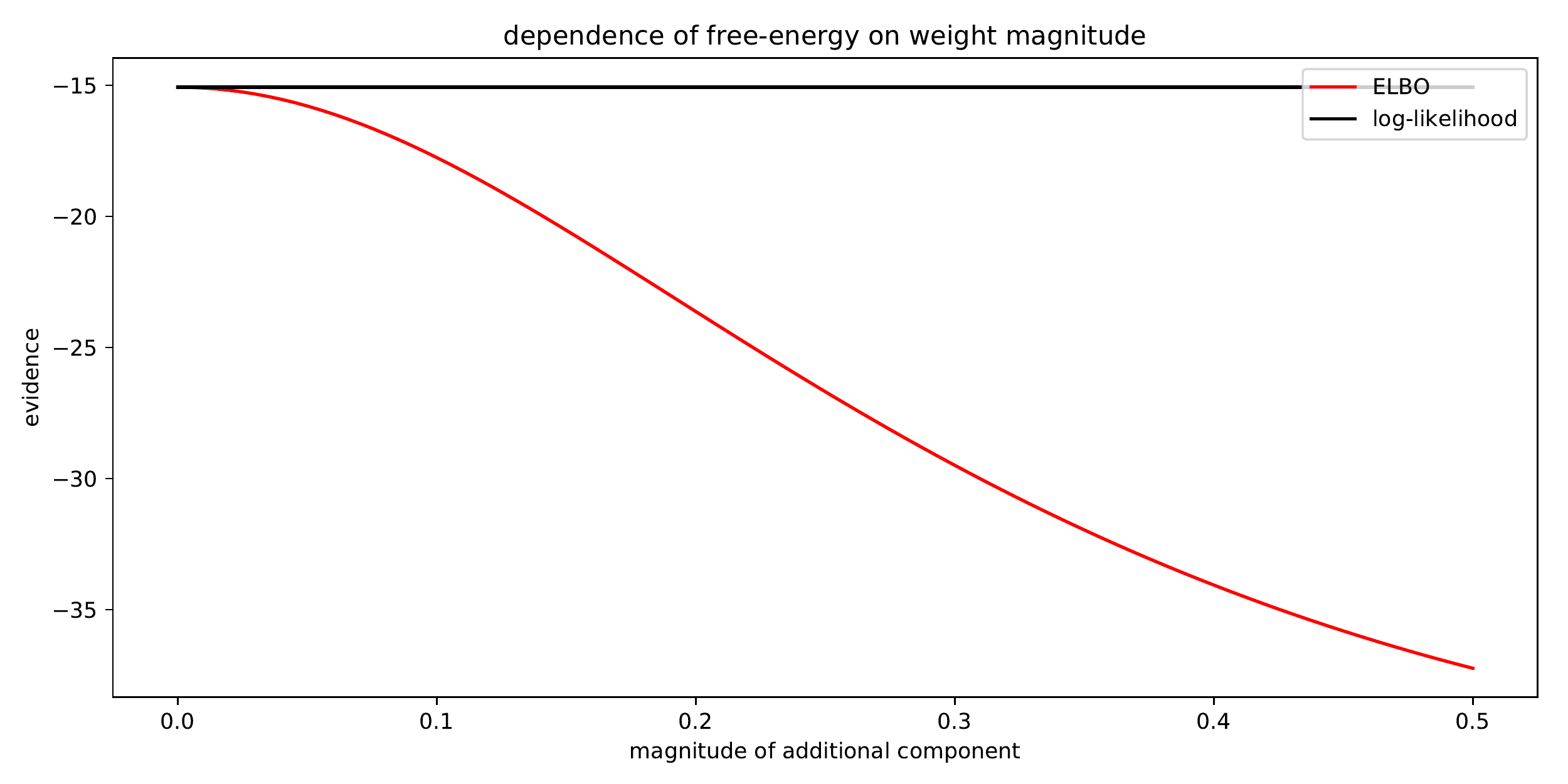}
\end{center}

\subsection{Experiment 3: The $\beta$-VAE also yields the PCA components, changing $\beta$ has no effect on the direction of the estimated components in the FA model}

How does the setting of $\beta$ change things? Here we rerun experiment 1 for different values of $\beta$.

\begin{center}
\includegraphics[width=0.66\linewidth]{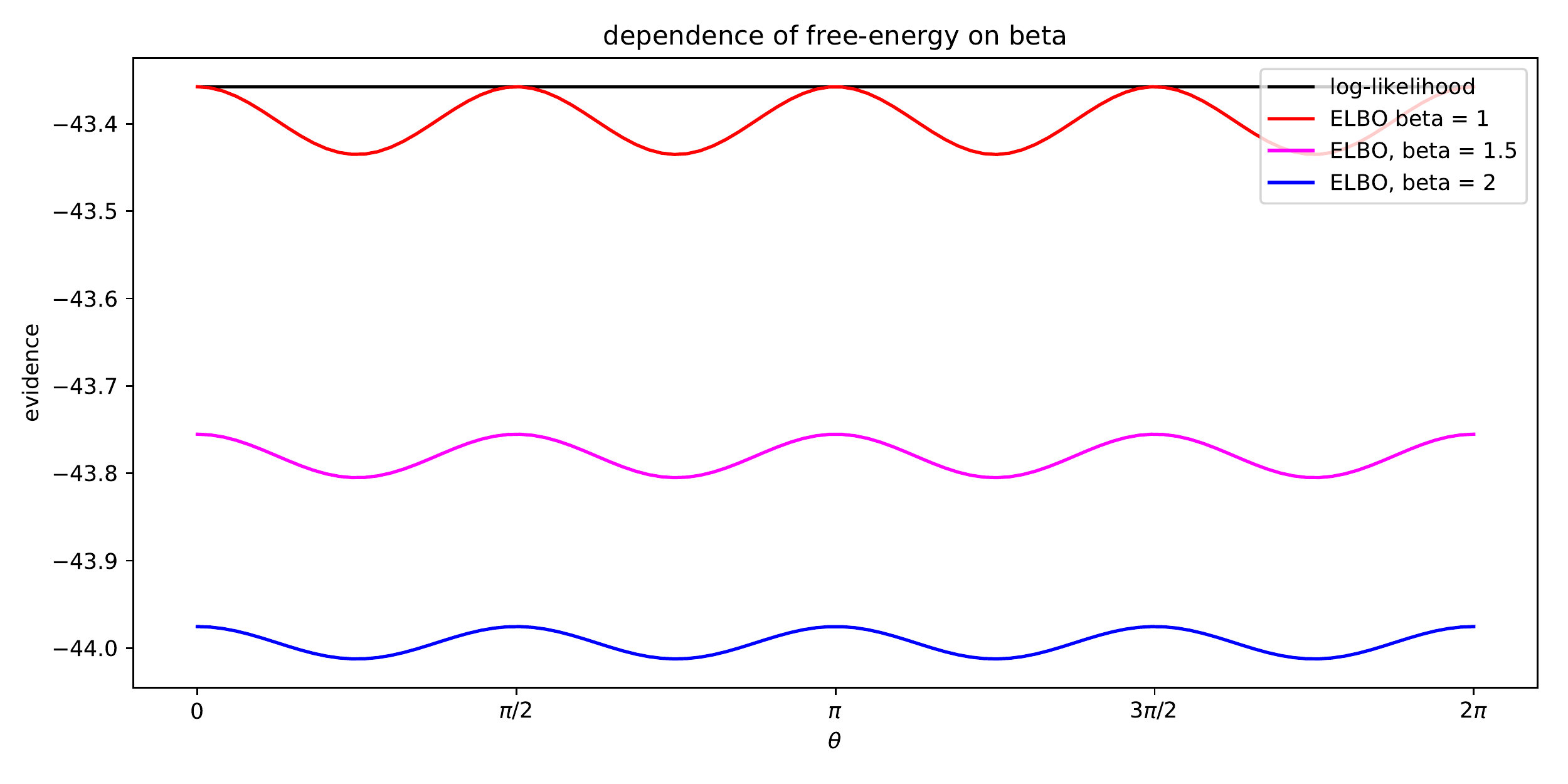}
\end{center}

In this example, changing $\beta$ in this example just reduces the amplitude of the fluctuations in the free-energy, but it does not change the directions found. A similar observation applies to the pruning experiment.

Increasing $\beta$ will increase the uncertainty in the posterior as it is like reducing the number of observations (or increasing the observation noise, from the perspective of $q$).

\subsection{Summary of Factor Analysis Experiments}

The behaviours introduced by the $\beta$-VAE appear relatively benign, and perhaps even helpful, in the linear case: VI is breaking the degeneracy of the maximum likelihood solution in a sensible way: selecting amongst the maximum likelihood solutions to find those that have orthogonal components and removing spurious latent dimensions. This should be tempered by the fact that the $\beta$ generalization recovered precisely the same solutions and so it was necessary to obtain the desired behaviour in the PCA case.

Similar effects will occur in deep generative models, not least since these typically also have a Gaussian prior over latent variables, and these latents are initially linearly transformed, thereby resulting in a similar degeneracy to factor analysis.

However, the behaviours above benefited from the fact that maximum-likelihood solutions could be found in which the posterior distribution over latent variables factorized. In real world examples, for example in deep generative models, this will not be case. In such cases, these same effects will cause the variational free-energy and its $\beta$-generalization to \textbf{bias the estimated parameters far away from maximum-likelihood settings, toward those settings that imply factorized Gaussian posteriors over the latent variables}.

\subsection{Independent Component Analysis}

We now apply VI and the $\beta$ free-energy method to ICA. We're interested the properties of the variational objective and the  $\beta$-VI objective and so we
1. fit the data using the true generative model to investigate the biases in VI and $\beta$-VI
2. do not use amortized inference, just optimizing the approximating distributions for each data point (this is possible for these small examples).

The linear independent component analysis generative model we use is defined as follows. Let $\mathbf{x} \in \mathbb{R}^L$ and $\mathbf{z} \in \mathbb{R}^K$.  

\begin{align}
& \text{for} \;\; n=1...N \nonumber \\
& \;\; \;\;\;\text{for} \;\; k=1...K  \nonumber \\
&\;\; \;\;\;\;\; \;\;\;z_{n,k} \sim \text{Student-t}(0,\sigma,v), \nonumber \\
& \;\; \;\;\;\mathbf{x}_n \sim \mathcal{N}(W \mathbf{z}_n,D) \;\; \text{where} \;\; D = \text{diag}( \mathrm{[\sigma^2_1,...,\sigma^2_D]}) \nonumber 
\end{align}

We apply mean-field variational inference, $q(\mathbf{z}_n) = \prod_k q_{n,k}(z_{k,d})$, and use Gaussian distributions for each factor $q_{n,k}(z_{n,k}) = \mathcal{N}(z_{n,k}; \mu_{n,k},\sigma_{n,k}^2)$.

The free-energy is computed as follows: The reconstruction term is identical to PCA: an avergage of a quadratic form wrt to a Gaussian, which is analytic. The KL is broken down into the differential entropy of q which is also analytic and the cross-entropy with the prior which we evaluate by numerical integration (finite differences). There is a cross-entropy term for each latent variable which is one reason why the code is slow (requiring N 1D numerical integrations). The gradient of the free-energy wrt the parameters $W$ and the means and variances of the Gaussian q distributions are computed using autograd. 

In order to be as certain as possible that we are finding a global maximum of the free-energies, all experiments initialise at the true value of the parameters and then ensure that each gradient step improves the free-energy. Stochastic optimization or a procedure that accepted all steps regardless of the change in the objective would be faster, but they might also move us into the basis of attraction of a worse (local) optima.

\subsection{Experiment 1: Learning in over-complete ICA}
\label{sec:ica-example}

Now we define the dataset. We use a very sparse Student's t-distribtion with $v=3.5$. For $v<4$ the the kurtosis is undefined so the model is fairly simple to estimate (it's a long way away from the degenerate factor analysis case which is recovered in the limit $v \rightarrow \infty$).

We use three latent components and a two dimensional observed space. The directions of the three weights are shown in blue below with data as blue circles.

\begin{center}
\includegraphics[width=0.66\linewidth]{figures/supplement/ICA-model.pdf}
\end{center}

First we run variational inference finding components (shown in red below) which are more orthogonal than the true directions. This bias is in this directions as this reduces the dependencies (explaining away) in the underlying posterior.

\begin{center}
\includegraphics[width=0.66\linewidth]{figures/supplement/ICA-beta-1.pdf}
\end{center}

Second we run $\beta$-VI with $\beta=5$. Two  components are now found that are orthogonal with one component pruned from the solution.
\begin{center}
\includegraphics[width=0.66\linewidth]{figures/supplement/ICA-beta-5.pdf}
\end{center}
In this case the bias is so great that the true component directions are not discovered. Instead the components are forced into the orthogonal setting regardless of the structure in the data.

\subsection{Summary of Independent Component Analysis experiment}

The ICA example illustrates that this approach -- of relying on a bias inherent in VI to discover meaningful components -- will sometimes return meaningful structure (e.g. in the PCA experiments above). However it does not seem to be a sensible way of doing so in general. For example, explaining away often means that the true components will be entangled in the posterior, as is the case in the ICA example, and the variational bias will then move us away from this solution. The $\beta$-VI generalisation only enhances this undesirable bias.